\documentclass[letterpaper, 10 pt, journal, twoside]{IEEEtran}
\IEEEoverridecommandlockouts

\usepackage{soul}

\usepackage{cite}
\usepackage{textcomp}

\usepackage{amsmath,amssymb}
\usepackage{contour}
\usepackage[cal = pxtx, scr = dutchcal]{mathalpha}

\usepackage{fontawesome}
\usepackage{colortbl}
\usepackage{wrapfig}
\usepackage{graphicx}
\usepackage[dvipsnames]{xcolor}

\usepackage[colorlinks, linkcolor=black, citecolor=black]{hyperref}
\usepackage{subfig}
\usepackage{caption}
\usepackage{float}
\usepackage{romannum}
\usepackage[T1]{fontenc}
\usepackage{fontawesome}

\usepackage{multirow}
\usepackage{booktabs}
\usepackage{tcolorbox}
\usepackage{makecell}
\usepackage{hhline}
\usepackage{scalerel}
\usepackage{listings}

\usepackage{footnote}
\makesavenoteenv{tabular}

\usepackage{array}
\usepackage{xspace}
\usepackage{bigstrut}
\usepackage[export]{adjustbox}
\usepackage{ragged2e}
\usepackage[linesnumbered, ruled,vlined]{algorithm2e}

\usepackage{stfloats}
\usepackage{paralist}
\usepackage{framed}
\usepackage{color}
\usepackage{tabularray}

\usepackage{stfloats}

\usepackage{boldline}
\usepackage{orcidlink}

\newcommand{\curly}{\mathrel{\leadsto}}

\newcommand{\texTTT}{\fontfamily{lmtt}\selectfont}

\let\oldnl\nl
\newcommand{\nonl}{\renewcommand{\nl}{\let\nl\oldnl}}

\DeclareFontFamily{OT1}{pzc}{}
\DeclareFontShape{OT1}{pzc}{m}{it}{<-> s * [1.2] pzcmi7t}{}
\DeclareMathAlphabet{\mathpzc}{OT1}{pzc}{m}{it}
\DeclareMathAlphabet{\mathdutchcal}{U}{dutchcal}{m}{n}

\newcommand{\ourapproach}{{\small \sf CURE}\xspace}

\usepackage{tikz}

\usetikzlibrary{arrows.meta}

\newcommand\edgeone{
\begin{tikzpicture}
  \draw[black,  arrows={-Triangle[angle=90:3pt,black,fill=black]}] (0,0.0) -- (0.5,0.0);
\end{tikzpicture}\xspace}

\newcommand\edgetwo{
\begin{tikzpicture}
  \draw[black,  arrows={Triangle[angle=90:3pt,black,fill=black]-Triangle[angle=90:3pt,black,fill=black]}] (0,0.0) -- (0.5,0.0);   
\end{tikzpicture}\xspace}

\newcommand\edgethree{
\begin{tikzpicture}
  \draw[black,  arrows={Circle[open]-Triangle[angle=90:3pt,black,fill=black]}] (0,0.0) -- (0.5,0.0);   
\end{tikzpicture}\xspace}

\newcommand\edgefour{
\begin{tikzpicture}
  \draw[black,  arrows={Circle[open]-Circle[open]}] (0,0.0) -- (0.5,0.0);   
\end{tikzpicture}\xspace}

\DeclareMathOperator*{\argmax}{arg\,max}
\DeclareMathOperator*{\argmin}{arg\,min}

\newcommand{\bs}{\boldsymbol}
\renewcommand{\mathbf}[1]{{\boldsymbol #1}}
\hypersetup{
	colorlinks   = true,
	citecolor    = blue,
	linkcolor    = blue,
    urlcolor     = blue,
}

\begin{document}
\bstctlcite{IEEEexample:BSTcontrol}

\title{\sf{CURE}: Simulation-Augmented Auto-Tuning in Robotics}

\author{Md Abir Hossen$^{\orcidlink{0000-0002-7956-0515}}$, Sonam Kharade$^{\orcidlink{0000-0001-7905-9387}}$, Jason M. O'Kane$^{\orcidlink{0000-0002-1536-4822}}$, \textit{Senior Member, IEEE}, \\ Bradley Schmerl$^{\orcidlink{0000-0001-7828-622X}}$, \textit{Senior Member, IEEE,} 
David Garlan$^{\orcidlink{0000-0002-6735-8301}}$, \textit{Life Fellow, IEEE,} and Pooyan Jamshidi$^{\orcidlink{0000-0002-9342-0703}}$

\thanks{This work was supported by National Science Foundation~(Award 2107463) and National Aeronautics and Space Administration~(Award 80NSSC20K1720).}
\thanks{M. A. Hossen, S. Kharade, and P. Jamshidi are with the University of South Carolina, SC, USA (e-mail: abir.hossen786@gmail.com; skharade@mailbox.sc.edu; pjamshid@cse.sc.edu)}
\thanks{J. M. O'Kane is with Texas A\&M University, TX, USA (e-mail: jokane@tamu.edu)}
\thanks{B. Schmerl, and D. Garlan are with the Carnegie Mellon University, PA, USA~(e-mail: schmerl@cs.cmu.edu; garlan@cs.cmu.edu)}
}

\maketitle

\begin{abstract}
Robotic systems are typically composed of various subsystems, such as localization and navigation, each encompassing numerous configurable components~(e.g., selecting different planning algorithms). Once an algorithm has been selected for a component, its associated configuration options must be set to the appropriate values. Configuration options across the system stack interact non-trivially. Finding optimal configurations for highly configurable robots to achieve desired performance poses a significant challenge due to the interactions between configuration options across software and hardware that result in an exponentially large and complex configuration space. These challenges are further compounded by the need for transferability between different environments and robotic platforms. Data efficient optimization algorithms~(e.g., Bayesian optimization) have been increasingly employed to automate the tuning of configurable parameters in cyber-physical systems. However, such optimization algorithms converge at later stages, often after exhausting the allocated budget~(e.g., optimization steps, allotted time) and lacking transferability. This paper proposes {\small\sf{CURE}}---a method that identifies causally relevant configuration options, enabling the optimization process to operate in a reduced search space, thereby enabling faster optimization of robot performance. \ourapproach abstracts the causal relationships between various configuration options and robot performance objectives by learning a causal model in the source~(a low-cost environment such as the Gazebo simulator) and applying the learned knowledge to perform optimization in the target~(e.g., \textit{Turtlebot 3} physical robot). We demonstrate the effectiveness and transferability of \ourapproach by conducting experiments that involve varying degrees of deployment changes in both physical robots and simulation.
\end{abstract}

\begin{IEEEkeywords}
robotics and cyberphysical systems, causal inference, optimization, robot testing.
\end{IEEEkeywords}

\setlength{\textfloatsep}{20pt}

\section{Introduction}
\IEEEPARstart{A} robotic system is composed of hardware and software components that are integrated within a physical machine. These components interact to achieve specific goals in a physical environment. Unfortunately, robots are prone to a wide variety of faults~\cite{khalastchi2018fault}. Incorrect configurations~(called \textit{misconfigurations}) in robotic algorithms are one of the most prevalent causes of such faults~\cite{kim2023patchverif, jung2021swarmbug, 10137745}. Misconfigurations can cause various bugs~\cite{xie2022rozz, kim2019rvfuzzer} leading to crashes, robots becoming unstable, deviations from planned trajectory, controller faults, and non-responsiveness. Several studies have reported misconfigurations as one of the key reasons for cyber-physical system failures. Such misconfigurations caused $19.6\%$ of Unmanned Aerial Vehicle~(UAV) bugs~\cite{wang2021exploratory}, $27.25\%$ of autonomous vehicle bugs~\cite{garcia2020comprehensive} (a faulty configuration in actuation layer even caused the vehicle to collide with a static object on the curb~\cite{kim2022drivefuzz}) and $55\%$ of traffic dispatch algorithm bugs~\cite{valle2023automated}. All of these issues were fixed by configuration changes.

\begin{figure}[!t]
\vspace{-0.3em}
\centering
\centerline{\includegraphics[width=.94\columnwidth, trim={1em 0 0.8em 0.5em},clip]{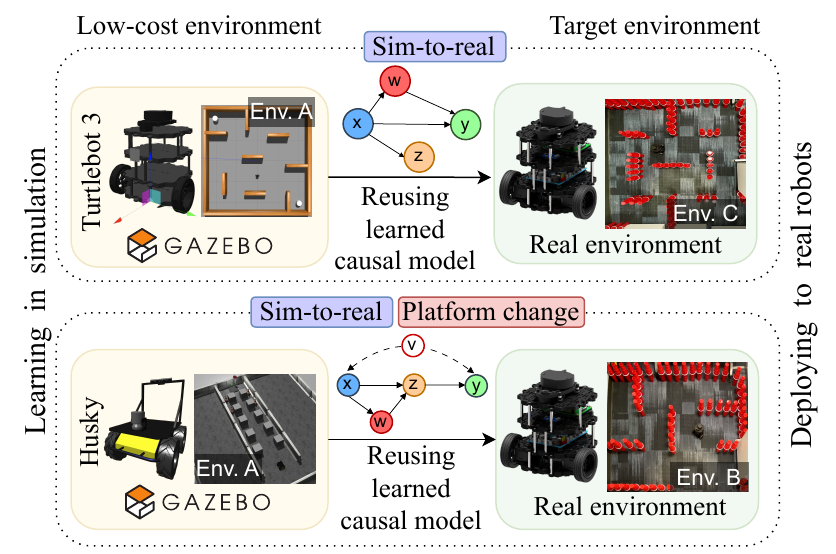}}
\caption{\small{Sim-to-real: applying the knowledge of the learned causal model using \textit{Turtlebot 3} in simulation to the \textit{Turtlebot 3} physical robot. Sim-to-real \& Platform change: transferring the causal model learned using \textit{Husky} in simulation to the \textit{Turtlebot 3} physical robot.}}
\label{fig:trans_setup}
\vspace{-1em}
\end{figure}

Most robotic algorithms require customization through configuration parameters to suit certain tasks and situations. For example, most UAV controllers include a wide range of configurable parameters that can be customized to different vehicles, flight conditions, or even particular tasks~(e.g., when speed is more important than energy use). Finding configurations that optimize performance on a given task is a challenging problem for designers and end users~\cite{configOpti}. A developer might request a feature such as \emph{``Create a tool to automatically tune navigation2 node parameters using state-of-the-art machine learning techniques."}~\cite{tuningfeature}. In another instance, a developer encounters a planner performance issue~\cite{plannerfailing} and asks \emph{``I have tuned this for almost 5-6 hours. Sometimes it is going towards the goal but still failing in the middle of the trajectory.''} After several back-and-forth communications, the algorithm designer concludes, \emph{``I cannot provide personalized tuning assistance to every user.''} Additionally, developers aim to maintain the performance of the tuned parameters when deployment changes~(e.g., from \textit{ROS1} to \textit{ROS2}) to avoid re-tuning. Specifically, the optimal configuration determined in one environment often becomes suboptimal in another, as demonstrated in Fig.~\ref{fig:suboptimal}. 

\noindent\textbf{Our Solution.} In this work, we propose \ourapproach~(\underline{C}ausal \underline{U}nderstanding and \underline{R}emediation for \underline{E}nhancing Robot Performance), a multi-objective optimization method that finds optimal configurations for robotic platforms, converges faster than the state-of-the-art, and transfers well from simulation to real robot and even to new untrained platforms. \ourapproach has two main phases. In Phase 1, \ourapproach reduces the search space by eliminating configuration options that do not affect the performance objective causally. For this, we collect observational data in a low-cost source environment, such as simulation. Then, a causal model is learned on the basis of the data, representing the underlying causal mechanisms that influence robot performance. We then estimate the causal effects of options on performance objectives. Finally, we reduce the search space to a subset of options that have non-negligible causal effects. In Phase 2, \ourapproach performs traditional Bayesian optimization in the target environment, but only over the reduced search space, to find the optimal configuration. We show that \ourapproach not only finds the optimal configuration faster than the state-of-the-art, but the learned causal model in the simulation speeds up optimization in the real robot. The results demonstrate that the learned causal model is transferable across similar but different settings, that is, environments, mission/tasks, and for new robotic platforms. In other words, the existence of a common abstract structure~(the causal relations between options, system-level variables, and performance objectives) is invariant across domains, and the behavior of specific features of the environment remains constant across domains. 

\noindent\textbf{Evaluations.} 
We evaluated \ourapproach in terms of its \emph{effectiveness} and \emph{transferability} across two tasks: navigation and manipulation. The navigation task forms the core of our experiments, using two highly configurable robotic systems~(\textit{Husky} and \textit{Turtlebot 3}) under varying degrees of deployment changes. The manipulation task involves simulating a robot arm~(\emph{Franka Emika Panda}) in \emph{Gazebo} to demonstrate \ourapproach's adaptability by complementing the effectiveness evaluation. We compared \ourapproach with traditional multi-objective Bayesian optimization~(MOBO) using the \textsc{Ax} framework~\cite{Ax}, and RidgeCV~\cite{hoerl1970ridge, ridgeCV} integrated with MOBO to reduce the search space. Our results indicate that compared to MOBO, \ourapproach finds a configuration that improves performance by $2\times$ and achieves this improvement with gains in efficiency of $4.6\times$ when we transfer the knowledge learned from \textit{Husky} in simulation to \textit{Turtlebot 3} physical robot.

\noindent\textbf{Contributions.} The contributions of our work are as follows:
\begin{itemize}
    \item We propose \ourapproach, a multi-objective optimization method that operates in the reduced search space involving causally relevant configuration options and allows faster convergence.
    
    \item We conducted a comprehensive empirical study by comparing \ourapproach with state-of-the-art optimization methods in both simulation and real robots under different severities of deployment changes, and studied effectiveness and transferability.
    
    \item The code and data are available at: \url{https://github.com/softsys4ai/cure}
\end{itemize}
\section{Related Work}
In this work, we focus on performance optimization through the lens of causality. Specifically, we learn a causal model from a low-cost environment and utilize causal knowledge to optimize performance in the target system. This section groups related work into four categories: optimizing robotic parameters, machine learning for performance modeling, transfer learning strategies, and causal analysis in configurable systems.

\paragraph{Optimization techniques in robotic configurations}
Researchers have considered robotic algorithms as a black box, as the objective functions in most robotic problems can only be accessible through empirical experiments. Evolutionary algorithms~\cite{binch2020context, zhou2011multiobjective} have been used to find optimal configurations in Dynamic‐Window Approach~(DWA)~\cite{fox1997dynamic} algorithm. However, the application of evolutionary algorithms in robotic systems is hindered by the limited availability of observations and the difficulty in extracting meaningful information from these observations due to the presence of noise. Approaches such as variational heteroscedastic Gaussian process regression (VHGP)~\cite{ariizumi2016multiobjective} and Bayesian optimization with safety constraints~\cite{berkenkamp2023bayesian} attempt to address these challenges, but struggle with high-dimensional search spaces, yield only local improvements, and lack transferability across different environments and platforms. Furthermore, the complexity of environmental dynamics models, coupled with the biases introduced by optimization formulation, poses significant challenges. Moreover, formalizing safety constraints that allow for computationally efficient solutions, specifically solutions in polynomial time with closed-form expressions, is complex if at all feasible.

\paragraph{Learning based methods for performance modeling}
Expanding on traditional optimization techniques, machine learning methods offer diverse approaches to improve robotic performance. Approaches such as learning from demonstration~\cite{argall2009survey}, learning human-aware path planning~\cite{perez2018learning}, and mapping sensory inputs to robot actions~\cite{pfeiffer2017perception, kahn2021badgr} have been widely applied to robot navigation beyond fine-tuning configuration parameters, as opposed to heavily relying on human expertise. These methods aim to replace classical methods, casting doubt on the robustness, generality, and safety of the systems. To provide a deeper understanding of performance behavior in robotic algorithms, performance influence models~\cite{chen2022performance, siegmund2015performance, 8919029} can be used. These models predict system performance by capturing important options and interactions that influence performance behavior using machine learning and sampling heuristics. However, performance influence models face limitations in adapting to unexpected environments due to not being able to capture changes in the performance distribution and often produce incorrect explanations~\cite{10137745}. In addition, the collection of training data for these models is costly and requires extensive human supervision.

\paragraph{Transfer learning for performance modeling}
Addressing the challenges of adapting to unexpected environments and costly data collection in learning-based methods, transfer learning accelerates optimization by selectively reusing knowledge from previous tasks. Techniques such as simulation-to-real learning~\cite{rai2019using, kaushik2022safeapt} and transferring Pareto frontiers across different platforms~\cite{valov2020transferring} improve sampling efficiency and improve training data sets. Each of these techniques uses the predicted transfer learning frameworks based on correlational analysis. However, changes in the environment and robotic platform can cause a distribution shift. The ML models used in these transfer learning approaches are vulnerable to spurious correlations~\cite{iqbal2023cameo, zhou2021examining}.

\paragraph{Causal analysis in configurable systems}
While machine learning techniques excel in uncovering correlations between variables, their ability to identify causal links is limited~\cite{pearl2009causality}. Using the information encoded in causal models, we can benefit from analyses that are only possible when
we explicitly employ causal models, such as interventional and counterfactual analyses~\cite{pearl2009causality, spirtes2000causation}. Causal analysis has been used for various debugging and optimization tasks in configurable systems, including finding the root cause of intermittent failures in database applications~\cite{fariha2020causality}, detecting and understanding the root causes of the defect~\cite{johnson2020causal, DubslaffCausality2022}, and improving fault localization~\cite{9402143}. The causality analysis in these studies is confined to a single environment and platform, while our approach transfers causal knowledge across different environments and platforms. In robotic systems, the causal models learned in simulation are used to find explanations for failures in real robots~\cite{diehl2022did, 10137745}. However, such methods are limited to identifying root causes of failures, whereas our approach extends beyond diagnosis to also prescribe remedies, new configuration option values that rectify the failure.

\section{Problem Formulation and Challenges} \label{sec:problem description}
In this section, we first motivate our work by illustrating how an optimal configuration found in one environment often becomes suboptimal in another. We then formally define the problem and describe the challenges.

\subsection{Motivating scenario}
\begin{figure}[!t] 
 \vspace{-1em}
  \subfloat[]{\includegraphics[width=.5\columnwidth]{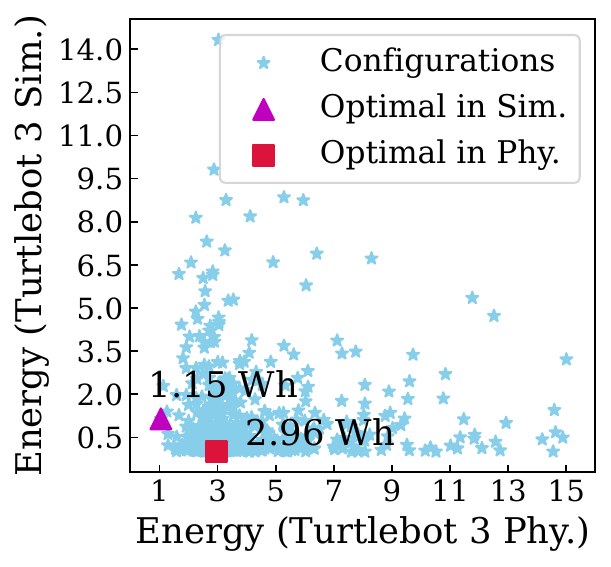}\label{fig:sim_to_phy}}
  \hfill
  \subfloat[]{\includegraphics[width=0.5\columnwidth,]{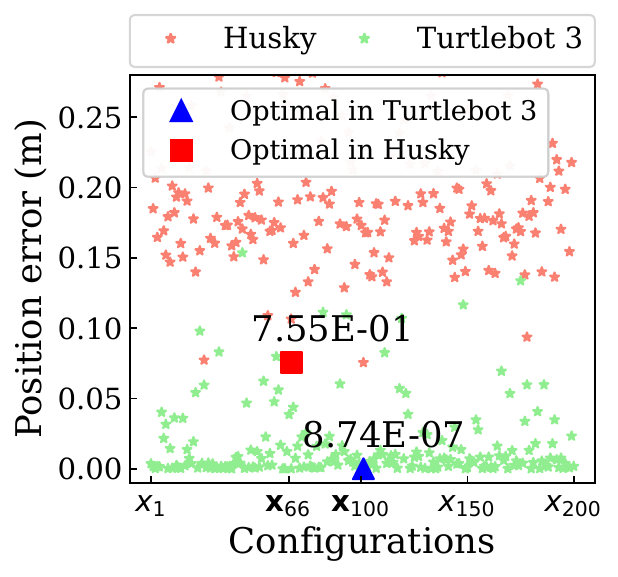}\label{fig:husky_to_tbot}}
  \caption{\small{Non-transferability of optimal configurations across different environments/platforms: (a)~optimal configuration for \textit{Turtlebot 3} in simulation differs from its physical counterpart; and (b)~optimal configuration for \textit{Turtlebot 3} is not suitable in \textit{Husky}.}
  \label{fig:suboptimal}}
  \vspace{-0.5em}
\end{figure}

We motivate our work by demonstrating the non-transferability of traditional Bayesian optimization through a simple experiment for robot navigation. In particular, we explore two deployment scenarios: (i) \textbf{Sim2Real}:~transferring the optimal configurations for energy consumption identified from simulations to the \textit{Turtlebot~3} physical robot~(Fig.~\ref{fig:sim_to_phy}), and (ii) \textbf{Real2Real}:~transferring the optimal configurations for position error\footnote{defined as the Euclidean distance between goal position and robot's actual position} identified from \textit{Husky} to \textit{Turtlebot~3}~(Fig.~\ref{fig:husky_to_tbot}). In both scenarios, we observe that the optimal configurations identified by Bayesian optimization in the source environments fail to retain their optimality in the target environment. We observe that energy consumption increases by $2.57\times$, and a significant increase in position error is observed by $8.64\times10^5$ times.

\subsection{Problem formulation}
Consider a highly configurable robot with $d$ distinct configurations. Let $X_i$ indicate the configuration parameter $i$, which can be assigned a value from a finite domain $Dom(X_i)$. In general, $X_i$ may be set to (i) a real number (e.g. the number of iterative refinements in a localization algorithm, the frequency of the controller) within specified bounds, denoted as $X_i \in [\underline{X}_i, \Bar{X}_i]$, where $\underline{X}_i$ and $\Bar{X}_i$ are the lower and upper bounds, respectively, (ii) binary (e.g., whether to enable recovery behaviors) or (iii) categorical (e.g., planner algorithm names). The configuration space is mathematically a Cartesian product of all the domains of the parameters of interest $\mathcal{X}=Dom(X_1) \times \dots \times Dom(X_d)$. Then, a configuration $\mathbf{x}$, which is in the configuration space $\mathbf{x} \in \mathcal{X}$, can be instantiated by setting a specific value for each option within its domain, $\mathbf{x}=\langle X_1=x_1, X_2=x_2, \dots, X_d=x_d \rangle$. Finding a configuration that uniformly optimizes all objectives is typically not possible; instead, there is a trade-off between them. Pareto optimal solutions signify the prime balance among all objectives. In the context of minimization, a configuration $\mathbf{x}$ is said to \emph{dominate} another configuration $\mathbf{x'}$ if $f(\mathbf{x}) \leq f(\mathbf{x'})$. A configuration $\mathbf{x} \in \mathcal{X}$ is called \emph{Pareto-optimal} if it is not dominated by any other configuration $\mathbf{x'} \in \mathcal{X}$, where $\mathbf{x} \not = \mathbf{x'}$. The goal is to find $\mathbf{x}^*$, a configuration that gives rise to Pareto-optimal performance in the multi-objective space~(e.g., $f_1:$ failure rate, $f_2:$ mission time, $f_3:$ energy consumption), given some constraints ($h:$ safety). Here, we assume that the performance measure can be evaluated in experiments for any configuration $\mathbf{x}$, and we do not know the underlying functional representation of the performance. The problem can be generalized by defining an arbitrary number of performance objectives (if they can be computed over a finite time horizon). Mathematically, we represent performance objectives as black-box functions that map from a configuration space to a real-valued one: $f (\mathbf{x}): \mathcal{X} \rightarrow\mathcal{R}$. In practice, we learn $f$ by sampling the configuration space and collecting the observations data, i.e.,~$y_i=f(\mathbf{x}_i)+\epsilon_i$ with $\epsilon \sim \mathcal{N}(0,\sigma^{2})$. In other words, we only partially know the response function through observations $\mathcal{D}=\{(\mathbf{x}_i,y_i)\}_{i=1}^d, |\mathcal{D}|\ll|\mathcal{X}|$. We define the problem formally as follows:
\begin{equation}
    \label{eq:problem}
    \boldsymbol{x^*} = \argmin_{\mathbf{x} \in \mathcal{X}} f_1(\mathbf{x}), f_2(\mathbf{x}),\dots, f_m(\mathbf{x}), s.t.: \space h(\mathbf{x}) \geq 0,
\end{equation}
where $\mathbf{x^*} \in \mathcal{X}$ is a Pareto-optimal configuration and adhere to the safety constraints.

\begin{figure*}[!t]
\begin{center}
\centerline{\includegraphics[width=2\columnwidth]{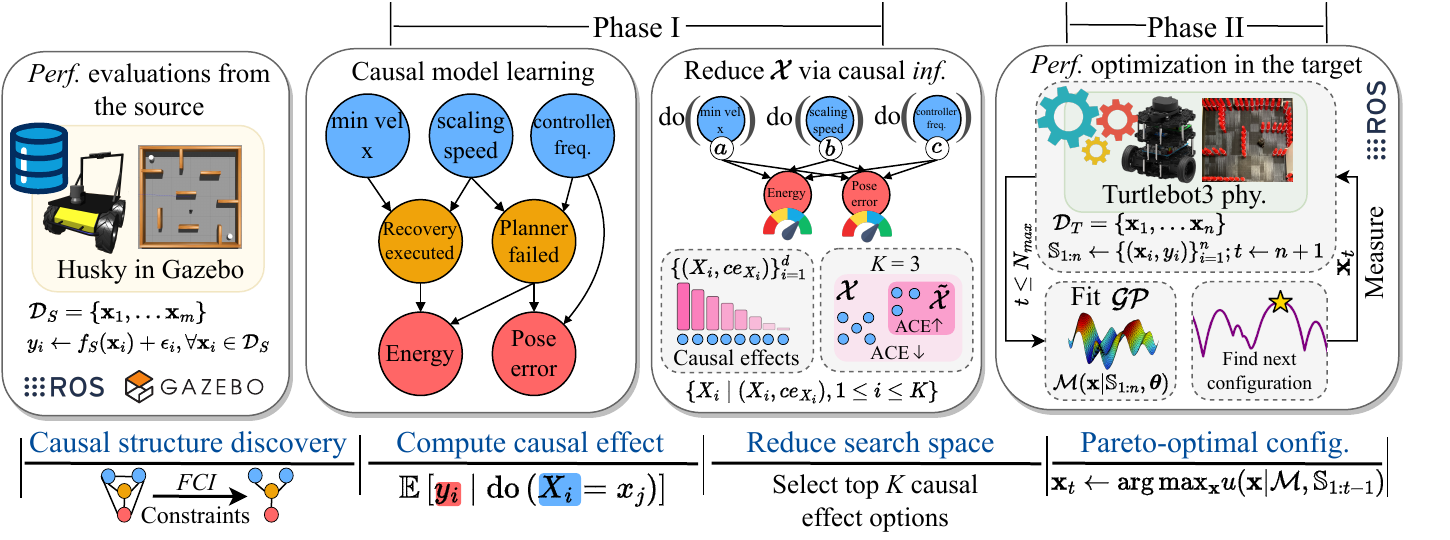}}
\caption{\small{\ourapproach overview.}}
\label{fig:cure_overview}
\end{center} 
\vspace{-2em}
\end{figure*}

\subsection{Challenges}
In this article, our objective is to propose a solution to address the following key challenges:

\paragraph{Software-hardware interactions and exponentially growing configuration space} 
A robotic system consists of software components (e.g., localization, navigation, and planning), hardware components (e.g., computer and sensors onboard), and middleware components (e.g., ROS), with most components being configurable. The configuration space of only 100 parameters with only 10 possible values for each comprises of $10^{100}$ possible configurations.  (For comparison, the number of atoms in the universe is estimated to be only $10^{82}$.) Therefore, the task of finding Pareto-optimal configurations for highly configurable robots and other cyberphysical systems is orders of magnitude more difficult because of software-hardware interactions, compared with software systems. 

\paragraph{Reality gap and negative transfer from sim to real} 
Robot simulators have been extensively used in testing new behaviors before the new component is used in real robots. However, the measurements from simulators typically contain noise, and the observable effect for some configuration options may not be the same in a real robot operating in a real environment, and in some cases, such effect may even have the opposite effect. Therefore, any reasoning based on the model predictions learned based on simulation data may become misleading. Such a reality gap between the sim and real exists due to unobservable confounders as a result of simplifications in the sim. Still, there exist stable relationships between configuration options and performance objectives in the two environments that can facilitate performance optimization of real robots.

\paragraph{Multiple objectives} 
It is common to find multiple performance objectives in mission specifications~(e.g., mission time, energy, and safety). Typically, the objectives involved in the specification are independent of each other~\cite{abdessalem2020automated}, but in some cases they can be correlated and conflicting; for example, faster task completion could lead to higher energy consumption. Therefore, finding the optimal configuration~(for a given robotic platform in a specific environment and for a specific task) should be treated as a multi-objective optimization problem.

\paragraph{Costly acquisition of training data and the safety critical nature of robotic systems}
Algorithm parameters can be manually adjusted by experiments on real robots or by using massive amounts of training data when the robotic system contains elements that are difficult to hard-code~(e.g., computer vision components)~\cite{lillicrap2015continuous}. However, collecting training data from real robots is time-consuming and often requires constant human supervision~\cite{gupta2018robot}. To guarantee the safe behavior of the robot, the practitioner must either meticulously select configurations that are safe or acquire an ample amount of representative data that lead to safe behavior.

\section{CURE: \underline{C}ausal \underline{U}nderstanding and \underline{R}emediation for \underline{E}nhancing Robot Performance} \label{sec:method}

To solve the optimization problem described in~\S\ref{sec:problem description}, we propose a novel approach, called \ourapproach. The high-level overview of \ourapproach is shown in Fig.~\ref{fig:cure_overview}. \ourapproach works in two phases. In Phase I, \ourapproach reduces the search space for the optimization problem using data from the source environment, while in Phase II, \ourapproach performs a black-box optimization in the reduced search space on the target platform. To elaborate on the details, in Phase I, \ourapproach learns a structural causal model that enforces structural relationships and constraints between variables using performance evaluations from the source platform~(e.g., \textit{Husky} in simulation). Specifically, we learn a causal model for a set of random samples\footnote{Instead of random samples, other partial designs (e.g., Latin Hypercube) could have been used, however, we experimentally found that random samples give rise to more reliable conditional independence tests in the structure learning algorithm.} taken in the source environment\footnote{Here the source environment could be a simulator like Gazebo or another robotic platform. The assumption is that the source is an environment in which we can intervene at a lower cost.}. The configuration options are then ranked by measuring their \emph{average causal effect} on the performance objectives through causal interventions. Options with the largest causal effect are selected to reduce the search space. Next, in Phase II, \ourapproach performs a black-box optimization in the reduced search space given a fixed sampling budget in the target platform~(e.g., the physical \textit{Turtlebot 3}). Specifically, \ourapproach searches for Pareto-optimal configurations in the target, iteratively fits a surrogate model to the samples, and selects the next sample based on an acquisition function until the budget is exhausted. \ourapproach's high-level procedure is described in Algorithm \ref{alg:cure-alg}.

\subsection{Phase I: Reducing the search space via causal inference}
Phase~I begins by recording performance metrics for $s$ initial configurations $\{(\mathbf{x}_1,y_1), \dots ,(\mathbf{x}_{s},y_{s})\}$ in the source environment~(Algorithm~\ref{alg:cure-alg}: lines 1-2). We define three types of variables to learn the causal structure: (i)~software-level configuration options~(e.g., hyperparameters in different algorithms~\cite{navcore}) and hardware-level options (e.g., sensor frequency), (ii)~intermediate performance metrics~(e.g., different system events in ROS) that map the influence of configuration options on performance objectives, and (iii)~end-to-end performance objectives (e.g., task completion rate, mission time). We also define structural constraints~(e.g., $X_i \nrightarrow X_j$) over the causal structure to incorporate domain knowledge that facilitates learning with low sample sizes\footnote{e.g., there should not be any causal connections between configuration options and their values are determined independently.}. 

To discover the causal structure, we use an existing structure learning algorithm \emph{Fast causal inference}~(FCI). We select FCI because (i) it can identify unobserved confounders~\cite{spirtes2000causation, glymour2019review}, and (ii) it can handle variables of various typologies, such as nominal, ordinal, and categorical given a valid conditional independence test. Algorithm \ref{alg:scm-learn} describes the details of our causal learning procedure. It starts by constructing an undirected fully connected graph $G$, where the nodes represent the variables (options, intermediate variables, performance metrics). Next, we evaluate the independence of all pairs of variables conditioned on all remaining variables using Fisher’s~z test~\cite{connelly2016fisher} to remove the edges between independent variables. Finally, a \emph{partial ancestral graph}~(PAG) is generated~(Algorithm~\ref{alg:scm-learn}: line 2), orienting the undirected edges using the edge orientation rules~\cite{spirtes2000causation,glymour2019review,colombo2012learning}. 

\begin{algorithm}[t]
\DontPrintSemicolon
\SetKwFor{ForEach}{for each}{do}{end}
\SetKwRepeat{Do}{do}{while}
\caption{\ourapproach}\label{alg:cure-alg}
\KwIn{Configuration space $\mathcal{X}$, Maximum budget $N_{\operatorname{max}}$, Response function $f$, Kernel function $\mathbf{K}_\theta$, Hyper-parameters $\boldsymbol{\theta}$, Design sample size $n$, and learning cycle $N_l$}
\KwOut{$\mathbf{x}^*$ and learned model $\mathcal{M}$}

\nonl \rule{0.08\textwidth}{1pt} \textbf{Dimension Reduction Phase} \rule{0.08\textwidth}{1pt}

        Sample $s \le N_{\operatorname{max}}$ random configurations from $\mathcal{X}$ within the bounds $X_i \in [\underline{X}_i, \Bar{X}_i]$ to form the initial design sample set $\mathcal{D}_{S}=\{\mathbf{x}_1, \dots ,\mathbf{x}_s\}$
        
         Obtain \emph{performance measurements} of the initial design in the source environment, $y_i \gets f_{S}(\mathbf{x}_i)+\epsilon_i,\forall\mathbf{x}_i\in\mathcal{D}_{S}$

        $\mathcal{G} \gets$ Learn a causal model on $\mathcal{D}_S$ using Algorithm \ref{alg:scm-learn}.
        
        Estimate the average causal effects of the configuration options by intervening on $X_i$: $\operatorname{CE}_{X_i} \gets 1/N \sum_{j=1}^{N} \mathbb{E}\left[Y_i \mid \text{do}\left(X_i=x_j\right)\right] - \mathbb{E}\left[Y_i \mid \text{do}\left(X_i=a\right)\right]$, where $a$ is the default value of option $X_i$.      
                
        Reduce the search space by selecting the top $K$ options with the largest causal effect:  $\Tilde{\mathcal{X}} \subset \mathcal{X}$

        \nonl \rule{0.06\textwidth}{1pt} \textbf{Configuration Optimization Phase} \rule{0.06\textwidth}{1pt}
                
         Choose an initial sparse design (Sobol sequences) in $\Tilde{\mathcal{X}}$ to find an initial design samples $\mathcal{D}_{T}=\{\mathbf{x}_1, \dots ,\mathbf{x}_n\}$
        
         Obtain \emph{performance measurements} of the initial design in the target environment, $y_i \gets f_{T}(\mathbf{x}_i)+\epsilon_i,\forall\mathbf{x}_i\in\mathcal{D}_{T}$
        
         $\mathbb{S}_{1:n} \gets \{(\mathbf{x}_{i},y_{i})\}_{i=1}^n; t \gets n+1$ 
        
         $\mathcal{M}(\mathbf{x}|\mathbb{S}_{1:n},\boldsymbol{\theta})\gets$ Fit a $\mathcal{GP}$ model to the design 
          
        \While{$t \le N_{\operatorname{max}}$}{
             {\eIf{$(t \mod N_l=0)$}{ 
             
             $\boldsymbol{\theta} 
             \gets$ \emph{Learn} the kernel hyper-parameters by maximizing the likelihood} 
            {
             Find \emph{next configuration} $\mathbf{x}_t$ by optimizing the selection criteria over the estimated response surface given the data, $\mathbf{x}_t \gets \argmax_\mathbf{x} u(\mathbf{x}| \mathcal{M},\mathbb{S}_{1:t-1})$
            
             Obtain performance for the \emph{new configuration} $\mathbf{x}_t$, $y_t \gets f_{T}(\mathbf{x}_t)+\epsilon_t$
            
              Add the newly measured configuration to the measurement set: $\mathbb{S}_{1:t}=\{\mathbb{S}_{1:t-1},(\mathbf{x}_{t},y_{t})\}$
            
             \textit{Re-fit} a new GP model $\mathcal{M}(\mathbf{x}|\mathbb{S}_{1:t},\boldsymbol{\theta})$ 
            
             $t \gets t+1$
             }
    }
    }
         $(\mathbf{x}^*,y^*)=\min \mathbb{S}_{1:N_{\operatorname{max}}}$
\end{algorithm}

A PAG is composed of directed, undirected, and partially directed edges. The partially directed edges must be
fully resolved to discover the true causal relationships. We employ the information-theoretic \emph{LatentSearch} algorithm proposed by Kocaoglu~\cite{Kocaoglu2020} to orient partially directed edges in PAG through entropic causal discovery~(line 3). For each partially directed edge, we follow two steps: (i) establish if we can generate a latent variable (with low entropy) to serve as a common cause between two vertices; (ii) if such a latent variable does not exist, then pick the direction which has the lowest entropy. For the first step, we assess whether there could be an unmeasured confounder (say $Z$) that lies between two partially oriented nodes (say $X$ and $Y$). \emph{LatentSearch} outputs a joint distribution $q(X, Y, Z)$ that can be used to compute the entropy $H(Z)$ of the unmeasured confounder $Z$. Following the Kocaoglu guidelines, we set an entropy threshold $\theta_r=0.8 \times min\left\{H(X), H(Y)\right\}$. If the entropy $H(Z)$ of the unmeasured confounder falls \textit{below} this threshold, then we declare that there is a simple unmeasured confounder $Z$ (with a low enough entropy) to serve as a common cause between $X$ and $Y$ and accordingly replace the partial edge with a bidirected (\hspace{-1.5mm}\edgetwo) edge. 
When there is no latent variable with sufficiently low entropy, there are two possibilities: (i) the variable $X$ causes $Y$; then there is an arbitrary function $f(\cdot)$ such that $Y=f(X,E)$, where $E$ is an exogenous variable (independent of $X$) that accounts for system noise; or (ii) the variable $Y$ causes $X$; then there is an arbitrary function $g(\cdot)$ such that $X=g(Y,\tilde{E})$, where $\tilde{E}$ is an exogenous variable (independent of $Y$) that accounts for noise in the system. The distribution of $E$ and $\tilde{E}$ can be inferred from the data. With these distributions, we measure the entropies $H(E)$ and $H(\tilde{E})$. If $H(E) < H(\tilde{E})$, then it is simpler to explain $X$ causes $Y$ (that is, the entropy is lower when $Y=f(X,E)$) and we choose $X$\edgeone $Y$. Otherwise, we choose $Y$\edgeone$X$. 

The final causal model is an acyclic-directed mixed graph~(ADMG). When interpreting a causal model, we view the nodes as variables and the arrows as the assumed direction of causality, whereas the absence of an arrow shows the absence of direct causal influence between variables. To quantify the influence of a configuration option on a performance objective, we need to locate the causal paths. A causal path~$P_{X \curly Y}$ is a directed path that originates from a configuration option $X$ to a subsequent non-functional property $\mathcal{S}$~ (e.g. planner failed) and ends at a performance objective $Y$. For example, $X \ \edgeone \ \mathcal{S} \ \edgeone \ Y$ denotes $X$ causes $Y$ through a subsequent node $\mathcal{S}$ on the path. We discover $P_{X \curly Y}$ by backtracking the nodes corresponding to each of the performance objectives until we reach a node without a parent. We then measure the \emph{average causal effect}~(ACE), by measuring the causal effects of the configuration options on the performance metrics and taking the average over the causal paths. We then rank the configuration options according to their ACE: $\{(X_i, \operatorname{CE}_{X_i})\}_{i=1}^d$, where $\operatorname{CE}_{X_i} \geq \operatorname{CE}_{X_{i+1}}$ for all $i < d$. Finally, we select a subset of configuration options with the highest ACE: $\{X_i \mid (X_i, \operatorname{CE}_{X_i}), 1 \leq i \leq K\}$, $K \leq d$, and reduce the search space to $\Tilde{\mathcal{X}} \subset \mathcal{X}$~(Algorithm~\ref{alg:cure-alg}: lines 4-5).

\begin{algorithm}[t]
\DontPrintSemicolon
\SetKwFor{ForEach}{for each}{do}{end}
\SetKwRepeat{Do}{do}{while}
\caption{Causal Model Learning}\label{alg:scm-learn}
\KwIn{Design samples $\mathcal{D}=\{\mathbf{x}_1, \dots ,\mathbf{x}_m\}$ from $\mathcal{X}$ with outcomes $y_i = f(\mathbf{x}_i)+\epsilon_i,\forall\mathbf{x}_i\in\mathcal{D}$}
\KwOut{Acyclic-directed mixed graph $G_{\operatorname{ADMG}}$}

        Initialize a fully connected undirected graph $G$

        Apply Fisher's z test to remove the edges between independent variables and then orient the edges to get $G_{\operatorname{PAG}}$.

        \ForEach{partial edge in $G_{\operatorname{PAG}}$}{
            Resolve the partial edge using the LatentSearch algorithm~\cite{Kocaoglu2020}. 
        }

        The resolved graph composed of directed and bi-directed edges: $G_{\operatorname{ADMG}}$\;
\end{algorithm}

\subsection{Phase II: Performance optimization through black-box optimization with limited budget}
\begin{figure}[t]
	\begin{center}
		\includegraphics[width=6cm]{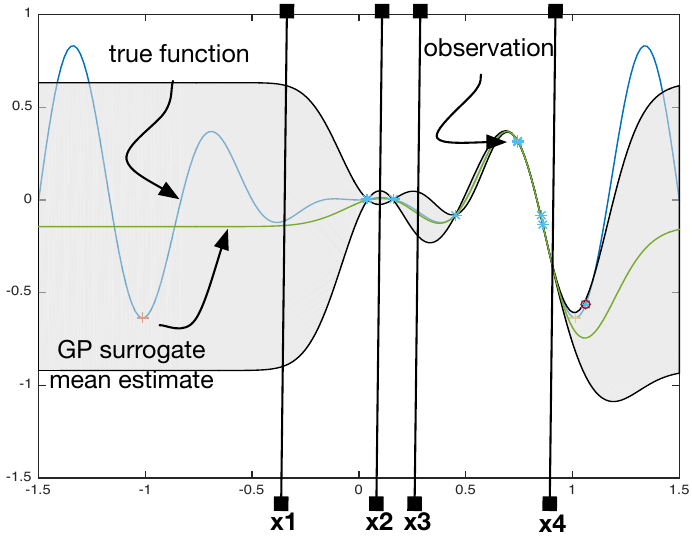}
		\caption{An example of 1D GP model: GPs provide mean estimates and uncertainty in estimations, \emph{i.e.}, variance.} 
		\label{fig:example-gp}
	\end{center}
 \vspace{-1em}
\end{figure}

In the configuration optimization phase (lines 6-18), we search for Pareto optimal configurations using an active learning approach that operates in the reduced search space in the target environment. Here the target environment is typically the target robotic platform that we want to optimize. The assumption is that any intervention in the target environment is costly and that we typically assume a small sampling budget. In some situations, we could assume that the cost of measuring configurations varies. For example, if the likelihood of violating safety confidence is high for a specific configuration, we could assign a higher cost to that configuration because it may damage the robot. We leave this assumption for future work. Specifically, we start by bootstrapping optimization by randomly sampling the reduced configuration space to produce an initial design $\mathcal{D} = \{\mathbf{x}_1, \dots ,\mathbf{x}_n\}$, where $\mathbf{x}_i\in \Tilde{\mathcal{X}}$. After obtaining the measurements regarding the initial design, \ourapproach then fits a GP model to the design points $\mathcal{D}$ to form our belief about the underlying response function. The while loop in Algorithm \ref{alg:cure-alg} iteratively updates the belief until the budget runs out: As we accumulate the data $\mathbb{S}_{1:t}=\{(\mathbf{x}_{i},y_{i})\}_{i=1}^t$, where $y_i=f_{T}(\mathbf{x}_i)+\epsilon_i$ with $\epsilon \sim \mathcal{N}(0,\sigma^2)$, a prior distribution $\Pr(f_{T})$ and the likelihood function $\Pr(\mathbb{S}_{1:t}|f_{T})$ form the posterior distribution: $\Pr(f_{T}|\mathbb{S}_{1:t}) \propto \Pr(\mathbb{S}_{1:t}|f_{T})\Pr(f_{T})$. We describe the steps of Phase~II as follows:

\paragraph{Bayesian optimization with GP} Bayesian optimization is a sequential design strategy that allows us to perform global optimization of black-box functions \cite{shahriari2015taking}. The main idea of this method is to treat the black-box objective function $f(\bs x)$ as a random variable with a given prior distribution and then optimize the posterior distribution of $f(\bs x)$, given experimental data. In this work, we use GPs to model this black-box objective function at each point $\bs x\in \mathbb{X}$. That is, let $\mathbb{S}_{1:t}$ be the experimental data collected in the first $t$ iterations, and let $\mathbf{x}_{t+1}$ be a candidate configuration that we can select to run the next experiment. Then the probability that this new experiment could find an optimal configuration using the posterior distribution will be assessed:
\begin{gather*}
    \Pr(f_{t+1}|\mathbb{S}_{1:t},\mathbf{x}_{t+1})\sim \mathcal{N}(\mu_{t}(\mathbf{x_{t+1}}),\sigma_t^2(\mathbf{x}_{t+1})),
\end{gather*}
where $\mu_{t}(\mathbf{x_{t+1}})$ and $\sigma_t^2(\mathbf{x}_{t+1})$ are suitable estimators of the mean and standard deviation of a normal distribution used to model this posterior. The main motivation behind the choice of GPs as prior here is that it offers a framework in which reasoning can be based not only on mean estimates, but also on variance, providing more informative decision making. The other reason is that all the computations in this framework are based on a solid foundation of \emph{linear algebra}. Fig~\ref{fig:example-gp} illustrates Bayesian optimization based on GP using a one-dimensional response surface. The blue curve represents the unknown true posterior distribution, while the mean is shown in green, and the confidence interval 95\% is shaded. Stars indicate measurements carried out in the past and recorded in $\mathbb{S}_{1:t}$ (i.e., observations). The configuration corresponding to $\mathbf{x}_1$ has a large confidence interval due to the lack of observations in its neighborhood. On the contrary, $\mathbf{x}_4$ has a narrow confidence since neighboring configurations have been experimented with. The confidence interval in the neighborhood of $\mathbf{x}_2$ and $\mathbf{x}_3$ is not large, and correctly our approach does not decide to explore these zones. The next configuration $\mathbf{x}_{t+1}$, indicated by a small circle on the right side of $\mathbf{x}_4$, is selected based on a criterion that will be defined later. A GP is a distribution over functions, specified by its mean and covariance:
\begin{equation} \label{eq:GP draw}
    y=f(\mathbf{x}) \sim \mathcal{GP} (\mu(\mathbf{x}), k(\mathbf{x}, \mathbf{x}')),
\end{equation}
where $k(\mathbf{x}, \mathbf{x}')$ defines the distance between $\mathbf{x}$ and $\mathbf{x}'$. 
Assume $\mathbb{S}_{1:t}=\{(\mathbf{x}_{1:t},y_{1:t}) | y_i:=f(\mathbf{x}_i)\}$ to be the collection of observations $t$. 
The function values are drawn from a multi-variate Gaussian distribution $\mathcal{N}(\boldsymbol{\mu},\mathbf{K})$, where $\mathbf{\mu}:=\mu(\mathbf{x}_{1:t})$,
\begin{equation} \label{eq:covariance}
    \mathbf{K}:=
    \begin{bmatrix}
    k(\mathbf{x}_1,\mathbf{x}_1)  &  \dots & k(\mathbf{x}_1,\mathbf{x}_t)   \\
    \vdots  & \ddots &  \vdots \\
    k(\mathbf{x}_t,\mathbf{x}_1)  &  \dots & k(\mathbf{x}_t,\mathbf{x}_t) 
    \end{bmatrix}
\end{equation}
In the while loop in \ourapproach, given the observations we accumulated so far, we intend to fit a new GP model:
\begin{equation} \label{eq:joint-gp}
    \begin{bmatrix}
    \mathbf{f}_{1:t} \\
    f_{t+1}
    \end{bmatrix}
    \sim \mathcal{N}(\boldsymbol{\mu},
    \begin{bmatrix}
    \mathbf{K}+\sigma^2\mathbf{I} & \mathbf{k} \\
    \mathbf{k}^\intercal & k(\mathbf{x}_{t+1},\mathbf{x}_{t+1}) 
    \end{bmatrix}),
\end{equation}
where $\mathbf{k}(\mathbf{x})^\intercal=[k(\mathbf{x},\mathbf{x}_1) \quad k(\mathbf{x},\mathbf{x}_2) \quad \dots \quad k(\mathbf{x},\mathbf{x}_t)]$ and $\mathbf{I}$ is identity matrix.
Given Eq.~\eqref{eq:joint-gp}, the new GP model can be drawn from this new Gaussian distribution:
\begin{equation} \label{eq:gp-surrogate}
    \begin{aligned}
    \Pr(f_{t+1}|\mathbb{S}_{1:t},\mathbf{x}_{t+1})=\mathcal{N}(\mu_{t}(\mathbf{x_{t+1}}),\sigma_t^2(\mathbf{x}_{t+1})),
    \end{aligned}
\end{equation}
where
\begin{align}
    \label{eq:gp-surrogate-mean-sigma}
    \mu_{t}(\mathbf{x})&=\mu(\mathbf{x})+\mathbf{k}(\mathbf{x})^\intercal (\mathbf{K}+\sigma^2\mathbf{I})^{-1} (\mathbf{y}-\boldsymbol{\mu}) \\
    \sigma_t^2(\mathbf{x})&=k(\mathbf{x},\mathbf{x})+\sigma^2\mathbf{I} - \mathbf{k}(\mathbf{x})^\intercal (\mathbf{K}+\sigma^2\mathbf{I})^{-1} \mathbf{k}(\mathbf{x})
\end{align}
These posterior functions are used to select the next point $\mathbf{x}_{t+1}$. 

\paragraph{Configuration selection criteria}
The selection criteria is defined as $u:\mathbb{X}\rightarrow\mathbb{R}$ that selects $\mathbf{x}_{t+1}\in\mathbb{X}$, should $f(\cdot)$ be evaluated next (\emph{step 7}):
\begin{equation} \label{eq:next-point}
\mathbf{x}_{t+1}=\argmax_{\mathbf{x}\in \mathbb{X}} u(\mathbf{x}|\mathcal{M},\mathbb{S}_{1:t})
\end{equation}

\noindent Although there are several different criteria in the literature for multiobjective optimization~\cite{knowles2006parego, hernandez2016predictive, ponweiser2008multiobjective}, \ourapproach utilizes Expected Hypervolume Improvement~(EHVI). EHVI has demonstrated its strength in balancing exploration and exploitation, and in producing Pareto fronts with excellent coverage and faster optimization~\cite{daulton2020differentiable}. EHVI operates by assessing the expected improvement of a given point in the solution space in terms of the hypervolume measure---a widely accepted metric for comparing the quality of solutions in multi-objective optimization. EHVI is particularly useful in robotic applications, where the solution landscape can be highly complex and multi-dimensional. 
\begin{figure}[t]
	\begin{center}
		\includegraphics[width=8.5cm]{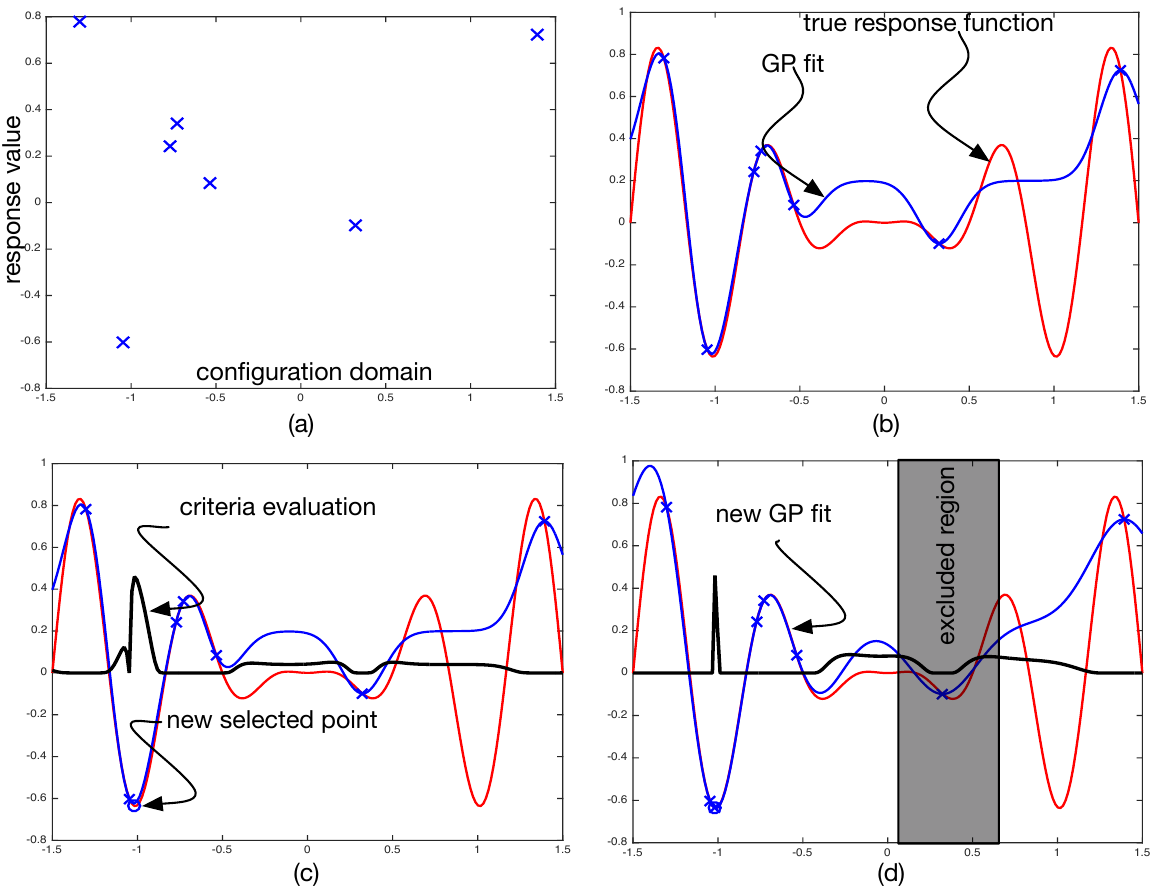}
		\caption{Illustration of configuration parameter optimization: (a) initial observations; (b) a GP model fit; (c) choosing the next point; (d) refitting a new GP model.}
		\label{fig:illustration-gp}
	\end{center}
  \vspace{-1em}
\end{figure}
The steps of Algorithm~\ref{alg:cure-alg} are illustrated in Fig~\ref{fig:illustration-gp}. First, an initial design based on random sampling is produced~(Fig~\ref{fig:illustration-gp}a). Second, a GP model is fitted to the initial design~(Fig~\ref{fig:illustration-gp}b). The model is then used to calculate the selection criteria~(Fig~\ref{fig:illustration-gp}c). Finally, the configuration that maximizes the selection criteria is used to run the next experiment and provide data to reconstruct a more accurate model~(Fig~\ref{fig:illustration-gp}d).

\paragraph{Model fitting}
Here, we provide some practical considerations to make GPs applicable for configuration optimization. In \ourapproach, as shown in Algorithm \ref{alg:cure-alg}, the covariance function $k:\mathcal{X}\times\mathcal{X}\rightarrow\mathbb{R}$ dictates the structure of the response function that we fit to the observed data. For integer variables, we implemented the Mat\'ern kernel~\cite{gpml}. The main reason behind this choice is that along each dimension of the configuration response functions, a different level of smoothness can be observed. Mat\'ern kernels incorporate a smoothness parameter $\nu>0$ that allows greater flexibility in modeling such functions. The following is a variation of the Mat\'ern kernel for $\nu=1/2$:
\begin{align}
\label{eq:matern}
k_{\nu=1/2}(\mathbf{x}_i,\mathbf{x}_j)&= \theta_0^2\exp(-r),
\end{align}
where $r^2(\mathbf{x}_i,\mathbf{x}_j)=(\mathbf{x}_i-\mathbf{x}_j)^\intercal \mathbf{\Lambda} (\mathbf{x}_i-\mathbf{x}_j)$ for some positive semidefinite matrix $\mathbf{\Lambda}$. For categorical variables, we implement the following \cite{hutter2009automated}:
\begin{align}
    \label{eq:categorical}
    k_{\theta}(\mathbf{x}_i,\mathbf{x}_j)=\exp(\Sigma_{\ell=1}^{d}(-\theta_\ell\delta(\mathbf{x}_i\neq\mathbf{x}_j))),
\end{align}
where $d$ is the number of dimensions (\emph{i.e.}, the number of configuration parameters), $\theta_\ell$ adjust the scales along the function dimensions and $\delta$ is a function gives the distance between two categorical variables using Kronecker delta \cite{hutter2009automated,shahriaritaking}.
{\sf TL4CO} uses different scales $\{\theta_\ell, \ell=1\dots d\}$ on different dimensions as suggested in \cite{gpml,shahriaritaking}, this technique is called Automatic Relevance Determination (ARD).
After learning the hyper-parameters (\emph{step 6}), if the $\ell$-th dimension turns out to be irrelevant, then $\theta_\ell$ will be a small value, and therefore will be discarded. This is particularly helpful in high-dimensional spaces where it is difficult to find the optimal configuration. Although the kernel controls the structure of the estimated function, the prior mean $\mu(\mathbf{x}):\mathbb{X}\rightarrow\mathbb{R}$ provides a possible offset for our estimate. By default, this function is set to a constant $\mu(\mathbf{x}):=\mu$, which is inferred from observations~\cite{shahriaritaking}. However, the prior mean function is a way of incorporating expert knowledge, and if it is available, then we can use this knowledge. Fortunately, we have collected extensive experimental measurements and based on our datasets, we observed that, for robotic systems, there is typically a significant distance between the minimum and the maximum of each function~(Fig.~\ref{fig:obs_surf}, \ref{fig:obs_surf_lowace}). Therefore, a linear mean function $\mu(\mathbf{x}):=\mathbf{a}\mathbf{x}+b$ allows for more flexible structures and provides a better fit for the data than a constant mean. We only need to learn the slope for each dimension and the offset (denoted $\mu_\ell=(\mathbf{a},b)$). Due to the heavy learning computation~(\emph{step 12} in Algorithm \ref{alg:cure-alg}), this process is computed only for every $N_l^{th}$ iteration. 
To learn the hyperparameters of the kernel and also the prior mean functions, we maximize the marginal likelihood~\cite{shahriaritaking} of the observations $\mathbb{S}_{1:t}$. To do that, we train the GP model \eqref{eq:gp-surrogate-mean-sigma} with $\mathbb{S}_{1:t}$.
We optimize the marginal likelihood using multi-started quasi-Newton hill climbers~\cite{gpml}. For this purpose, we used the Ax + BoTorch library. Using the kernel defined in \eqref{eq:categorical}, we learn $\boldsymbol{\theta}:=(\theta_{0:d},\mu_{0:d},\sigma^2)$ which comprises the hyperparameters of the kernel and the mean functions. The learning is performed iteratively, resulting in a sequence of $\boldsymbol{\theta}_i$ for $i=1\dots\lfloor \frac{N_{\operatorname{max}}}{N_\ell} \rfloor$. 
\begin{figure}[!t]  
\vspace{-0.5em}
  \centering
  \subfloat[Husky in Gazebo]{\includegraphics[width=.52\columnwidth]{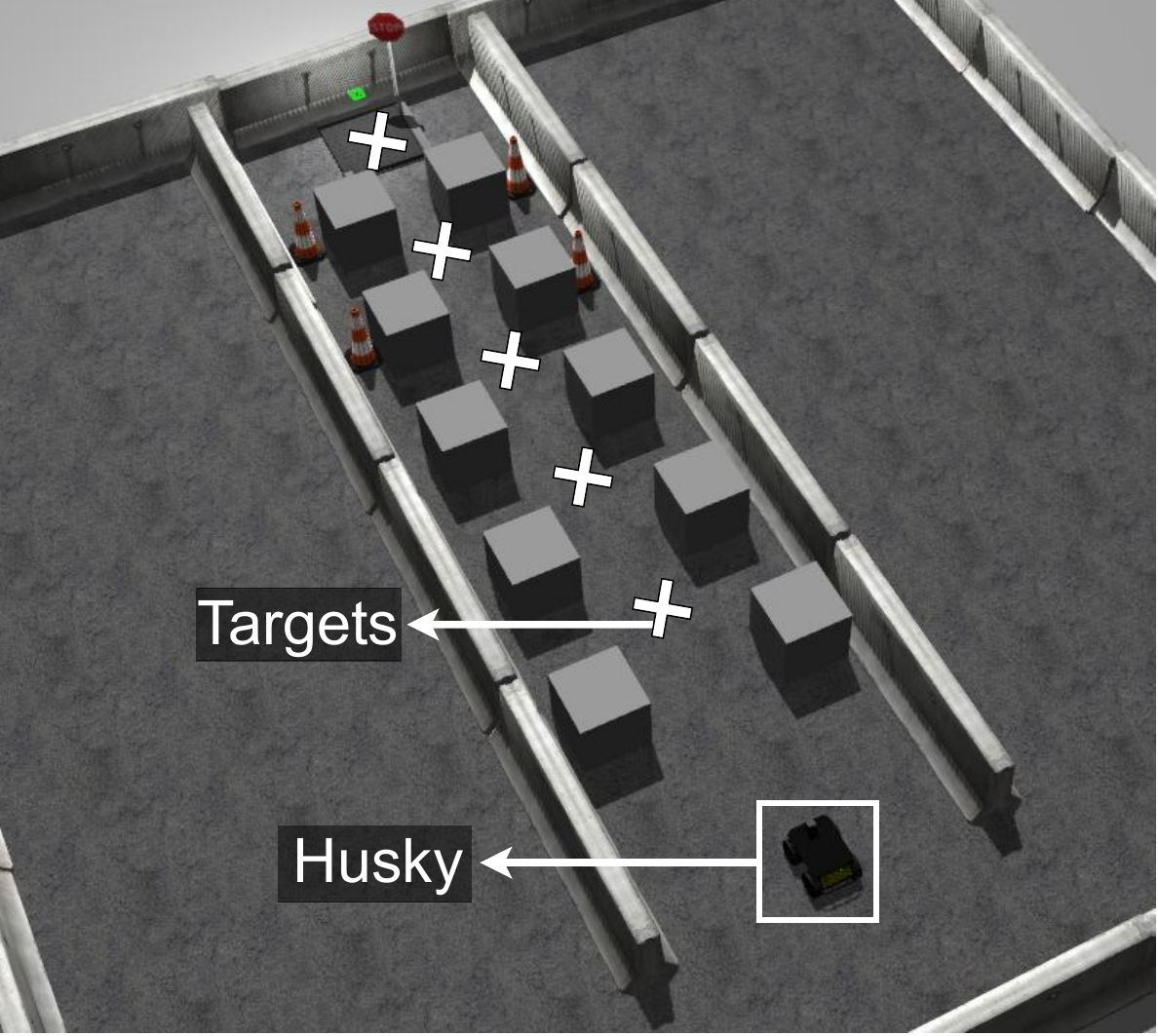}\label{fig:husky_env}}
  \hfill
  \subfloat[Turtlebot 3 in Gazebo]{\includegraphics[width=0.462\columnwidth,]{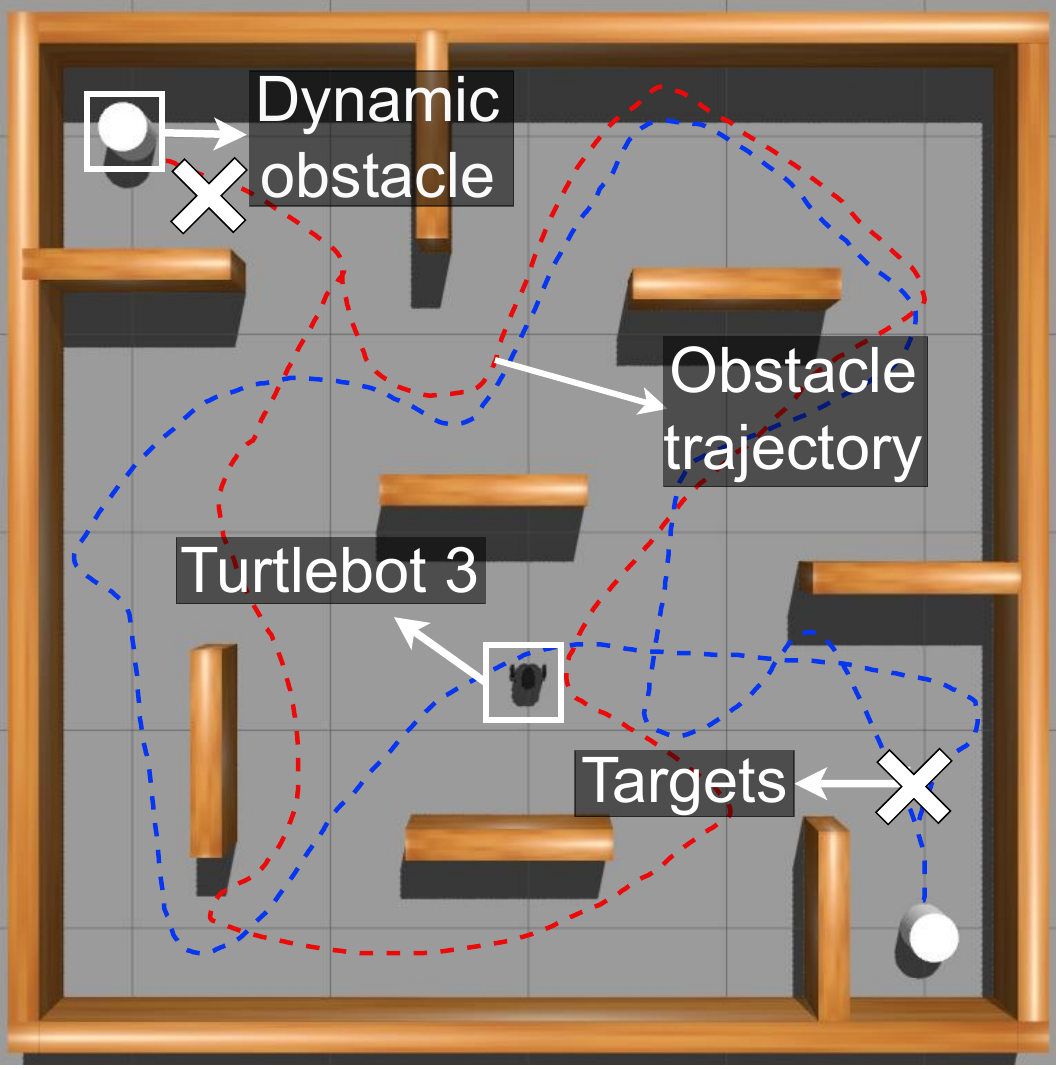}\label{fig:tbot_env}}
  \caption{\small{Simulated environments for \textit{Husky} and \textit{Turtlebot 3}. The dashed lines in (b) show the trajectory of the dynamic obstacles.}
  \label{fig:env_and_penalty}}
  \vspace{-1.5em}
\end{figure}

\section{Experiments and Results} \label{sec:exp}
To evaluate this work, we answer the following research questions~(RQs)
\begin{itemize}
    \item RQ1 (Effectiveness): How effective is \ourapproach in (i)~ensuring optimal performance; (ii)~utilizing the budget; and (iii)~respecting the safety constraints compared to the baselines?
    \item RQ2 (Transferability): How does the effectiveness of \ourapproach change when the severity of deployment changes varies~(e.g., environment and platform change)?
\end{itemize}
{We answered these questions in a robot navigation task, using \textit{Husky} and \textit{Turtlebot 3} platforms. Additionally, to illustrate adaptability of \ourapproach to different tasks, we also demonstrate RQ1 on a robot manipulation task, using the \textit{Franka Emika Panda} platform in \textit{Gazebo}.

\begin{figure}[!t]
    \vspace{-1.4em}
    \begin{center}
    \centerline{\includegraphics[width=1\columnwidth, angle=-90]{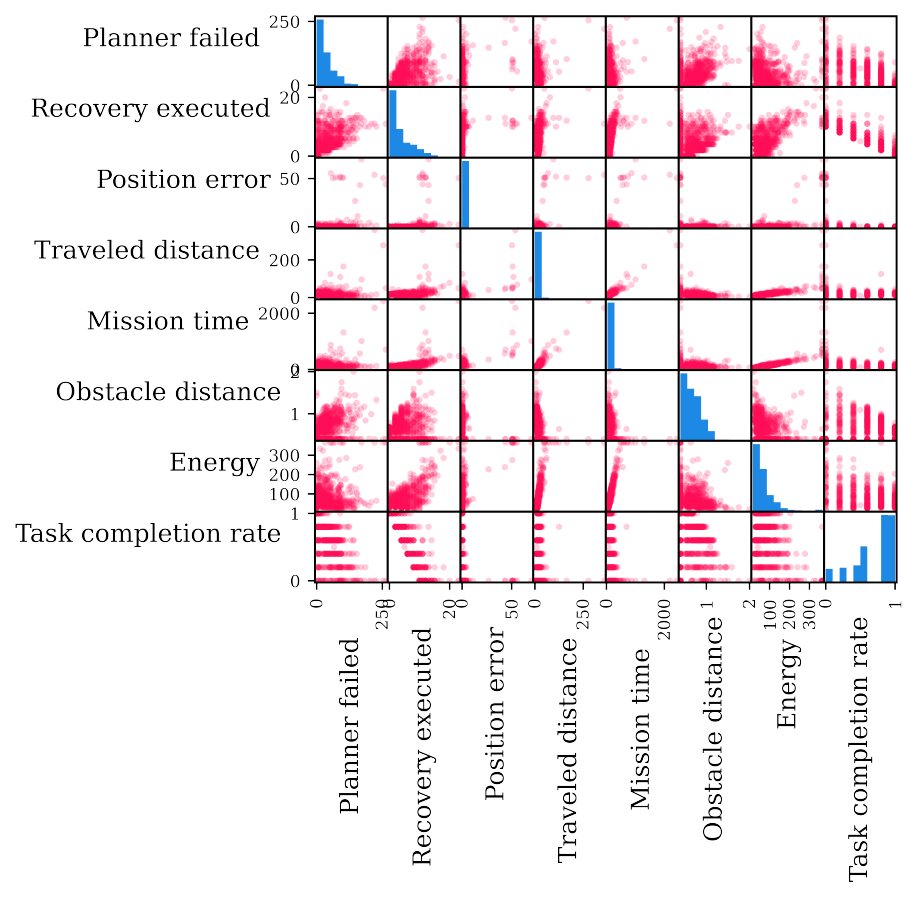}}
    \caption{\small{Correlation between different performance objectives derived from observational data.}}
    \label{fig:obs_data}
    \end{center} 
    \vspace{-2.4em}
\end{figure}

\subsection{Experimental setup} \label{sec:exp_rq1}
\paragraph{Robot navigation} We simulate \textit{Husky} and \textit{Turtlebot 3} in \textit{Gazebo} to collect the observational data  by measuring the performance metrics~(e.g., planner failed) and performance objectives~(e.g., energy consumption) under different configuration settings to train the causal model. Note that we use simulator data to evaluate the transferability of the causal model to physical robots, but \ourapproach also works with data from physical robots. We deploy the robot in a controlled indoor environment and direct the robot to navigate autonomously to the target locations~(Fig.~\ref{fig:husky_env}). The robot was expected to encounter obstacles and narrow passageways, where the locations of the obstacles were unknown prior to deployment. The mission was considered successful if the robot reached each of the target locations. We fixed the goal tolerance parameters~({\texTTT{xy\_goal\_tolerance=0.2}}, and {\texTTT{yaw\_goal\_tolerance=0.1}}) to determine whether a target was reached. We defined the following properties for the ROS Navigation Stack~\cite{navcore}: (i)~\textit{Task completion rate}: $\mathcal{T}_{cr}=(\sum\mathrm{Tasks_{completed}})/(\sum \mathrm{Tasks})$; (ii)~\textit{Traveled distance}: Distance traveled from start to destination; (iii)~\textit{Mission time}: Total time to complete a mission (iv)~\textit{Position error}: Euclidean distance between the actual target position and the position reached by the robot, $E_{dist}=\sqrt{\sum^n_{i=1}(t_i - r_i)^2}$, where $t$ and $r$ denote the target and position reached by the robot, respectively; (v)~\textit{Recovery executed}: Number of {\texTTT{rotate recovery}} and {\texTTT{clear costmap recovery}} executed per mission; and (vi)~\textit{Planner failed}: Number of times the planner failed to produce a path during a mission.

\paragraph{Robot manipulation}
We simulate the \textit{Franka Emika Panda} in \textit{Gazebo} and perform a pick-and-place task using the \textit{Moveit}~\cite{moveit} motion planning framework. To learn a causal model, we measure the following performance objectives under different configuration settings: (i)~\textit{Average trajectory jerk:} Rate of change of acceleration, averaged across all joints and time steps, we define $\text{average jerk} = \frac{1}{N} \sum_{t=1}^{N} \sqrt{\sum_{j=1}^{7} \left( \frac{a_j(t) - a_j(t-1)}{\Delta t} \right)^2 }$, where $N$ is the total number of time steps, $a_j(t)$ is the acceleration of joint $j$ at time $t$, and $\Delta t$ is the time interval between consecutive time steps; and (ii)~\textit{Task execution time}: The total execution time from picking up an object to placing.

\begin{figure}[!t]
  \vspace{-0.7em} 
  \centering
  \subfloat[Pareto front]{\includegraphics[width=.5\columnwidth]{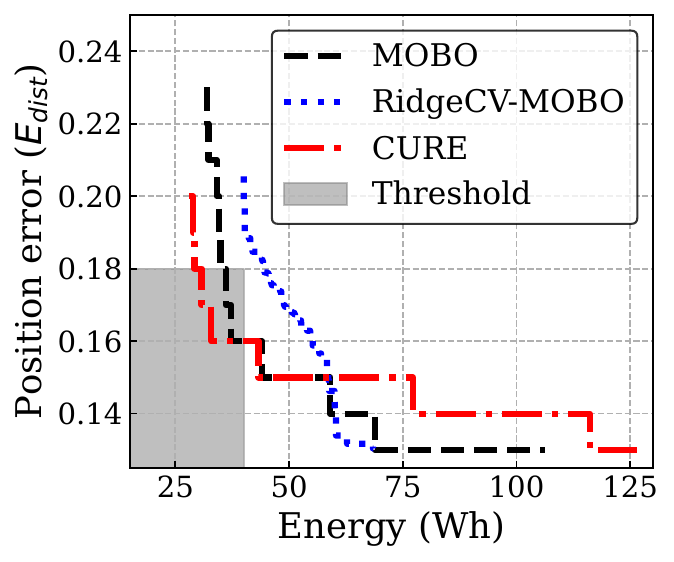}\label{fig:rq1_pareto}}
  \hfill
  \subfloat[Hypervolume]{\includegraphics[width=0.5\columnwidth,]{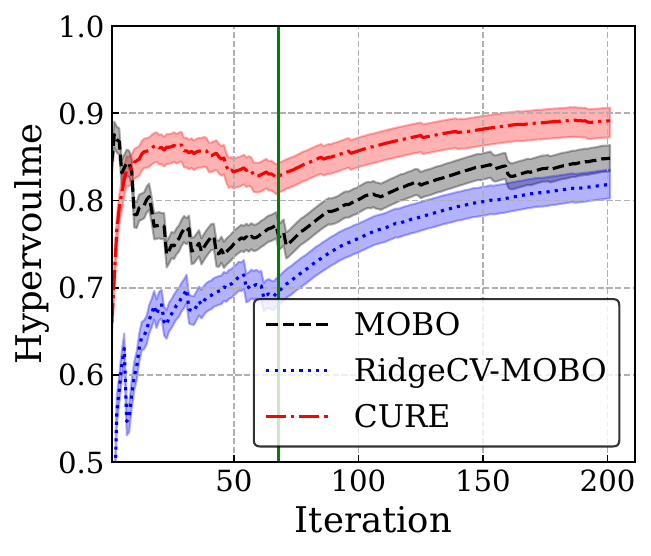}\label{fig:rq1_hypervolume}}
  \hfill
  \subfloat[Efficiency]{\includegraphics[width=.5\columnwidth,]{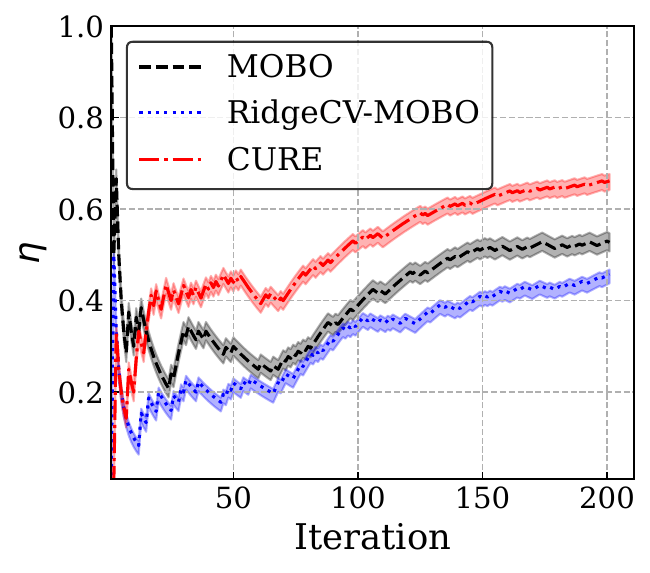}\label{fig:rq1_efficiency}} 
  \hfill
  \subfloat[Safety]{\includegraphics[width=0.5\columnwidth,]{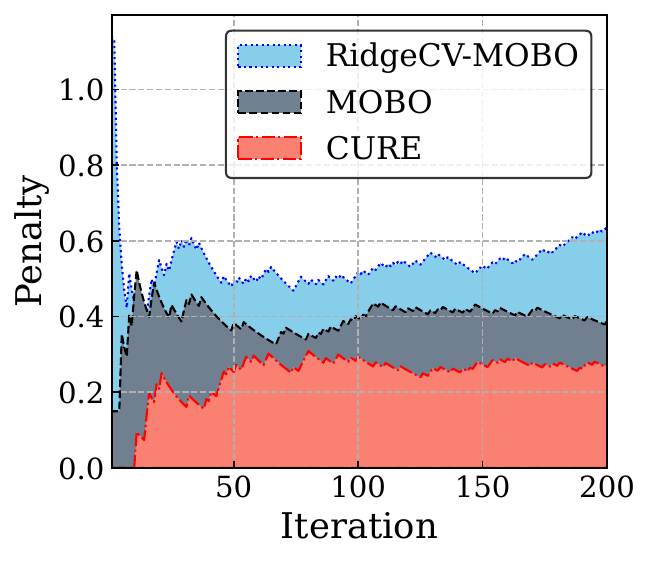}\label{fig:rq1_safety}}  
  \caption{\small{Effectiveness of \ourapproach and baseline methods for the navigation task: (a)~Pareto front; (b)~Hypervolume, (c)~efficiency; and (d)~safety penalty response obtained by \ourapproach and other approaches for \textit{Husky} in simulation. The vertical green line in (b) shows the number of initial trails before fitting the GP model.}}
  \label{fig:rq1_performance}
  \vspace{-0.6em} 
\end{figure}

\subsection{Evaluation}
To learn a causal model from the source~(a low-cost environment), we generated the values for the configurable parameters using random sampling and recorded the performance metrics~(the intermediate layer of the causal model that maps the influence of the configuration options to the performance objective) for different values of the configurable parameters. We use a budget of $200$ iterations for each method. When running each method for the same budget, we compare the Pareto front~($PF$) and Pareto hypervolume~($HV$). The Pareto front is the set of objective vectors corresponding to all Pareto-optimal configurations in the configuration space $\mathcal{X}$. The Pareto hypervolume is commonly used to measure the quality of an estimated Pareto front~\cite{cao2015using, zitzler1999multiobjective}. We define the Pareto front and hypervolume as follows:
\begin{equation}
    PF = \{( f_j(\mathbf{x}))_{j=1}^{m} \mid \mathbf{x} \in \mathcal{X} \text{ is Pareto-optimal}\},
\end{equation}
\begin{equation}\label{eq:hypervolume}
        HV(\mathbf{x^*}, f^{\operatorname{ref}}) = \Lambda\left( \bigcup_{\mathbf{x}_n^* \in \mathbf{x^*}} \prod_{j=1}^{m} [f_j(\mathbf{x}_n^*), f_j^{\operatorname{ref}}] \right),
\end{equation}
\noindent where $HV(\mathbf{x^*}, f^{\operatorname{ref}})$ resolves the size of the dominated space covered by a non-dominated set $\mathbf{x^*}$, $f^{\operatorname{ref}}$ refers to a user-defined reference point in the objective space, and $\Lambda(.)$ refers to the Lebesgue measure. In our experiments, we fixed the $f^{\operatorname{ref}}$ points to the maximum observed values of each objective among all the methods. 

To compare the efficiency of each method, we define an efficiency metric $\eta = (\sum_{k=1}^n \mathcal{T}_k)/(\sum_{k=1}^n k)$, where $\mathcal{T}_k$ is a binary variable taking values $0$ or $1$, denoting the success of a task during the $k^{th}$ iteration. We also compare the number of unsuccessful execution~(e.g., when the robot failed to complete a task) and the number of constraint violations~(e.g., when the robot completed the task but violated a constraint). We compared \ourapproach with the following baselines:
\begin{itemize}
    \item MOBO: We implement multiobjective Bayesian optimization~(MOBO) using \textsc{Ax}~\cite{Ax}---an optimization framework that can optimize discrete and continuous configurations.
    \item RidgeCV~\cite{ridgeCV, hoerl1970ridge}: A feature extraction method that selects the important features based on the highest absolute coefficient. We use RidgeCV to determine the important configuration options and generate a reduced search space which consists of only the important configuration options. We then perform an optimization using MOBO on the reduced search space.
\end{itemize}

\begin{figure}[!t]
  \vspace{-0.5em} 
  \centering
  \subfloat[Hypervolume]{\includegraphics[width=.5\columnwidth]{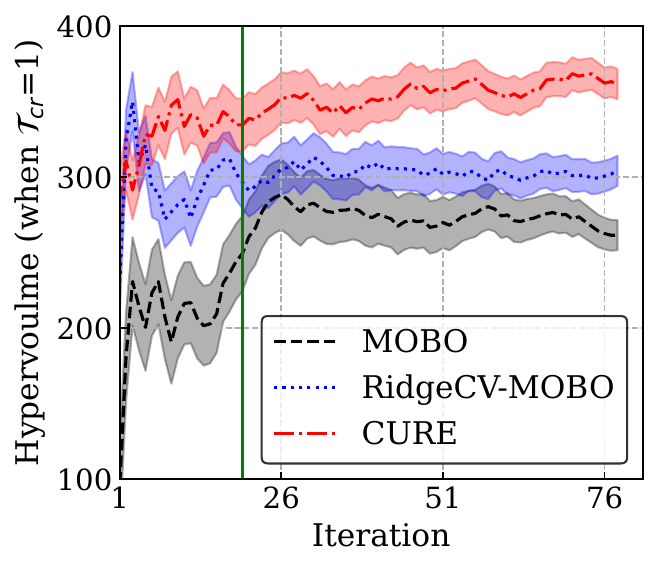}\label{fig:rq1_pareto_task2}}
  \hfill
  \subfloat[Efficiency]{\includegraphics[width=0.49\columnwidth,]{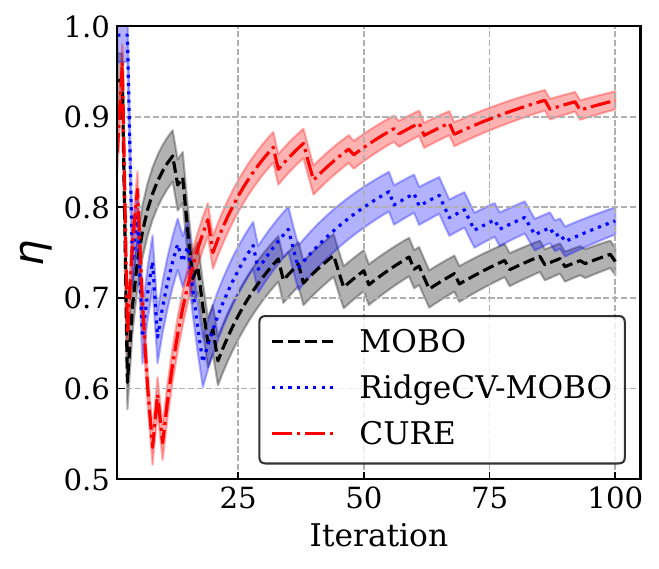}\label{fig:rq1_hypervolume_task2}} 
  \caption{\small{Effectiveness of \ourapproach and baseline methods for the manipulation task: (a)~Hypervolume; and (b)~Efficiency.}}
  \label{fig:rq1_performance_task2}
  \vspace{-0.5em}
\end{figure}

\subsection{RQ1: Effectiveness} \label{sec:rq1}
We evaluated the effectiveness of \ourapproach in finding an optimal configuration compared to the baselines. We collect observational data by running a mission $1000$ times from \textit{Husky} in simulation under different configuration settings and recorded the performance objectives. In Fig.~\ref{fig:obs_data}, the histograms of performance objectives are depicted along the diagonal line, while scatter plots illustrating pairs of performance objectives are displayed outside the diagonal. The histograms of performance objectives, namely planner failed, recovery executed, obstacle distance, and energy, have shapes similar to one half of a Gaussian distribution. Scatter plots depicting different pairs of performance objectives, such as mission time, distance traveled, and energy, exhibit positive linear relationships. We selected energy and position error as the two performance objectives given the imperative to incorporate uncorrelated objectives in the multi-objective optimization framework, underscored by their lowest correlation coefficient, ensuring the diversity of the optimization criteria. We then learn a causal model using observational data. The search space was reduced according to the estimated causal effects on performance objectives and constraints by selecting top $K$ configuration options~(e.g., $\{\mathrm{Energy}_{topK}\} \cup \{\mathrm{PoseError}_{topK}\} \cup \{\mathrm{Safety}_{topK}\}$) and performed optimization using Algorithm~\ref{alg:cure-alg}.

\begin{figure*}[!t]
  \vspace{-1.5em}
  \centering
  \subfloat[MOBO]
  {\includegraphics[width=.62\columnwidth]{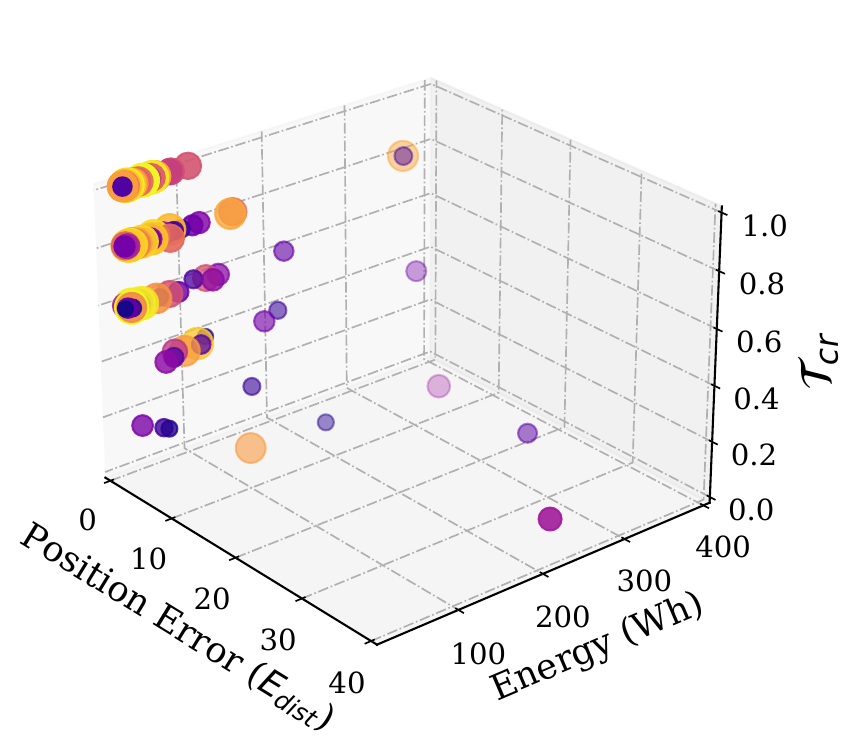}\label{fig:rq1_budgetutil_mobo}} 
  \hfill
  \subfloat[RidgeCV-MOBO]
  {\includegraphics[width=.62\columnwidth]{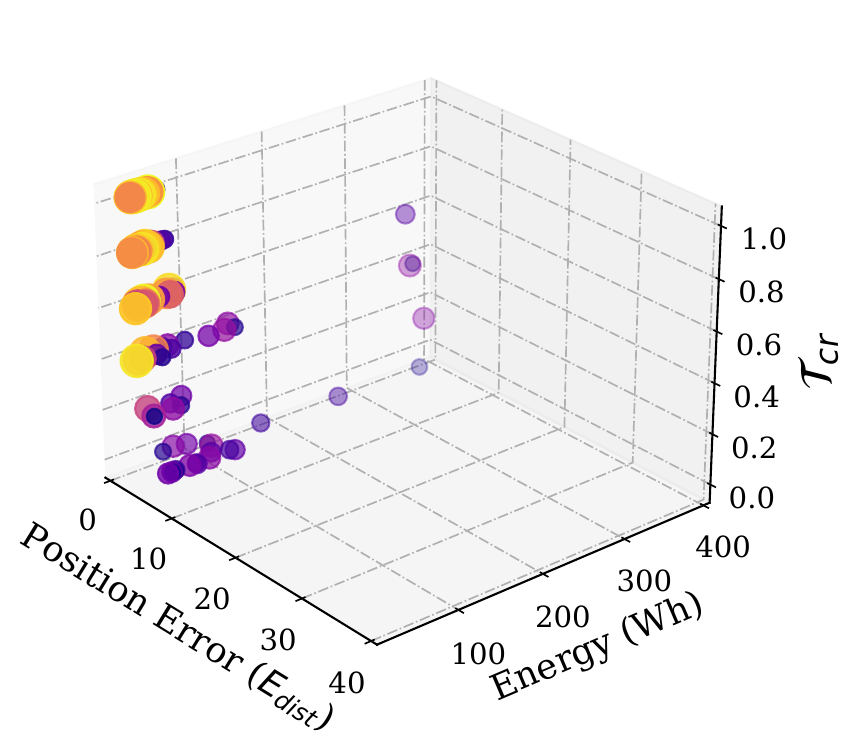}\label{fig:rq1_budgetutil_ridgecv}}
  \hfill
  \subfloat[\ourapproach]{\includegraphics[width=0.75\columnwidth,]{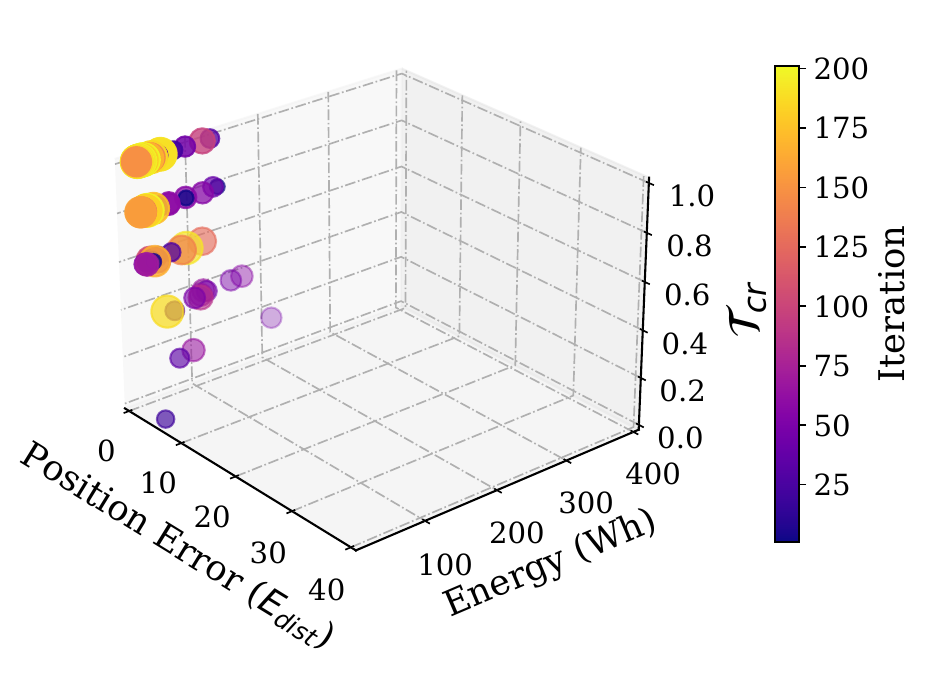}\label{fig:rq1_budgetutil_cure}}
  \caption{\small{\ourapproach demonstrates a denser surface response near the Pareto front and achieved higher $\mathcal{T}_{cr}$ in fewer iterations for the navigation task}, resulting in better budget utilization compared to baselines.}
  \label{fig:rq1_budgetutil} 
\end{figure*}

\paragraph{Setting} For the \textit{Husky} robot, we set the objective thresholds $\mathrm{Energy_{Th}} = 40 \ \mathrm{Wh}$ and $\mathrm{PoseError_{Th}} = 0.18 \ \mathrm{m}$. We compute the hypervolume using Eq.~\eqref{eq:hypervolume} by setting the $f^{\operatorname{ref}}$ points at $400$ for energy and $35$ for position error within the coordinate system. 
\begin{wrapfigure}{r}{4cm}
\includegraphics[width=4cm]{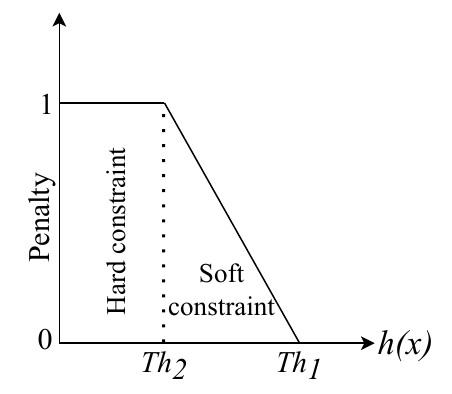}
\caption{\small{Penalty function.}}\label{fig:penalty_func}
\end{wrapfigure}
We incorporate the safety constraint~$h(\mathbf{x})$ by defining a test case, where the robot must maintain a minimum distance from obstacles to avoid collisions. We incorporate a user defined penalty function~(Fig.~\ref{fig:penalty_func}) for each instance $0 \leq \alpha h(\mathbf{x}) \leq 1$ that penalizes $\mathcal{T}_{cr}$ if $h(\mathbf{x})$ is violated. In Fig.~\ref{fig:penalty_func}, $Th_1$ is a soft constraint threshold and $Th_2$ is a hard constraint threshold. That is, we penalize $\mathcal{T}_{cr}$ gradually if $Th_1 > h(x) > Th_2$ and give the maximum penalty if $h(x) < Th_2$ to ensure safety.  In our experiments, we set $Th_1 = 0.25$ and $Th_2 = 0.18$. We defined the safety constraint: $\mathcal{T}_{cr} - \frac{1}{N} \sum_{k=0}^N\alpha_k h(\mathbf{x}) \geq \theta$, where $\theta$ is a user-defined threshold. In our experiments, we set $\theta = 0.8$. For the manipulation task, we set the $f^{\operatorname{ref}}$ points at $16$ for task execution time and $113$ for average trajectory jerk.

\paragraph{Results} 
\ourapproach performed better than MOBO and RidgeCV-MOBO in finding a Pareto front with a higher hypervolume, as shown in Fig.~\ref{fig:rq1_performance}. In our experiments, we observed a comparable Pareto front between \ourapproach and MOBO~(Fig.~\ref{fig:rq1_pareto}), which can be attributed to MOBO's exploration of an extensive search space that includes all possible configuration options. On the contrary, \ourapproach confines its exploration to a reduced search space, composing only configuration options with a greater causal effect on performance objectives. Although \ourapproach and MOBO have a similar Pareto front, \ourapproach achieved a higher hypervolume with a less amount of budget~(Fig.~\ref{fig:rq1_hypervolume}). Fig.~\ref{fig:rq1_budgetutil} illustrates the budget utilization of \ourapproach and baseline methods. \ourapproach demonstrated better budget utilization, as reflected in the increased density of purple-colored data points surrounding the Pareto front and the achievement of a higher $\mathcal{T}_{cr}$ in fewer iterations compared to the baseline methods. When comparing the penalty response given, we observed \ourapproach selected configuration options that achieved the lower penalty, as shown in Fig.~\ref{fig:rq1_safety}. Furthermore, \ourapproach outperformed the baselines in terms of efficiency, achieving a $1.3\times$ improvement over MOBO and achieved this improvement $2\times$ faster compared to MOBO~(as shown in Fig.~\ref{fig:rq1_efficiency}). RidgeCV-MOBO, however, underperformed, mainly because it was unable to identify the core configuration options influencing the performance objectives~(Fig.~\ref{fig:rq1_hypervolume}, \ref{fig:rq1_efficiency}, \ref{fig:rq1_budgetutil_ridgecv}). Moreover, \ourapproach continuously outperformed the baselines in the manipulation task~(Fig.~\ref{fig:rq1_performance_task2}). Therefore, \ourapproach is more effective in finding optimal configurations compared to the baselines.

\begin{figure*}[!t]
  \vspace{-1em}
  \centering
  \subfloat[Hypervolume]{\includegraphics[width=0.4\columnwidth,]{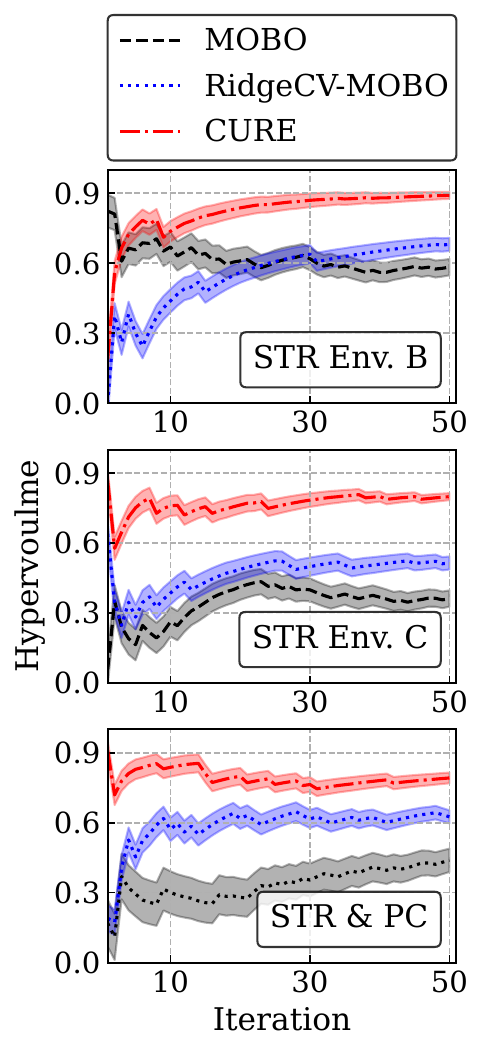}\label{fig:rq2_hypervolume}}  
  \hfill
  \subfloat[Pareto front]{\includegraphics[width=0.4\columnwidth]{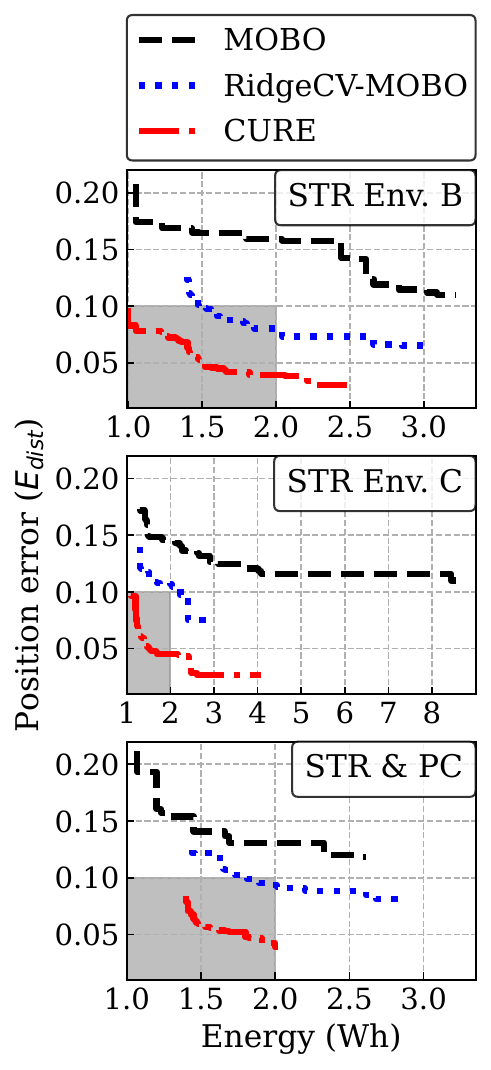}\label{fig:rq2_pareto}}
  \hfill  
  \subfloat[Efficiency]{\includegraphics[width=0.4\columnwidth]{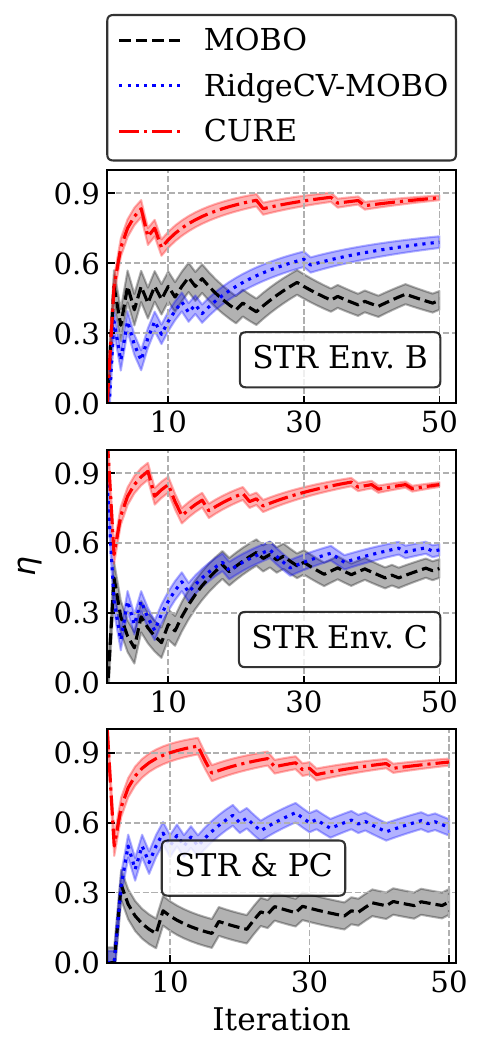}\label{fig:rq2_efficiency}}
  \hfill  
  \subfloat[Penalty]{\includegraphics[width=0.4\columnwidth]{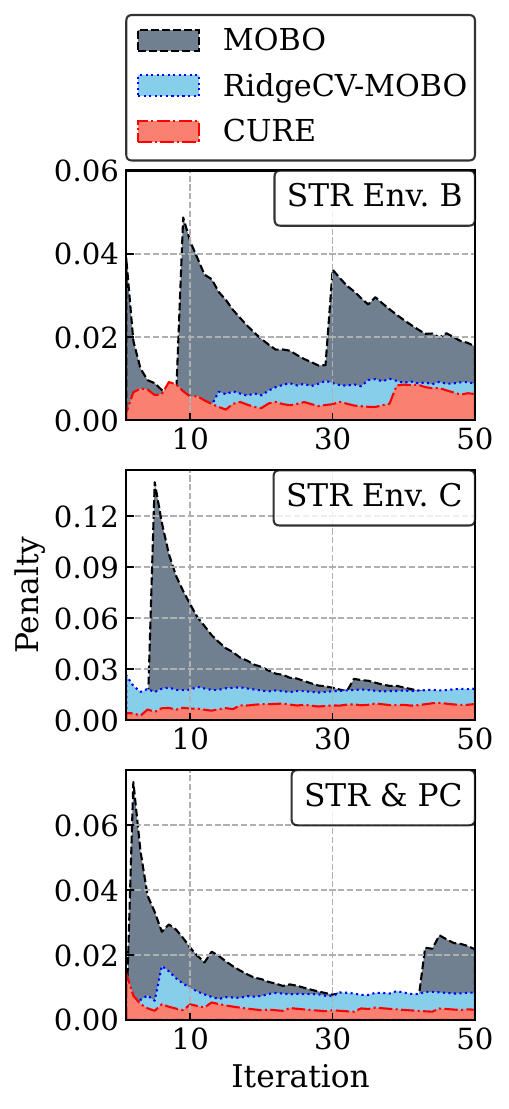}\label{fig:rq2_penalty}}
  \hfill  
  \subfloat[Violations and failures]{\includegraphics[width=0.44\columnwidth]{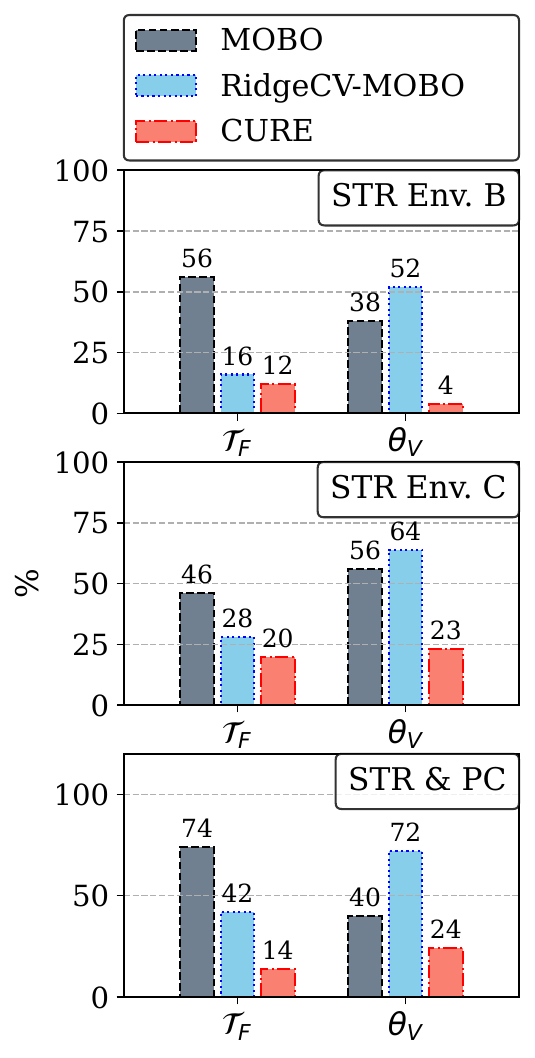}\label{fig:rq2_vio}} 
  \caption{\small{Transferability of \ourapproach and baseline methods for the navigation task}: (a)~Hypervolume; (b)~Pareto front; (c)~Efficiency; (d)~safety penalty response; and (e)~$\theta_V$ and $\mathcal{T}_F$; under varying severity of deployment changes.}
  \label{fig:rq2_trans}
  \vspace{-0.5em}
\end{figure*}

\subsection{RQ2: Transferability} \label{sec:rq2} 
Understanding \ourapproach's sensitivity to different degrees of deployment changes, such as transfer of the causal model learned from a source platform~(e.g., \textit{Gazebo} simulation) to a target platform~(e.g., real robot), is critical. Sensitivity analysis is especially crucial for such scenarios, considering that distribution shifts can occur during deployment changes. 
We answer RQ2 through an empirical study. We examine different levels of severity in deployment changes, where severity is determined by the number of changes involved. For example, a deployment change is considered more severe when both the robotic platform and the operating environment change, as opposed to changes limited solely to the environment.

\paragraph{Setting} We consider \textit{Husky} and \textit{Turltebot 3} in simulation as the source and \textit{Turtlebot 3} physical robot as the target. We evaluate two deployment scenarios~(Fig.~\ref{fig:trans_setup}): (i)~\textit{Sim-to-real}: We trained the causal model using Algorithm~\ref{alg:scm-learn} on observational data obtained by conducting a mission $1000$ times using \textit{Turtlebot 3} in \textit{Gazebo} environment~(Fig.~\ref{fig:tbot_env}). The robot was expected to encounter dynamic obstacles~(the trajectories of the obstacles are shown in Fig.~\ref{fig:tbot_env}). The mission was considered successful if \textit{Turtlebot 3} reached each of the target locations. Subsequently, we used the causal model learned from simulation~(environment A) to the \textit{Turtlebot 3} physical robot for performance optimization in two distinct environments~(environment B and C). (ii)~\textit{Sim-to-real~(STR) \& Platform change~(PC)}: We consider the change of two categories, the Sim-to-real and robotic platform change. In particular, we applied the causal model used in RQ1~(learned using \textit{Husky} in simulation) to the \textit{Turtlebot 3} physical robot in a real environment, as shown in Fig.~\ref{fig:trans_setup}. We use the identical experimental setting for the \textit{Husky} as described in \S\ref{sec:rq1}. For \textit{Turtlebot 3}, we set the objective thresholds, $\mathrm{Energy_{Th}} = 2 \ \mathrm{Wh}$ and $\mathrm{PoseError_{Th}} = 0.1 \ \mathrm{m}$. We compute the hypervolume using~Eq.~\eqref{eq:hypervolume} by setting the $f^{\operatorname{ref}}$ points at $19.98$ for energy and $3$ for position error within the coordinate system. We also set $Th_1 = 0.25$ and $Th_2 = 0.15$ in the penalty function~(Fig.~\ref{fig:penalty_func}).

\paragraph{Results} As shown in Fig.~\ref{fig:rq2_trans}, \ourapproach continuously outperforms the baselines in terms of hypervolume~(Fig.~\ref{fig:rq2_hypervolume}), Pareto front~(Fig.~\ref{fig:rq2_pareto}), efficiency~(Fig.~\ref{fig:rq2_efficiency}), penalty response~(Fig.~\ref{fig:rq2_penalty}), and violations and failures~(Fig.~\ref{fig:rq2_vio}) for each severity changes. Specifically, compared to MOBO, \ourapproach finds a configuration with $1.5 \times$ higher hypervolume in Sim-to-real setting~(low severity), and $2 \times$ higher hypervolume when we change the platform in addition to sim-to-real~(high severity). Moreover, \ourapproach achieved efficiency gains of $2.2\times$, and $4.6 \times$ over MOBO with low and high severity of deployment changes, respectively. To provide insights into the factors contributing to \ourapproach's enhanced performance, we compared constraint violation $\theta_V$ and task failure~$\mathcal{T}_F$, revealing reductions of $48\%$ in $\theta_V$, while also demonstrating $28\%$ lower $\mathcal{T}_F$ under high severity changes compared to RidgeCV-MOBO. Therefore, we conclude that \ourapproach performs better compared to the baseline methods as the deployment changes become more severe.

\section{Performance and sensitivity analysis of \ourapproach}
To explain \ourapproach's advantages over other methods, we conducted a case study employing the same experimental setup described in \S\ref{sec:rq1}. We also demonstrate \ourapproach's sensitivity by varying the top $K$ values. Our key findings are discussed in the following.

\paragraph{\ourapproach's efficient budget utilization is attributed to a comprehensive evaluation of the core configuration options}
For a more comprehensive understanding of the optimization process, we visually illustrate the response surfaces of three pairs of options, each with varying degrees of ACE in energy. Fig.~\ref{fig:con_highace} contains options with high ACE values, while Fig.~\ref{fig:con_lowace} contains only options with lower ACE values. Options with ACE values close to the median are presented in Fig.~\ref{fig:con_midace}. We observe that response surfaces with higher ACE values are more complex compared to those with lower ACE values. Figs.~\ref{fig:con_highace}-\ref{fig:con_lowace} also show that \ourapproach explored a range of configurations within the range by systematically varying configurations associated with higher ACE values than those associated with lower ones. In particular, because they have the lowest ACE, the pair of options involving trans\_stopped\_vel and max\_scaling\_factor was not considered by \ourapproach in the optimization process, avoiding allocating the budget to less effective options. In contrast, both MOBO and RidgeCV-MOBO wasted the budget exploring less effective options~(Fig.~\ref{fig:con_lowace}). Note that the option pair involving Min\_vel\_x and scalling\_speed in Fig.~\ref{fig:con_highace}, which exhibits the highest ACE, was not identified by RidgeCV-MOBO. We also observe that due to having a larger search space~(entire configuration space), MOBO struggled to explore regions effectively~(exhibits a more denser data distribution) compared to \ourapproach. In our previous study~\cite{10137745}, we evaluated the accuracy of the key configuration options identified using causal inference through a comprehensive empirical study. Therefore, \ourapproach strategically prioritize core configuration options with high ACE values, ensuring efficient budget utilization and demonstrating a better understanding of such complex behavior, while bypassing less effective options.

\begin{figure}[!t]
  \centering
  \vspace{-0.5em}
  \subfloat[Causal structure overlaps]
  {\includegraphics[width=.5\columnwidth, valign=t]{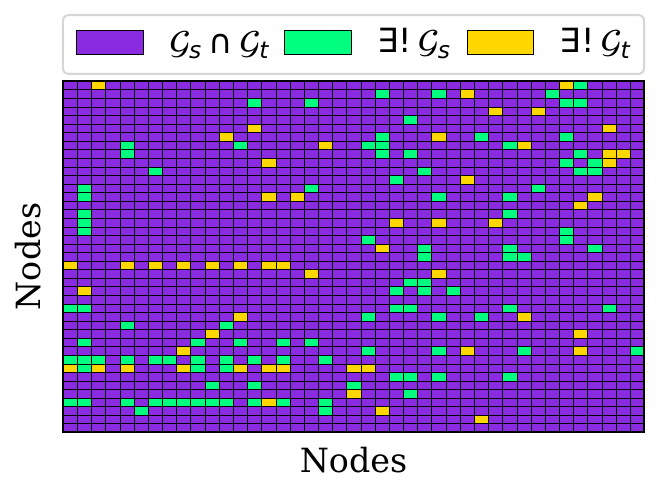}\label{fig:caual_overlap}}
  \hfill
  \subfloat[High ACE]{\includegraphics[width=0.5\columnwidth, valign=t]{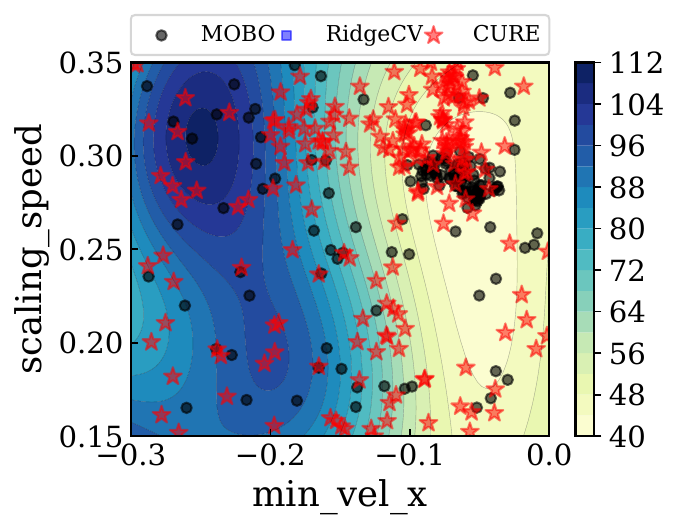}\label{fig:con_highace}}
  \hfill
  \subfloat[Median ACE]{\includegraphics[width=0.48\columnwidth,]{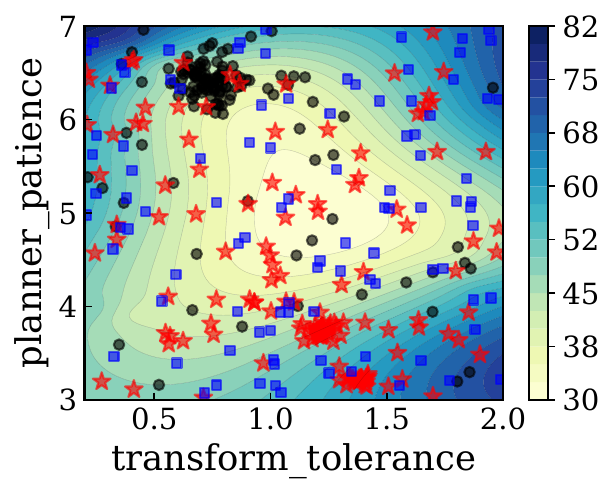}\label{fig:con_midace}} 
  \hfill
  \subfloat[Low ACE]{\includegraphics[width=0.515\columnwidth,]{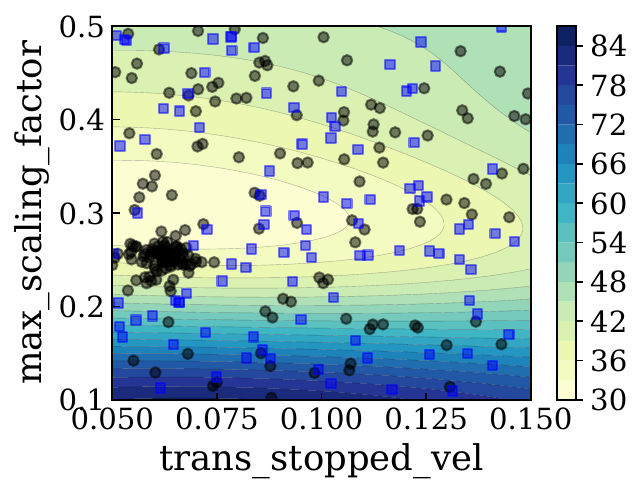}\label{fig:con_lowace}}  
  \caption{\small{(a) Significant overlap between causal structures~(common edges are represented as purple squares) developed in \textit{Husky}~($\mathcal{G}_s$) and \textit{Turtlebot 3}~($\mathcal{G}_t$). Unique edges are represented as green and yellow squares in $\mathcal{G}_s$ and $\mathcal{G}_t$, respectively. (b) (c) and (d)~represents contour plot with options of different causal effects. The color bar indicates the energy values, where lower values indicate better performance.}}
  \label{fig:dis_contours}
  \vspace{-0.5em}
\end{figure}

\paragraph{\ourapproach leverages the knowledge derived from the causal model learned on the source platform} In Fig.~\ref{fig:caual_overlap}, we compare the adjacency matrix between causal graphs learned from the source and target platforms, respectively. We compute the adjacency matrix $A$ from a causal graph $G=(V,E)$, where $V$ is the set of vertices and $E$ is the set of edges, as follows: 
\begin{equation}
        A[i][j] =
    \begin{cases}
        1, & \text{if } (i,j) \in E \\
        0, & \text{otherwise}
    \end{cases}
\end{equation}
where $(i,j)$ represents the edge from vertex $i$ to vertex $j$. In particular, both causal graphs share a significant overlap, providing a rationale for \ourapproach's enhance performance when transferring the causal model learned from a source~(e.g, \textit{Husky} in simulation) to a target~(e.g., \textit{Turtlebot 3} physical platform). Therefore, a causal model developed on one platform or environment can be leveraged as prior knowledge on another, demonstrating the cross-platform applicability and usefulness of the acquired causal understanding.

\paragraph{How sensitive is \ourapproach when the value of top $K$ varies?}
We investigate \ourapproach's performance with different $K$ values and how it affects the optimization process. We conduct a single-objective optimization on the \textit{Turtlebot 3} platform to demonstrate the sensitivity of \ourapproach. As shown in Fig~\ref{fig:topK_sensitivity}, there is a trade-off between the top $K$ values and the iterations required to achieve high-quality solutions. Smaller $K$ values allow the optimization process to quickly find low energy values but may limit exploration, leading to early plateauing. Conversely, larger $K$ values enable more extensive exploration, leading to more gradual improvements and potentially better solutions, but requiring more iterations. This is because, when the search space is smaller, the optimization process can exploit known good areas more effectively. In contrast, a larger search space requires more exploration, which extends the optimization process. One approach for selecting $K$ is to define a threshold on the ACE values and select options that exceed this threshold. This can be done by using a threshold defined as $\{X \mid X_{\mathrm{ACE}} > \mu_{\mathrm{ACE}} + \sigma_{\mathrm{ACE}}\}$, where $\mu_{\mathrm{ACE}}$ is the mean and $\sigma_{\mathrm{ACE}}$ is the standard deviation of the ACE values. Alternatively, a threshold based on the percentile of ACE values can be employed, such as selecting options with ACE values greater than the $75^{th}$ percentile. We leave this selection up to the practitioner as user preferences may vary depending on the task, environment, and robotic system.
\begin{figure}[!t]
    \vspace{0.5em}
    \begin{center}
    \centerline{\includegraphics[width=0.65\columnwidth]{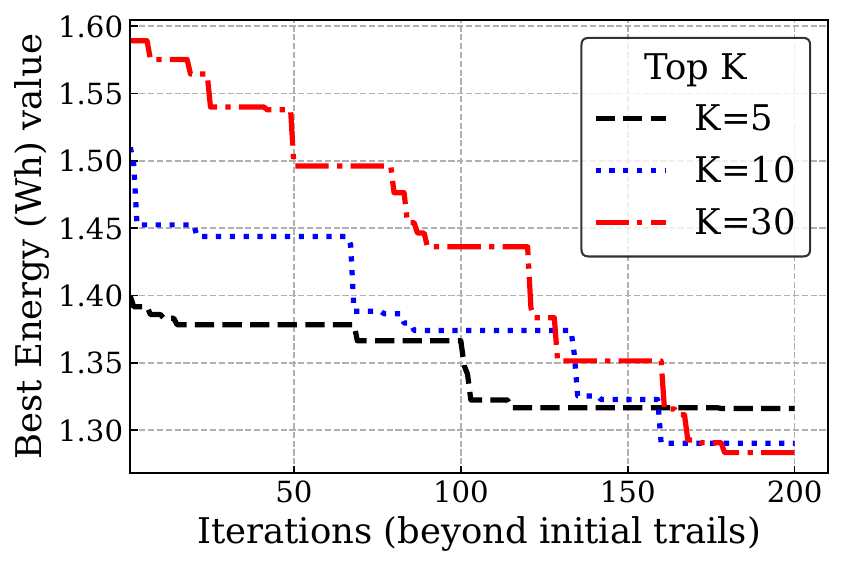}}
    \caption{\small{Sensitivity of \ourapproach under different top $K$ values.}}
    \label{fig:topK_sensitivity}
    \end{center} 
    \vspace{-2em}
\end{figure}
\section{Discussion}\label{sec:discussion}
\subsection{Usability of \ourapproach}
The design we have proposed is general and extendable to other robotic systems but would require some engineering effort. In particular, to apply \ourapproach to a novel problem, the practitioner must identify (i)~configuration options, (ii)~performance metrics, and (iii)~key performance indicators (KPIs). Note that the abstraction level of the variables in the causal model depends on the practitioner and can go all the way down to the hardware level. In defining the metrics and KPIs, guidelines provided by the National Institute of Standards and Technology (NIST) can be used~\cite{zimmerman2017metrics, falco2018performance}. These guidelines help classify variables as non-manipulable in the three-layer causal model design~\cite{10137745}, which simplifies the performance modeling process by allowing a clear distinction between configurable and performance variables. Moreover, we provide various performance metrics and performance objectives for mobile robot navigation and robot manipulation tasks in \S\ref{sec:exp}.

\subsection{Limitations}\label{sec:lim}
\paragraph{Causal model error} The NP-hard complexity of causal discovery introduces a challenge~\cite{chickering2004large}, implying that the identified causal model may not always represent the ground-truth causal relationships among variables. It is crucial to recognize the potential for discrepancies between the causal structure discovered and the actual structures. However, such causal models can still be employed to achieve better performance compared to ML-based approaches in systems optimization~\cite{dubslaff2022causality} and debugging tasks~\cite{hossen2023care}, because causal models avoid capturing spurious correlations~\cite{glymour2019review}.

\paragraph{Potential biases when transferring the causal model} Caution must be exercised when reusing the entire causal graph learned from the source platform, as differences between causal graphs in the two platforms~(as indicated by the green and yellow squares in Fig.~\ref{fig:caual_overlap}, representing edges unique to the source and target, respectively) can induce bias. It is crucial to discover new causal connections~(indicated by the yellow squares in Fig.~\ref{fig:caual_overlap}) on the target platform based on observations. Given the small number of edges to be discovered, this task can easily be accomplished with a limited number of observational samples from the target platform.

\subsection{Future directions}\label{sec:cure_future}
\paragraph{Incorporating Causal Gaussian Process~(CGP)} Using CGP in the optimization process has the potential to capture the behavior of the performance objective better compared to traditional GP~\cite{aglietti20a}. Unlike GP, CGP represents the mean using interventional estimates via do-calculus. This characteristic renders CGP particularly useful in scenarios with a limited amount of observational data or in areas where observational data is not available.

\paragraph{Updating the causal model at run-time} There is potential in employing an active learning mechanism that combines the source causal model $\mathcal{G}_s$ with a new causal model $\mathcal{G}_t$ learned from a small number of samples from the target platform. This approach is particularly promising considering the limitations discussed in \S\ref{sec:lim}.

\paragraph{Dynamically selecting top $K$ at run-time}
In our framework, $K$ is a hyperparameter and its value is defined by the practitioner. Motivated by Fig~\ref{fig:topK_sensitivity}, there is potential for implementing a dynamic selection approach. This approach would start with a lower $K$ and progressively increase the $K$ if the objective reaches a plateau.
\section{Conclusion}
We presented \ourapproach, a multi-objective optimization method that identified optimal configurations for robotic systems. \ourapproach converged faster than the baseline methods and demonstrated effective transferability from simulation to real robots, and even to new untrained platforms. \ourapproach constructs a causal model based on observational data collected from a source environment, typically a low-cost setting such as the \textit{Gazebo} simulator. We then estimate the causal effects of configuration options on performance objectives, reducing the search space by eliminating configuration options that have negligible causal effects. Finally, \ourapproach employs traditional Bayesian optimization in the target environment, but confines it to the reduced search space, thus efficiently identifying the optimal configuration. Empirically, we have demonstrated that \ourapproach not only finds the optimal configuration faster than the baseline methods, but the causal models learned in simulation accelerate optimization in real robots. Moreover, our evaluation shows the learned causal model is transferable across similar but different settings, encompassing different environments, mission/tasks, and new robotic systems.
\appendices
\newpage
\section{Additional Details}\label{sec:appendix_B}
\subsection{Background and definitions}
\subsubsection{Configuration space $\mathcal{X}$}\label{sec:config_space_def} Consider $X_i$ as the $i^{th}$ configuration option of a robot, which can be assigned a range of values~(e.g., categorical, boolean, and numerical). The configuration space $\mathcal{X}$ is a Cartesian product of all options and a configuration $\mathbf{x} \in \mathcal{X}$ in which all options are assigned specific values within the permitted range for each option. Formally, we define:
\begin{itemize}
    \item Configuration option: $X_1,X_2,\cdots, X_d$
    \item Option value: $x_1, \dots, x_d$    
    \item Configuration: $\mathbf{x} = \langle X_1=x_1,\dots,X_d=x_d \rangle$       
    \item Configuration space: $\mathcal{X}=Dom(X_1) \times \dots \times Dom(X_d)$      
\end{itemize}

\subsubsection{Partial Ancestral Graph (PAG)}\label{sec:pag}  Each edge in the PAG denotes the ancestral connections between the vertices. A PAG is composed of the following types of edges:
\begin{itemize}
    \item $A$\edgeone $B$: The vertex $A$ causes $B$.
    \item $A$\edgetwo$B$: There are unmeasured confounders between the vertices $A$ and $B$.
    \item $A$\edgethree $B$: $A$ causes $B$, or there are unmeasured confounders that cause both $A$ and $B$. 
    \item $A$\edgefour$B$: $A$ causes $B$, or $B$ causes $A$, or there are unmeasured confounders that cause both $A$ and $B$.
\end{itemize}
For a comprehensive theoretical foundation on these ideas, we refer the reader to~\cite{colombo2012learning, colombo2014order, pearl2016causal}

\subsubsection{Causal model $\mathcal{G}$}\label{sec:causal_model}
A causal model is an acyclic-directed mixed graph~(ADMG)~\cite{richardson2002ancestral, evans2014markovian} which encodes performance variables, functional
nodes (which defines functional dependencies between performance variables such as how variations in one or multiple variables determine variations in other variables), causal
links that interconnect performance nodes with each other
via functional nodes. An ADMG is defined as a finite collection of vertices, denoted by $V$, and directed edges $E_{\mathpzc{d}}$~(ordered pairs $E_{\mathpzc{d}} \subset V \times V$, such that $(v, v) \not \in E_{\mathpzc{d}}$ for any vertex $v$), together with bidirected edges, denoted by $E_{\mathpzc{b}}$~(unordered pairs of elements of $V$). If $(v,w) \in E_{\mathpzc{b}}$ then $v \leftrightarrow w$, and if in addition $(v, w) \in E_{\mathpzc{d}}$ then $v\; {\tikz[baseline=-2pt]{
\draw[<->] (0cm,0cm) -- (3mm,0cm);
\draw[->] (0cm,4pt) -- (3mm,4pt);}
}\; w$.

\subsubsection{Causal paths $P_{X \curly Y}$}
 
We define $P=\langle v_0, v_1,\dots,v_n \rangle$ so that the following conditions hold:
\begin{itemize}
    \item $v_o$ is the root cause of the functional fault and $v_n = y_F$.
    \item $\forall\ 0 \leq i \leq n,\ v_i \in V$ and $\forall\ 0 \leq i \leq n,\ (v_i, v_{i+1}) \in (E_{\mathpzc{d}} \vee E_{\mathpzc{b}})$.
    \item $\forall\ 0 \leq i \leq j \leq n$, $v_i$ is a counterfactual cause of $v_j$.
    \item $\left|P\right|$ is maximized.
\end{itemize}

\begin{figure}[t]
\centering
\centerline{\includegraphics[width=1\columnwidth]{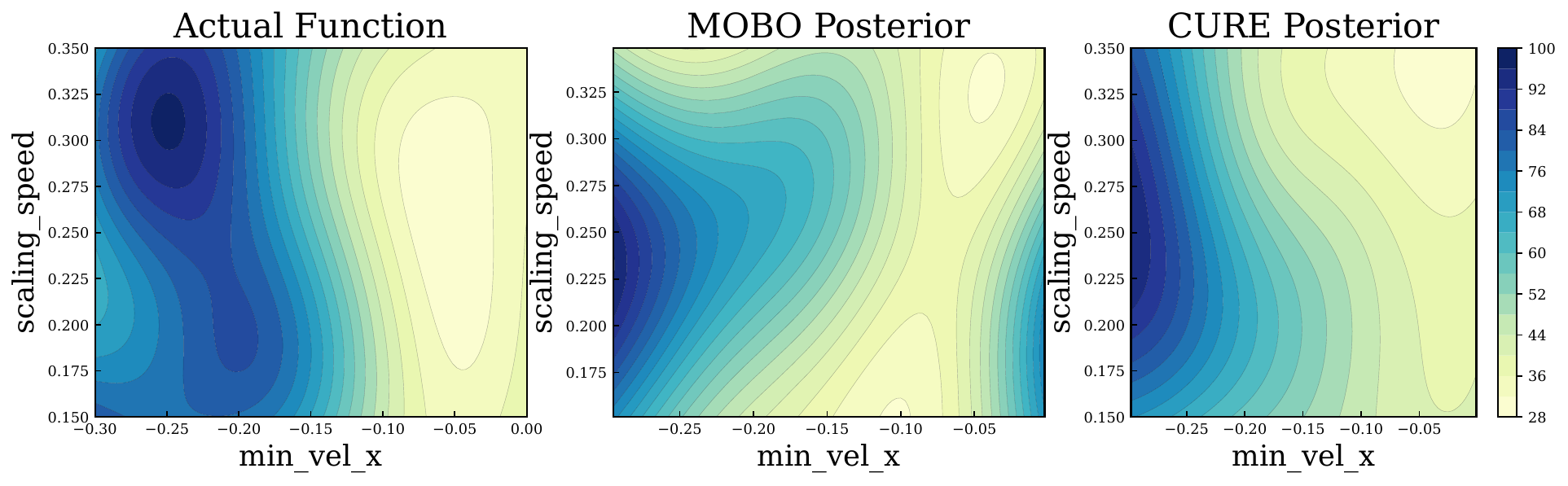}}
\caption{\small{\ourapproach demonstrates a better understanding about the performance behavior compared to MOBO. The actual function was derived from $1000$ observational samples. The color bar indicates energy values.}}
\label{fig:dis_contours_appen}
\end{figure}

\begin{figure}[t]
\centering
\centerline{\includegraphics[width=0.65\columnwidth]{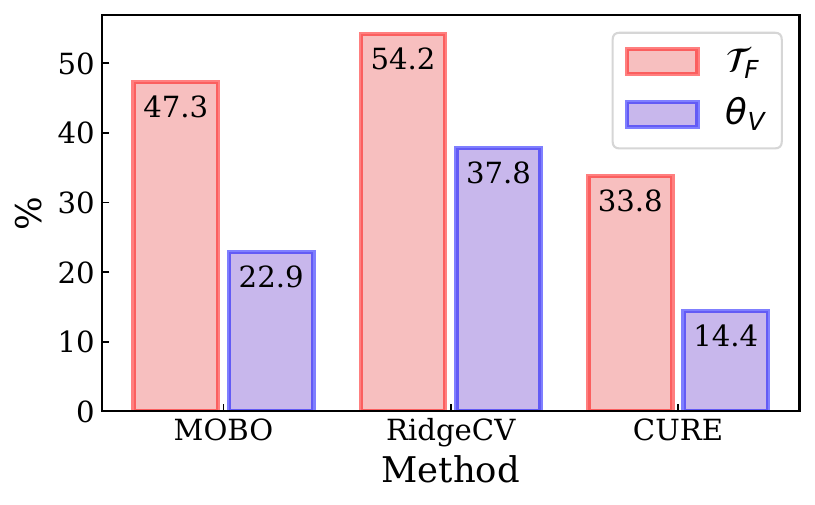}}
\caption{\small{$\theta_V$ and $\mathcal{T}_F$ for RQ1.}}
\label{fig:husky_vio}
\end{figure}

\begin{table}[!t]
\centering
\caption{Configuration options in \textit{move base}.}
\label{tab:move_base}
\begin{tblr}{
  cell{1}{1} = {r=2}{},
  cell{1}{2} = {c=2}{},
  hline{1,3,10} = {-}{},
  hline{2} = {2-3}{},
}
Configuration Options & Option Values/Range & \\
 & Husky & Turtlebot 3\\
\textsf{controller\_frequency} & 3.0 - 7.0 & 5.0 - 15.0\\
\textsf{planner\_patience} & 3.0 - 7.0 & 3.0 - 7.0\\
\textsf{controller\_patience} & 3.0 - 7.0 & 10.0 - 20.0\\
\textsf{conservative\_reset\_dist} & 1.0 - 5.0 & 1.0 - 5.0\\
\textsf{planner\_frequency} & 0.0 & 5.0\\
\textsf{oscillation\_timeout} & 5.0 & 3.0\\
\textsf{oscillation\_distance} & 0.5 & 0.2
\end{tblr}
\end{table}

\begin{table}[!t]
\centering
\caption{Configuration options in \textit{costmap common}.}
\label{tab:costmap_common}
\begin{tblr}{
  cell{1}{1} = {r=2}{},
  cell{1}{2} = {c=2}{c},
  hline{1,3,7} = {-}{},
  hline{2} = {2-3}{},
}
Configuration Options & Option Values/Range & \\
 & Husky & Turtlebot 3\\
\textsf{publish\_frequency} & 1.0 - 6.0 & 5.0 - 20.0\\
\textsf{resolution} & 0.02 - 0.15 & 0.02 - 0.15\\
\textsf{transform\_tolerance} & 0.2 - 2.0 & 0.2 - 2.0\\
\textsf{update\_frequency} & 1.0 - 6.0 & 5.0 - 20.0
\end{tblr}
\end{table}

\begin{table}[!t]
\centering
\caption{Configuration options in \textit{costmap common inflation}.}
\label{tab:costmap_common_inflation}
\begin{tblr}{
  cell{1}{1} = {r=2}{},
  cell{1}{2} = {c=2}{c},
  hline{1,3,5} = {-}{},
  hline{2} = {2-3}{},
}
Configuration Options & Option Values/Range & \\
 & Husky & Turtlebot 3\\
\textsf{cost\_scaling\_factor} & 1.0 - 20.0 & 3.0 - 20.0\\
\textsf{inflation\_radius} & 0.3 - 1.5 & 0.3 - 2.0
\end{tblr}
\end{table}

\setlength{\textfloatsep}{10pt}
\begin{table}[!t]
\centering
\caption{Configuration options in \textit{DWAPlannerROS}.}
\label{tab:DWAPlannerROS}
\begin{tblr}{
  cell{1}{1} = {r=2}{},
  cell{1}{2} = {c=2}{c},
  hline{1,3,35} = {-}{},
  hline{2} = {2-3}{},
}
Configuration Options & Option Values/Range & \\
 & Husky & Turtlebot 3\\
\textsf{acc\_lim\_theta} & 1.5 - 5.2 & 2.0 - 4.5\\
\textsf{acc\_lim\_trans} & 0.1 - 0.5 & 0.05 - 0.3\\
\textsf{acc\_lim\_x} & 1.0 - 5.0 & 1.5 - 4.0\\
\textsf{acc\_lim\_y} & 0.0 & 0.0\\
\textsf{angular\_sim\_granularity} & 0.1 & 0.1\\
\textsf{forward\_point\_distance} & 0.225 - 0.725 & 0.225 - 0.525\\
\textsf{goal\_distance\_bias} & 5.0 - 40.0 & 10.0 - 40.0\\
\textsf{max\_scaling\_factor} & 0.1 - 0.5 & 0.1 - 0.4\\
\textsf{max\_vel\_theta} & 0.5 - 2.0 & 1.5 - 4.0\\
\textsf{max\_vel\_trans} & 0.3 - 0.75 & 0.15 - 0.4\\
\textsf{max\_vel\_x} & 0.3 - 0.75 & 0.15 - 0.4\\
\textsf{max\_vel\_y} & 0.0 & 0.0\\
\textsf{min\_vel\_theta} & 1.5 - 3.0 & 0.5 - 2.5\\
\textsf{min\_vel\_trans} & 0.1 - 0.2 & 0.08 - 0.22\\
\textsf{min\_vel\_x} & -0.3 - 0.0 & -0.3 - 0.0\\
\textsf{min\_vel\_y} & 0.0 & 0.0\\
\textsf{occdist\_scale} & 0.05 - 0.5 & 0.01 - 0.15\\
\textsf{oscillation\_reset\_angle} & 0.1 - 0.5 & 0.1 - 0.5\\
\textsf{oscillation\_reset\_dist} & 0.25 & 0.25\\
\textsf{path\_distance\_bias} & 10.0 - 50.0 & 20.0 - 45.0\\
\textsf{scaling\_speed} & 0.15 - 0.35 & 0.15 - 0.35\\
\textsf{sim\_granularity} & 0.015 - 0.045 & 0.015 - 0.045\\
\textsf{sim\_time} & 0.5 - 3.5 & 0.5 - 2.5\\
\textsf{stop\_time\_buffer} & 0.1 - 1.5 & 0.1 - 1.5\\
\textsf{theta\_stopped\_vel} & 0.05 - 0.15 & 0.05 - 0.15\\
\textsf{trans\_stopped\_vel} & 0.05 - 0.15 & 0.05 - 0.15\\
\textsf{twirling\_scale} & 0.0 & 0.0\\
\textsf{vth\_samples} & 10 - 30 & 20 - 50\\
\textsf{vx\_samples} & 3 - 10 & 10 - 30\\
\textsf{vy\_samples} & 0 - 15 & 0 - 5\\
\textsf{xy\_goal\_tolerance} & 0.2 & 0.08\\
\textsf{yaw\_goal\_tolerance} & 0.1 & 0.17
\end{tblr}
\end{table}

\begin{table}[!t]
\centering
\caption{Configuration options in \textit{moveit chmop planning}.}
\label{tab:chomp}
\begin{tblr}{
  row{1} = {c},
  hline{1,20} = {-}{0.08em},
  hline{2} = {-}{},
}
Configuration options & Option Values/Range\\
\textsf{planning\_time\_limit} & 1.0 - 10.0\\
\textsf{max\_iterations} & 1 - 500\\
\textsf{max\_iterations\_after\_collision\_free} & 1 - 10\\
\textsf{smoothness\_cost\_weight} & 0.05 - 5.0\\
\textsf{obstacle\_cost\_weight} & 0.0 - 1.0\\
\textsf{learning\_rate} & 0.001 - 0.5\\
\textsf{smoothness\_cost\_velocity} & 0.0 - 10.0\\
\textsf{smoothness\_cost\_acceleration} & 0.0 - 10.0\\
\textsf{smoothness\_cost\_jerk} & 0.0 - 10.0\\
\textsf{ridge\_factor} & 0.0 - 0.01\\
\textsf{use\_pseudo\_inverse} & True, False\\
\textsf{pseudo\_inverse\_ridge\_facto}r & 0.00001 - 0.001\\
\textsf{joint\_update\_limit} & 0.05 - 5.0\\
\textsf{collision\_clearance} & 0.05 - 2.0\\
\textsf{collision\_threshold} & 0.01 - 0.15\\
\textsf{use\_stochastic\_descent} & True, False\\
\textsf{enable\_failure\_recovery} & True, False\\
\textsf{max\_recovery\_attempts} & 0 - 10
\end{tblr}
\end{table}

\begin{figure}[t]
\vspace{-2em}
  \centering
  \subfloat[Interaction with energy]{\includegraphics[width=.5\columnwidth]{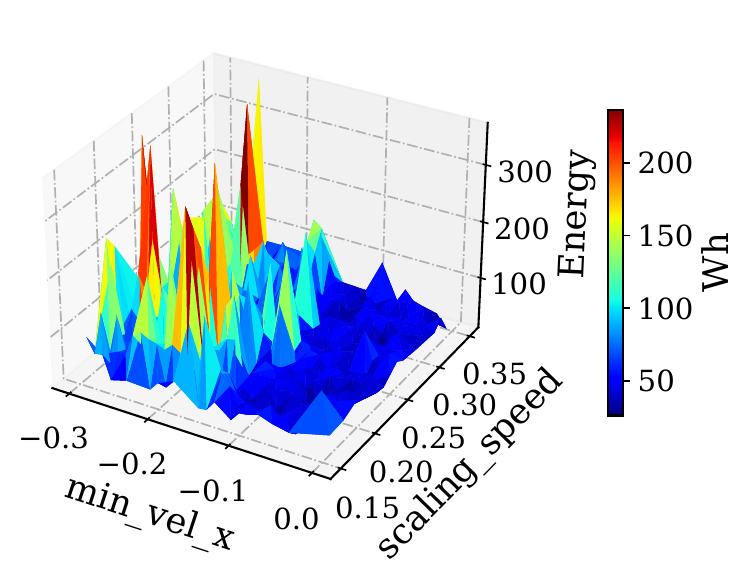}\label{fig:surf_energy}}
  \hfill
  \subfloat[Interaction with position error]{\includegraphics[width=0.5\columnwidth,]{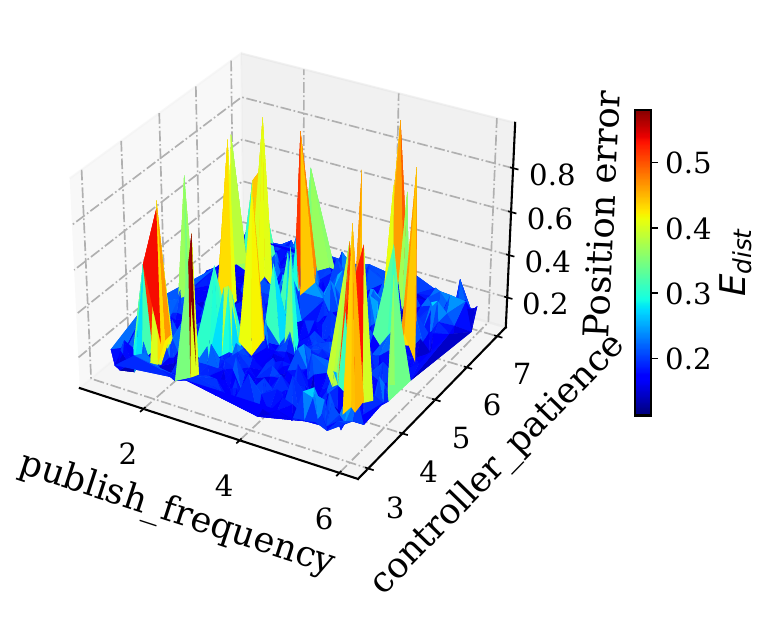}\label{fig:surf_poseerror}}
  \hfill
  \subfloat[Interaction with $\mathcal{T}_{cr}$]{\includegraphics[width=0.5\columnwidth,]{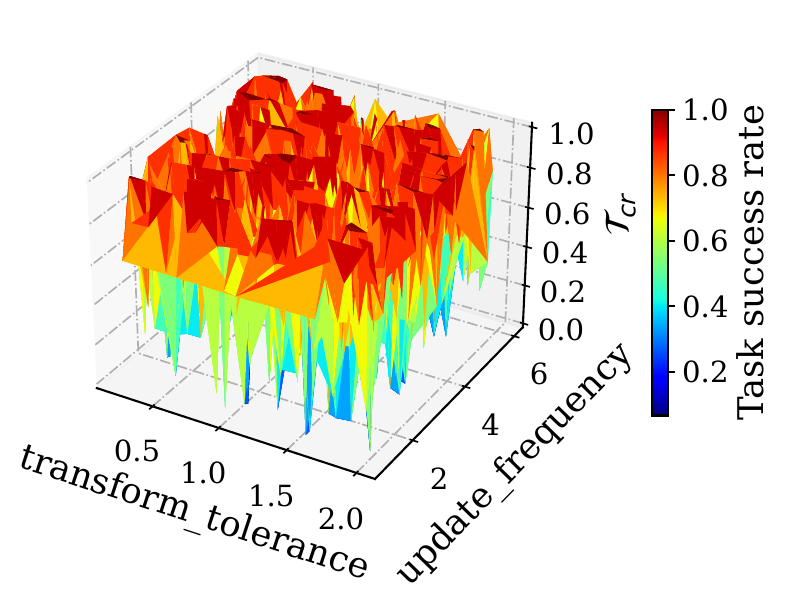}\label{fig:surf_tcr}}  
  \hfill
  \subfloat[Interaction with safety constraint]{\includegraphics[width=0.5\columnwidth,]{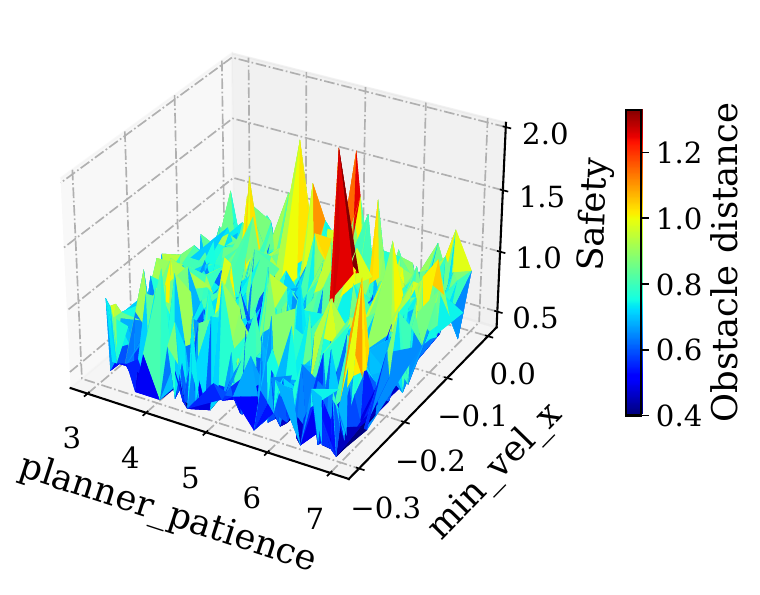}\label{fig:surf_safety}}   
  \caption{\small{Pairwise interactions between high ACE configuration options, performance objectives, and constraints, derived from observational data.}
  \label{fig:obs_surf}}
\end{figure}

\subsubsection{Why do robotic systems fail?} 
A robotic system may fail to perform a specific task or deteriorate from the desired performance due to (i)~Hardware faults: physical faults of the robot's equipment~(e.g., faulty controller), (ii)~Software faults: faulty algorithms and/or faulty implementations of correct algorithms~(e.g., incorrect cognitive behavior of the robot), (iii) interaction faults: failures that result from uncertainties in their environments. The software stack is typically composed of multiple components (e.g., localization, navigation), each with a plethora of configuration options~(different planner algorithms and/or parameters in the same planner algorithm). Similarly to software components, hardware components also have numerous configuration options. Incorrect configurations can cause a functional fault (the robot cannot perform a task successfully) or a non-functional fault (the robot may be able to finish tasks, but with undesired performance).

\subsubsection{Non-functional fault} The non-functional faults (interchangeably used as \textit{performance faults}) refer to cases where the robot can perform the specified task but cannot meet the specified performance requirements of the task specification. For example, the robot reached the target location(s); however, it consumed more energy. We define the non-functional property $\mathcal{N}=\{p_1, \dots, p_n\}$, where $p_1, \dots, p_n$ represents different non-functional properties of the robotic system~(e.g., energy, mission time) and $p_j$ is the value of $j^{th}$ $\mathcal{N}$. The specified performance goal is denoted as $p_{js}$. Performance failure occurs when $p_j \not \models p_{js}$. Extending the previous scenario, let $E^i$ be the energy consumption during task $i$ and let $T^i$ be the mission completion time. The specified performance goals for the task are indicated as $E_{s->t}<= en, T_{s->t} <= tt$ respectively. A non-functional fault can be defined as $N_F = (E^i > en) \lor (T^i > tt)$.
\begin{figure}[t]
    \vspace{-2.5em}
  \centering
  \subfloat[Interaction with energy]{\includegraphics[width=.5\columnwidth]{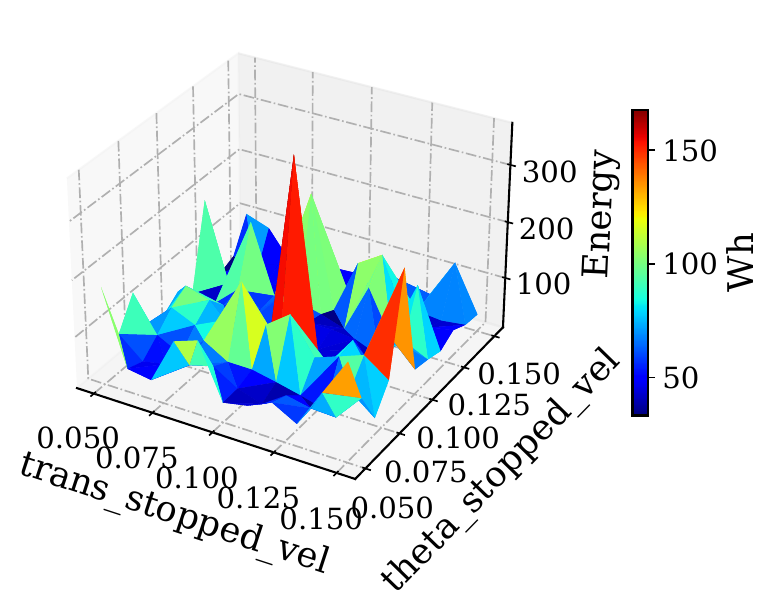}\label{fig:surf_energy_lowace}}
  \hfill
  \subfloat[Interaction with position error]{\includegraphics[width=0.5\columnwidth,]{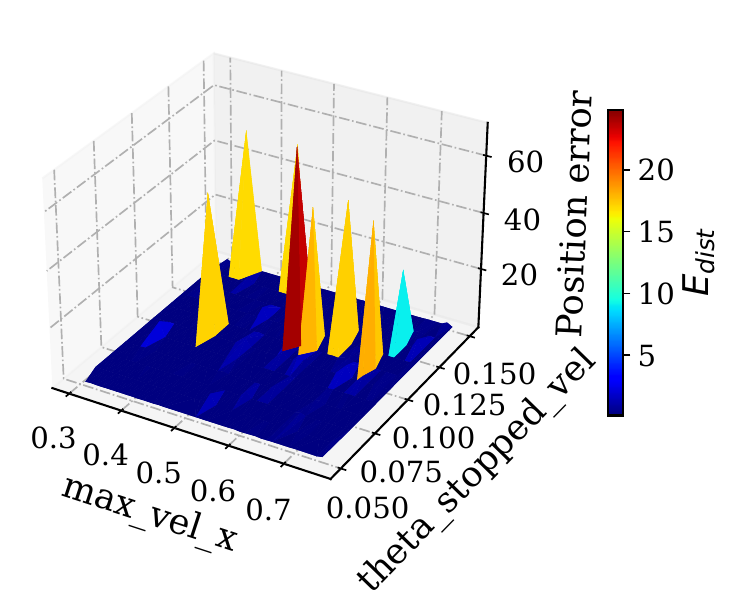}\label{fig:surf_poseerror_lowace}}
  \hfill
  \vspace{-0.6em}
  \subfloat[Interaction with $\mathcal{T}_{cr}$]{\includegraphics[width=0.5\columnwidth,]{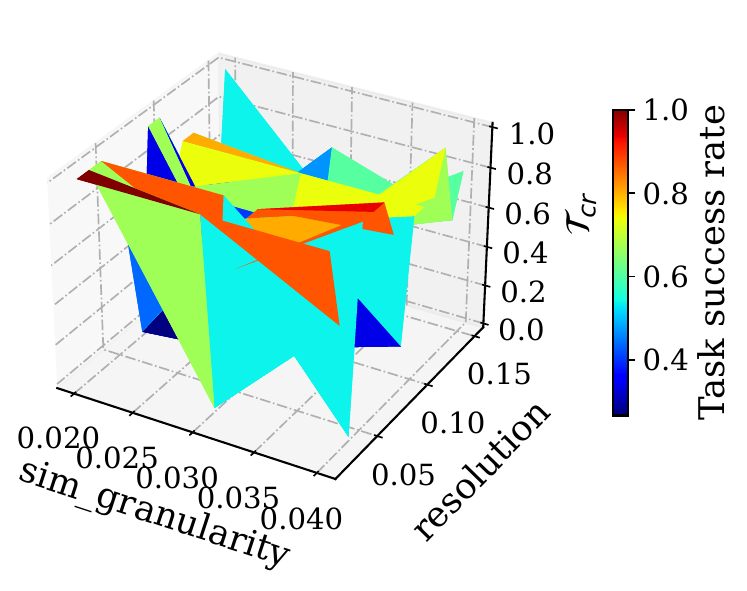}\label{fig:surf_tcr_lowace}} 
  \hfill
  \subfloat[Interaction with safety constraint]{\includegraphics[width=0.5\columnwidth,]{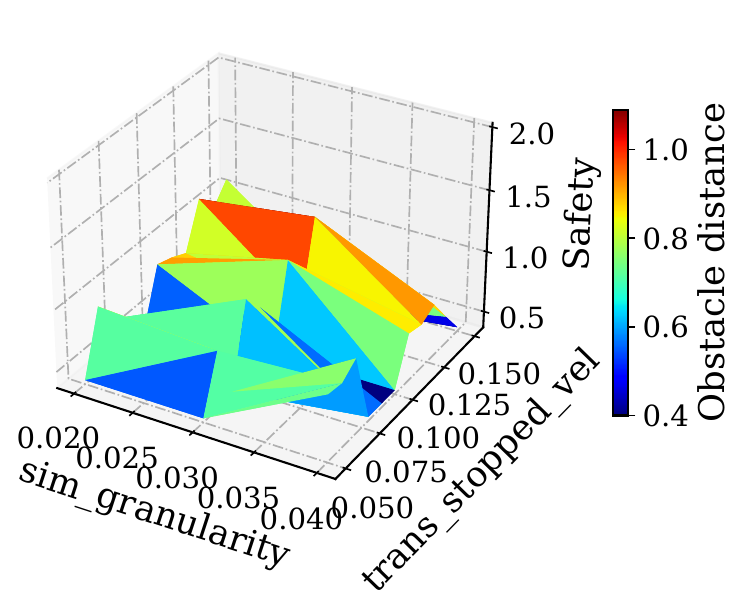}\label{fig:surf_safety_lowace}}   
  \caption{\small{Pairwise interactions between low ACE configuration options, performance objectives, and constraints, derived from observational data.}
  \label{fig:obs_surf_lowace}}
\end{figure}

\begin{table}[!t]
\centering
\caption{ACE values of the configuration options.}
\label{tab:ace_values}
\begin{tabular}{lllll} 
\hline
\multicolumn{1}{c}{\multirow{2}{*}{\begin{tabular}[c]{@{}c@{}}\\~Configuration Options\end{tabular}}} & \multicolumn{4}{c}{ACE} \\ 
\cline{2-5}
\multicolumn{1}{c}{} & Energy & \begin{tabular}[c]{@{}l@{}}Positional\\error\end{tabular} & $\mathcal{T}_{cr}$ & Safety \\ 
\hline
\rowcolor[rgb]{0.761,0.816,0.98} \textsf{scaling\_speed} & 199.349 & 65.980 & 0.454 & 0.633 \\
\rowcolor[rgb]{0.761,0.816,0.98} \textsf{min\_vel\_x} & 115.496 & 18.764 & 0.566 & 0.217 \\
{\cellcolor[rgb]{0.761,0.816,0.98}}\textsf{controller\_frequency} & {\cellcolor[rgb]{0.761,0.816,0.98}}25.370 & 3.695 & 0.050 & 0.019 \\
{\cellcolor[rgb]{0.761,0.816,0.98}}\textsf{publish\_frequency} & {\cellcolor[rgb]{0.761,0.816,0.98}}19.598 & {\cellcolor[rgb]{0.761,0.816,0.98}}6.026 & {\cellcolor[rgb]{0.761,0.816,0.98}}0.223 & 0.016 \\
{\cellcolor[rgb]{0.761,0.816,0.98}}\textsf{sim\_time} & {\cellcolor[rgb]{0.761,0.816,0.98}}15.570 & {\cellcolor[rgb]{0.761,0.816,0.98}}4.680 & 0.110 & {\cellcolor[rgb]{0.761,0.816,0.98}}0.062 \\
\textsf{acc\_lim\_x} & 12.589 & 3.050 & 0.013 & 0.016 \\
\textsf{stop\_time\_buffer} & 12.130 & 3.292 & 0.069 & 0.011 \\
\textsf{inflation\_radius} & 11.267 & 3.663 & 0.049 & 0.017 \\
\textsf{path\_distance\_bias} & 10.550 & 2.559 & 0.033 & 0.021 \\
{\cellcolor[rgb]{0.761,0.816,0.98}}\textsf{max\_vel\_theta} & 10.507 & 0.394 & 0.026 & {\cellcolor[rgb]{0.761,0.816,0.98}}0.062 \\
{\cellcolor[rgb]{0.761,0.816,0.98}}\textsf{update\_frequency} & 9.250 & 3.362 & {\cellcolor[rgb]{0.761,0.816,0.98}}0.118 & 0.019 \\
\textsf{vth\_samples} & 8.599 & 2.083 & 0.028 & 0.021 \\
\textsf{cost\_scaling\_factor} & 8.565 & 1.092 & 0.014 & 0.021 \\
\textsf{min\_vel\_theta} & 8.152 & 0.049 & 0.027 & 0.016 \\
\textsf{conservative\_reset\_dist} & 7.693 & 2.808 & 0.025 & 0.014 \\
{\cellcolor[rgb]{0.761,0.816,0.98}}\textsf{planner\_patience} & 7.532 & 2.656 & 0.022 & {\cellcolor[rgb]{0.761,0.816,0.98}}0.069 \\
{\cellcolor[rgb]{0.761,0.816,0.98}}\textsf{transform\_tolerance} & 7.103 & 3.892 & {\cellcolor[rgb]{0.761,0.816,0.98}}0.148 & 0.038 \\
\textsf{vy\_samples} & 5.614 & 2.107 & 0.021 & 0.017 \\
\textsf{goal\_distance\_bias} & 5.159 & 1.365 & 0.028 & 0.013 \\
\textsf{vx\_samples} & 4.901 & 0.847 & 0.088 & 0.014 \\
\textsf{forward\_point\_distance} & 4.877 & 1.100 & 0.032 & 0.006 \\
{\cellcolor[rgb]{0.761,0.816,0.98}}\textsf{controller\_patience} & 4.116 & {\cellcolor[rgb]{0.761,0.816,0.98}}4.613 & 0.031 & 0.016 \\
\textsf{acc\_lim\_theta} & 4.101 & 1.835 & 0.043 & 0.006 \\
\textsf{occdist\_scale} & 2.803 & 0.804 & 0.035 & 0.000 \\
\textsf{acc\_lim\_trans} & 2.349 & 0.818 & 0.015 & 0.003 \\
\textsf{max\_vel\_trans} & 2.080 & 0.307 & 0.007 & 0.000 \\
\textsf{oscillation\_reset\_angle} & 1.791 & 0.715 & 0.028 & 0.007 \\
\textsf{max\_vel\_x} & 1.150 & 0.057 & 0.000 & 0.003 \\
\textsf{min\_vel\_trans} & 0.948 & 0.146 & 0.002 & 0.000 \\
\textsf{resolution} & 0.188 & 0.266 & 0.010 & 0.001 \\
\textsf{sim\_granularity} & 0.114 & 0.000 & 0.001 & 0.000 \\
\textsf{trans\_stopped\_vel} & 0.106 & 0.042 & 0.002 & 0.000 \\
\textsf{max\_scaling\_factor} & 0.059 & 0.062 & 0.021 & 0.005 \\
\textsf{theta\_stopped\_vel} & 0.000 & 0.000 & 0.000 & 0.002 \\
\hline
\end{tabular}
\end{table}

\subsection{Additional details about experimental setup}
\subsubsection{Configuration Options in \textit{ROS nav core} and \textit{Moveit}}\label{sec:config_space}
Table~\ref{tab:move_base}-\ref{tab:DWAPlannerROS} shows the configuration space for each component in the ROS navigation stack and Table~\ref{tab:chomp} shows the configuration space in \textit{Moveit chomp planning} used in our experiments. We fixed the goal tolerance parameters~({\texTTT{xy\_goal\_tolerance}}, and {\texTTT{yaw\_goal\_tolerance}}) to determine if a target was reached. Complex interactions between options~(intra or inter components) give rise to a combinatorially large configuration space.

\subsection{Additional details for evaluation}
\subsubsection{RQ1 additional details} 

We also compared $\theta_V$ and $\mathcal{T}_F$, revealing reductions of $8.5\%$ in $\theta_V$, while also demonstrating lower $13.5\%$ $\mathcal{T}_F$ compared to MOBO as shown in Fig.~\ref{fig:husky_vio}.

\subsubsection{ACE values of configuration options}\label{sec:ace_values}
Table~\ref{tab:ace_values} shows the corresponding ACE values of the configuration options on the performance objectives and constraints. We set the top $K = 5$, represented by blue. Note that \ourapproach reduces the search space from $34$ configuration options to $10$ by eliminating configuration options that do not affect the performance objective causally.

\subsubsection{Observational data additional details}
In Fig.~\ref{fig:obs_surf} and Fig.~\ref{fig:obs_surf_lowace}, we visualize the interactions between core configuration options~(pairwise) and their influence on the energy, position error, task success rate, and the safety constraint from the observational data. We observe that the surface response of configuration options with higher ACE values is complex than those with lower ACE values.
\section{Artifact Appendix}\label{sec:appendix_A}
This appendix provides additional information about \ourapproach. We describe the steps to reproduce the results reported in \S\ref{sec:exp}, and \S\ref{sec:discussion} using \ourapproach. The source code and data are provided in a publicly accessible GitHub repository, allowing users to test them on any hardware once the software dependencies are met.
\begin{tcolorbox}[colback=blue!5!white,colframe=blue!75!black]
    \vspace{-0.5em}
    \textbf{Code:} \url{https://github.com/softsys4ai/cure}
    \vspace{-0.5em}
\end{tcolorbox}
\subsection{Description} \ourapproach is used for tasks such as performance optimization and performance debugging in robotic systems. Given the cost and human involvement associated with collecting training data from physical robots for these tasks, \ourapproach addresses these challenges by learning the performance behavior of the robot in simulations and transferring the acquired knowledge to physical robots. \ourapproach also works with data from physical robots, we use simulator data to evaluate the transferability of the causal model.
\begin{itemize}
    \item In offline mode, \ourapproach is compatible with any device utilizing \textit{Husky} and \textit{Turtlebot} in \textit{Gazebo} environment.
    \item In online mode, the performance measurements are directly taken from the physical robot. In the experiments, we have used \textit{Turtlebot 3} platform. 
    \item In debugging mode, users can query the root cause of a certain functional and non-functional faults. For example, what is the root cause of the task failure and higher energy consumption? 
    \item \ourapproach can also be applied to a different robotic platform. However, an interface is required to read the telemetry data from the new robotic platform. We have provided a tool for that in~\S\ref{sec:reval}.    
\end{itemize}

\subsection{Setup}
\subsubsection{Software Dependencies} \emph{Ubuntu 20.04 LTS} and \emph{ROS Noetic} are prerequisites for using \ourapproach. Additionally, \ourapproach is implemented by integrating and building on top of several existing tools:
\begin{itemize}
    \item \href{https://github.com/cmu-phil/causal-learn}{\color{blue!80}causal-learn} for structure learning. 
    \item \href{https://ananke.readthedocs.io/en/latest/}{\color{blue!80}ananke} for estimating the causal effects.
    \item \href{https://ax.dev/}{\color{blue!80}Ax} to perform MOBO.    
\end{itemize}

\subsubsection{Hardware Dependencies} \ourapproach is implemented both in simulation and in physical robots. There are no particular hardware dependencies to run \ourapproach in simulation mode. To evaluate the transferability, we used \textit{Turtlebot 3} physical robot with \textit{ROS Noetic} and \emph{Ubuntu 20.04 LTS}.

\subsubsection{Installation} The installation of dependencies and third-party libraries essential for testing our approach can be accomplished using the following commands.
\begin{lstlisting}[basicstyle=\sffamily\small, frame=none, backgroundcolor=\color{blue!5}]
$ git clone git@github.com:softsys4ai/cure.git
$ cd ~/cure
$ sh requirements.sh
$ catkin build
$ source devel/setup.bash
\end{lstlisting}

\subsection{Getting measurements and observational data}\label{sec:reval}
To collect the observational data and measure the performance, we developed a tool, \href{https://github.com/softsys4ai/Reval}{\color{blue!80}\textit{Reval}}, which currently supports \textit{Husky} and \textit{Turtlebot-3}. Note that, observational data collection is optional since all the datasets required to run experiments are already included in the {\texTTT{./cure/data}} directory. However, \emph{Reval} is itegrated with \ourapproach and actively utilized for measurement during optimization. The following steps are solely for observational data collection. The observational data is stored in a {\texTTT{CSV}} file located in the {\texTTT{./cure/src/reval/results}} directory. The following commands can be used for observational data collection.
\begin{lstlisting}[basicstyle=\sffamily\small, frame=none, backgroundcolor=\color{blue!5}]
$ cd ~/cure/src/Reval
$ python reval_husky_sim.py -v off -e 10
\end{lstlisting}

\subsection{Outlier data} We generated $10$ outlier samples for both \textit{Husky} and \textit{Turtlebot 3}, each exhibiting different degrees of percentile variations in the performance. In particular, the outlier data contains configuration options where the robot's performance is worse than $80^{th}$ - $90^{th}$ percentile. We have included the outlier data in the {\texTTT{.cure/data/bug/}} directory.

\subsection{Training causal model} The causal model was trained on $1000$ observational data obtained using \emph{Reval}. The following commands can be used for training and saving the causal model. The saved model can later be utilized for both inference purposes and transferring knowledge. We have already included the saved models both for \textit{Husky} and \textit{Turtlebot 3} in the {\texTTT{.cure/model}} directory.
\begin{lstlisting}[basicstyle=\sffamily\small, frame=none, backgroundcolor=\color{blue!5}]
$ cd ~/cure
$ python run_cure_MOO.py --robot Husky_sim \\
--train_data data/husky_1000.csv 
\end{lstlisting}

\subsection{Identifying root causes} \ourapproach can perform debugging tasks such as identifying the root cause of a functional and non-functional fault. The following commands can be used to determine the root causes from the outlier data using the saved causal model. In this example, we have used {\texTTT{Task success rate}} as a functional property, and {\texTTT{Energy, Positional\_error}} as non-functional properties. We display the root causes in the terminal.
\begin{lstlisting}[basicstyle=\sffamily\small, frame=none, backgroundcolor=\color{blue!5}]
$ cd ~/cure
$ python run_cure_MOO.py --robot Husky_sim \\
-l --model model/care_Husky_sim.model \\
--outlier_data data/husky_outlier.csv \\
-root_cause  --f Task_success_rate \\
--nf Energy Positional_error
\end{lstlisting}

\subsection{Major Claims}
In this paper, we make the following major claims:
\begin{itemize}
    \item \ourapproach ensures optimal performance while efficiently utilizing the allocated budget by identifying the root causes of configuration bugs.
    \item The causal models are transferable across different environments~(Sim-to-real) and different robotic systems~(\textit{Husky} sim. to \textit{Turtlebot 3} physical). 
\end{itemize}

\subsection{Experiments} To support our claims, we perform the following experiments.
\subsubsection{Setup}
\begin{itemize}
    \item Install the dependencies for \href{https://emanual.robotis.com/docs/en/platform/turtlebot3/quick-start/}{\color{blue!80}\textit{Turtlebot 3}} physical robot.
    \item Run {\texTTT{roscore}} on remote PC.
    \item  Run {\texTTT{Bringup}} on \textit{Turtlebot 3 SBC}.
\end{itemize}

\subsubsection{E1: Optimizing robot performance with emphasis on faster convergence} To support this claim, we have (i) trained the causal model, (ii) generated a reduced search space by identifying the core configuration options, and (iii) performed MOBO on the reduced search space. We reproduce the results reported in Fig.~\ref{fig:rq1_performance} and Fig.~\ref{fig:rq1_budgetutil}. This experiment would require $\approx15$ hours to complete. We also compare the results with baseline methods.

\noindent \textit{Execution.} To run the experiment, the following commands need to be executed:
\begin{lstlisting}[basicstyle=\sffamily\small, frame=none, backgroundcolor=\color{blue!5}, caption=CURE]
$ cd ~/cure
$ python run_cure_MOO.py --robot Husky_sim \\
-l --model model/care_Husky_sim.model \\
--outlier_data data/husky_outlier.csv \\
-root_cause  --f Task_success_rate --nf \\
Energy Positional_error Obstacle_distance \\
--top_k 5 -opt --f1 Energy --f2 \\
Positional_error --f1_pref 40.0 --f2_pref \\
0.18 --sc 0.25 --tcr 0.8 --hv_ref_f1 400.0 \\ 
--hv_ref_f2 15 --budget 200
\end{lstlisting}
\begin{lstlisting}[basicstyle=\sffamily\small, frame=none, backgroundcolor=\color{blue!5}, caption=MOBO]
$ cd ~/cure
$ python run_baselineMOO.py --robot \\
Husky_sim --f1 Energy --f2 Positional_error \\
--f1_pref 40.0 --f2_pref 0.18 --sc 0.25 \\
--tcr 0.8 --hv_ref_f1 400.0 --hv_ref_f2 \\
15 --budget 200
\end{lstlisting}
\begin{lstlisting}[basicstyle=\sffamily\small, frame=none, backgroundcolor=\color{blue!5}, , caption=RidgeCV-MOBO]
$ cd ~/cure
$ python run_baselineSF_MOO.py --robot \\
Husky_sim --data data/husky_outlier.csv --f \\ 
-l --model model/RidgeCV_Husky_sim_model \\
Task_success_rate --nf Energy \\
Positional_error Obstacle_distance --top_k 5 \\
-opt --f1 Energy --f2 Positional_error \\
--f1_pref 40.0 --f2_pref 0.18 --sc 0.25 
--tcr 0.8 --hv_ref_f1 400.0 --hv_ref_f2 15 \\
--budget 200
\end{lstlisting}
\noindent \textit{Results.} The results reported in this paper are stored in a {\texTTT{CSV}} file located in the {\texTTT{./cure/cure\_log}} directory. Note that, during hypervolume computation, the execution might show warnings if the ovserved $f^{ref}$ points are higher than the defined points. Therefore, we have computed the hypervoulme after the experiments are over from using {\texTTT{CSV}} file. Note that this experiment is conducted once without repetition; thus, there are no error bars.
\subsubsection{E2: Demonstrating transferability} To support this claim, we trained the (i) causal model using observational data collected from \textit{Turtlebot 3} in simulation and reuse the causal model in \textit{Turtlebot 3} physical robot, and (ii) causal model using observational data collected from \textit{Husky} in simulation and reuse the causal model in \textit{Turtlebot 3} physical robot for performance optimization. This experiment is anticipated to require $\approx 20$ to $24$ hours, contingent on the time needed for the complete charging of the \textit{Turtlebot 3} physical robot's battery.

\noindent{\textit{Execution.}} The following commands need to be executed to run the experiment. 

\begin{lstlisting}[basicstyle=\sffamily\small, frame=none, backgroundcolor=\color{blue!5}, caption=CURE]
$ cd ~/cure
$ python run_cure_MOO.py --robot 
Turtlebot3_phy -l --model \\
model/care_Turtlebot_sim.model \\
-root_cause --outlier_data \\
data/turtlebot_phy_outlier.csv \\
--f Task_success_rate --nf Energy \\
Positional_error Obstacle_distance --top_k 5 \\
-opt --f1 Energy --f2 Positional_error \\
--f1_pref 2.0 --f2_pref 0.1 --sc 0.25 --tcr \\
0.8 --hv_ref_f1 19.98 --hv_ref_f2 3.0 \\
--init_trails 15 --budget 50
\end{lstlisting}
Replace {\texTTT{-{}-model}} parameter to \small{model/care\_Husky\_sim.model} for \textit{Husky sim.} to \textit{Turtlebot 3} phy. experiment.
\begin{lstlisting}[basicstyle=\sffamily\small, frame=none, backgroundcolor=\color{blue!5}, caption=MOBO]
$ cd ~/cure
$ python run_baselineMOO.py --robot \\
Turtlebot3_phy --f1 Energy --f2 -l_opt \\
--json model/optimodels/mobo/ \\
Turtlebot3_sim_ax_client_snapshot_201 \\
Positional_error --f1_pref 2.0 --f2_pref \\
0.1 --sc 0.25 --tcr 0.8 --hv_ref_f1 \\
19.98 --hv_ref_f2 3.0 --budget 50
\end{lstlisting}
Replace {\texTTT{-{}-json}} parameter to \small{Husky\_sim\_ax\_client\_snapshot\_200} for \textit{Husky sim.} to \textit{Turtlebot 3} phy. experiment.
\begin{lstlisting}[basicstyle=\sffamily\small, frame=none, backgroundcolor=\color{blue!5}, caption=RidgeCV-MOBO]
$ cd ~/cure
$ python run_baselineSF_MOO.py --robot \\
Turtlebot3_phy --data \\
data/turtlebot_phy_outlier.csv -l --model \\ 
model/RidgeCV_Turtlebot3_sim_model \\
--f Task_success_rate --nf Energy \\
Positional_error Obstacle_distance --top_k 5 \\
-opt --f1 Energy --f2 Positional_error \\
--f1_pref 2.0 --f2_pref 0.1 --sc 0.25 --tcr \\
0.8 --hv_ref_f1 19.98 --hv_ref_f2 3.0 \\
--init_trails 15 --budget 50
\end{lstlisting}
Replace {\texTTT{-{}-model}} parameter to \small{model/RidgeCV\_Husky\_sim\_model} for \textit{Husky sim.} to \textit{Turtlebot 3} phy. experiment.

\noindent{\textit{Results.}} We store the result in a {\texTTT{CSV}} file located in the {\texTTT{./cure/cure\_log}} directory.

\subsubsection{Real time result visualization}
To visualize the results in real time, execute {\texTTT{python live\_plot.py -{}-hv\_ref\_f1 19.98 -{}-hv\_ref\_f2 3}} in a separate terminal when an experiment is running.

\bibliographystyle{IEEEtran}
\bibliography{reference}

\begin{thebibliography}{10}
\providecommand{\url}[1]{#1}
\csname url@samestyle\endcsname
\providecommand{\newblock}{\relax}
\providecommand{\bibinfo}[2]{#2}
\providecommand{\BIBentrySTDinterwordspacing}{\spaceskip=0pt\relax}
\providecommand{\BIBentryALTinterwordstretchfactor}{4}
\providecommand{\BIBentryALTinterwordspacing}{\spaceskip=\fontdimen2\font plus
\BIBentryALTinterwordstretchfactor\fontdimen3\font minus
  \fontdimen4\font\relax}
\providecommand{\BIBforeignlanguage}[2]{{%
\expandafter\ifx\csname l@#1\endcsname\relax
\typeout{** WARNING: IEEEtran.bst: No hyphenation pattern has been}%
\typeout{** loaded for the language `#1'. Using the pattern for}%
\typeout{** the default language instead.}%
\else
\language=\csname l@#1\endcsname
\fi
#2}}
\providecommand{\BIBdecl}{\relax}
\BIBdecl

\bibitem{khalastchi2018fault}
E.~Khalastchi and M.~Kalech, ``On fault detection and diagnosis in robotic
  systems,'' \emph{ACM Computing Surveys (CSUR)}, vol.~51, no.~1, pp. 1--24,
  2018.

\bibitem{kim2023patchverif}
H.~Kim, M.~O. Ozmen \emph{et~al.}, ``Patchverif: Discovering faulty patches in
  robotic vehicles,'' in \emph{32nd USENIX Security Symposium (USENIX Security
  23)}, 2023, pp. 3011--3028.

\bibitem{jung2021swarmbug}
C.~Jung, A.~Ahad \emph{et~al.}, ``Swarmbug: debugging configuration bugs in
  swarm robotics,'' in \emph{Proceedings of the 29th ACM Joint Meeting on
  European Software Engineering Conference and Symposium on the Foundations of
  Software Engineering}, 2021, pp. 868--880.

\bibitem{10137745}
M.~A. Hossen, S.~Kharade \emph{et~al.}, ``Care: Finding root causes of
  configuration issues in highly-configurable robots,'' \emph{IEEE Robotics and
  Automation Letters}, vol.~8, no.~7, pp. 4115--4122, 2023.

\bibitem{xie2022rozz}
K.-T. Xie, J.-J. Bai \emph{et~al.}, ``Rozz: Property-based fuzzing for robotic
  programs in ros,'' in \emph{2022 International Conference on Robotics and
  Automation (ICRA)}.\hskip 1em plus 0.5em minus 0.4em\relax IEEE, 2022, pp.
  6786--6792.

\bibitem{kim2019rvfuzzer}
T.~Kim, C.~H. Kim \emph{et~al.}, ``Rvfuzzer: Finding input validation bugs in
  robotic vehicles through control-guided testing.'' in \emph{USENIX Security
  Symposium}, 2019, pp. 425--442.

\bibitem{wang2021exploratory}
D.~Wang, S.~Li \emph{et~al.}, ``An exploratory study of autopilot software bugs
  in unmanned aerial vehicles,'' in \emph{Proceedings of the 29th ACM Joint
  Meeting on European Software Engineering Conference and Symposium on the
  Foundations of Software Engineering}, 2021, pp. 20--31.

\bibitem{garcia2020comprehensive}
J.~Garcia, Y.~Feng \emph{et~al.}, ``A comprehensive study of autonomous vehicle
  bugs,'' in \emph{Proceedings of the ACM/IEEE 42nd international conference on
  software engineering}, 2020, pp. 385--396.

\bibitem{kim2022drivefuzz}
S.~Kim, M.~Liu \emph{et~al.}, ``Drivefuzz: Discovering autonomous driving bugs
  through driving quality-guided fuzzing,'' in \emph{Proceedings of the 2022
  ACM SIGSAC Conference on Computer and Communications Security}, 2022, pp.
  1753--1767.

\bibitem{valle2023automated}
P.~Valle, A.~Arrieta \emph{et~al.}, ``Automated misconfiguration repair of
  configurable cyber-physical systems with search: an industrial case study on
  elevator dispatching algorithms,'' \emph{arXiv preprint arXiv:2301.01487},
  2023.

\bibitem{configOpti}
\BIBentryALTinterwordspacing
Challenges in optimizing configurations for robotic systems in real-world
  scenarios. [Online]. Available:
  \url{https://github.com/softsys4ai/cure/wiki/Real-World-Optimization-Issues}
\BIBentrySTDinterwordspacing

\bibitem{tuningfeature}
\BIBentryALTinterwordspacing
Automatic parameter tuning feature request in {ROS-2} navigation. [Online].
  Available: \url{https://github.com/ros-planning/navigation2/issues/1276}
\BIBentrySTDinterwordspacing

\bibitem{plannerfailing}
\BIBentryALTinterwordspacing
Local planner performance problem. in {ROS-2} navigation. [Online]. Available:
  \url{https://github.com/ros-planning/navigation2/issues/2439}
\BIBentrySTDinterwordspacing

\bibitem{Ax}
\BIBentryALTinterwordspacing
Facebook. Adaptive experimentation platform. [Online]. Available:
  \url{https://ax.dev/docs/why-ax.html}
\BIBentrySTDinterwordspacing

\bibitem{hoerl1970ridge}
A.~E. Hoerl and R.~W. Kennard, ``Ridge regression: Biased estimation for
  nonorthogonal problems,'' \emph{Technometrics}, vol.~12, no.~1, pp. 55--67,
  1970.

\bibitem{ridgeCV}
\BIBentryALTinterwordspacing
scikit learn. Ridgecv. [Online]. Available:
  \url{https://scikit-learn.org/stable/modules/generated/sklearn.linear_model.RidgeCV.html}
\BIBentrySTDinterwordspacing

\bibitem{binch2020context}
A.~Binch, G.~P. Das \emph{et~al.}, ``Context dependant iterative parameter
  optimisation for robust robot navigation,'' in \emph{2020 IEEE International
  Conference on Robotics and Automation (ICRA)}.\hskip 1em plus 0.5em minus
  0.4em\relax IEEE, 2020, pp. 3937--3943.

\bibitem{zhou2011multiobjective}
A.~Zhou, B.-Y. Qu \emph{et~al.}, ``Multiobjective evolutionary algorithms: A
  survey of the state of the art,'' \emph{Swarm and evolutionary computation},
  vol.~1, no.~1, pp. 32--49, 2011.

\bibitem{fox1997dynamic}
D.~Fox, W.~Burgard \emph{et~al.}, ``The dynamic window approach to collision
  avoidance,'' \emph{IEEE Robotics \& Automation Magazine}, vol.~4, no.~1, pp.
  23--33, 1997.

\bibitem{ariizumi2016multiobjective}
R.~Ariizumi, M.~Tesch \emph{et~al.}, ``Multiobjective optimization based on
  expensive robotic experiments under heteroscedastic noise,'' \emph{IEEE
  Transactions on Robotics}, vol.~33, no.~2, pp. 468--483, 2016.

\bibitem{berkenkamp2023bayesian}
F.~Berkenkamp, A.~Krause \emph{et~al.}, ``Bayesian optimization with safety
  constraints: safe and automatic parameter tuning in robotics,'' \emph{Machine
  Learning}, vol. 112, no.~10, pp. 3713--3747, 2023.

\bibitem{argall2009survey}
B.~D. Argall, S.~Chernova \emph{et~al.}, ``A survey of robot learning from
  demonstration,'' \emph{Robotics and autonomous systems}, vol.~57, no.~5, pp.
  469--483, 2009.

\bibitem{perez2018learning}
N.~P{\'e}rez-Higueras, F.~Caballero \emph{et~al.}, ``Learning human-aware path
  planning with fully convolutional networks,'' in \emph{2018 IEEE
  international conference on robotics and automation (ICRA)}.\hskip 1em plus
  0.5em minus 0.4em\relax IEEE, 2018, pp. 5897--5902.

\bibitem{pfeiffer2017perception}
M.~Pfeiffer, M.~Schaeuble \emph{et~al.}, ``From perception to decision: A
  data-driven approach to end-to-end motion planning for autonomous ground
  robots,'' in \emph{2017 ieee international conference on robotics and
  automation (icra)}.\hskip 1em plus 0.5em minus 0.4em\relax IEEE, 2017, pp.
  1527--1533.

\bibitem{kahn2021badgr}
G.~Kahn, P.~Abbeel \emph{et~al.}, ``Badgr: An autonomous self-supervised
  learning-based navigation system,'' \emph{IEEE Robotics and Automation
  Letters}, vol.~6, no.~2, pp. 1312--1319, 2021.

\bibitem{chen2022performance}
T.~Chen and M.~Li, ``Do performance aspirations matter for guiding software
  configuration tuning? an empirical investigation under dual performance
  objectives,'' \emph{ACM Transactions on Software Engineering and
  Methodology}, 2022.

\bibitem{siegmund2015performance}
N.~Siegmund, A.~Grebhahn \emph{et~al.}, ``Performance-influence models for
  highly configurable systems,'' in \emph{Proceedings of the 2015 10th Joint
  Meeting on Foundations of Software Engineering}, 2015, pp. 284--294.

\bibitem{8919029}
H.~Ha and H.~Zhang, ``Performance-influence model for highly configurable
  software with fourier learning and lasso regression,'' in \emph{2019 IEEE
  International Conference on Software Maintenance and Evolution (ICSME)},
  2019, pp. 470--480.

\bibitem{rai2019using}
A.~Rai, R.~Antonova \emph{et~al.}, ``Using simulation to improve
  sample-efficiency of bayesian optimization for bipedal robots,'' \emph{The
  Journal of Machine Learning Research}, vol.~20, no.~1, pp. 1844--1867, 2019.

\bibitem{kaushik2022safeapt}
R.~Kaushik, K.~Arndt \emph{et~al.}, ``{SafeAPT}: Safe simulation-to-real robot
  learning using diverse policies learned in simulation,'' \emph{IEEE Robotics
  and Automation Letters}, vol.~7, no.~3, pp. 6838--6845, 2022.

\bibitem{valov2020transferring}
P.~Valov, J.~Guo \emph{et~al.}, ``Transferring {P}areto frontiers across
  heterogeneous hardware environments,'' in \emph{Proceedings of the ACM/SPEC
  International Conference on Performance Engineering}, 2020, pp. 12--23.

\bibitem{iqbal2023cameo}
M.~S. Iqbal, Z.~Zhong \emph{et~al.}, ``Cameo: A causal transfer learning
  approach for performance optimization of configurable computer systems,''
  \emph{arXiv preprint arXiv:2306.07888}, 2023.

\bibitem{zhou2021examining}
C.~Zhou, X.~Ma \emph{et~al.}, ``Examining and combating spurious features under
  distribution shift,'' in \emph{International Conference on Machine
  Learning}.\hskip 1em plus 0.5em minus 0.4em\relax PMLR, 2021, pp.
  12\,857--12\,867.

\bibitem{pearl2009causality}
J.~Pearl, \emph{Causality}.\hskip 1em plus 0.5em minus 0.4em\relax Cambridge
  university press, 2009.

\bibitem{spirtes2000causation}
P.~Spirtes, C.~N. Glymour \emph{et~al.}, \emph{Causation, prediction, and
  search}.\hskip 1em plus 0.5em minus 0.4em\relax MIT press, 2000.

\bibitem{fariha2020causality}
A.~Fariha, S.~Nath \emph{et~al.}, ``Causality-guided adaptive interventional
  debugging,'' in \emph{Proceedings of the 2020 ACM SIGMOD International
  Conference on Management of Data}, 2020, pp. 431--446.

\bibitem{johnson2020causal}
B.~Johnson, Y.~Brun \emph{et~al.}, ``Causal testing: understanding defects'
  root causes,'' in \emph{Proceedings of the ACM/IEEE 42nd International
  Conference on Software Engineering}, 2020, pp. 87--99.

\bibitem{DubslaffCausality2022}
C.~Dubslaff, K.~Weis \emph{et~al.}, ``Causality in configurable software
  systems,'' in \emph{Proceedings of the 44th International Conference on
  Software Engineering}, ser. ICSE '22.\hskip 1em plus 0.5em minus 0.4em\relax
  New York, NY, USA: Association for Computing Machinery, 2022, p. 325–337.

\bibitem{9402143}
Y.~Küçük, T.~A.~D. Henderson \emph{et~al.}, ``Improving fault localization
  by integrating value and predicate based causal inference techniques,'' in
  \emph{2021 IEEE/ACM 43rd International Conference on Software Engineering
  (ICSE)}, 2021, pp. 649--660.

\bibitem{diehl2022did}
M.~Diehl and K.~Ramirez-Amaro, ``Why did i fail? a causal-based method to find
  explanations for robot failures,'' \emph{IEEE Robotics and Automation
  Letters}, vol.~7, no.~4, pp. 8925--8932, 2022.

\bibitem{abdessalem2020automated}
R.~B. Abdessalem, A.~Panichella \emph{et~al.}, ``Automated repair of feature
  interaction failures in automated driving systems,'' in \emph{Proceedings of
  the 29th ACM SIGSOFT International Symposium on Software Testing and
  Analysis}, 2020, pp. 88--100.

\bibitem{lillicrap2015continuous}
T.~P. Lillicrap, J.~J. Hunt \emph{et~al.}, ``Continuous control with deep
  reinforcement learning,'' \emph{arXiv preprint arXiv:1509.02971}, 2015.

\bibitem{gupta2018robot}
A.~Gupta, A.~Murali \emph{et~al.}, ``Robot learning in homes: Improving
  generalization and reducing dataset bias,'' \emph{Advances in neural
  information processing systems}, vol.~31, 2018.

\bibitem{navcore}
\BIBentryALTinterwordspacing
{ROS} navigation stack. [Online]. Available: \url{http://wiki.ros.org/nav_core}
\BIBentrySTDinterwordspacing

\bibitem{glymour2019review}
C.~Glymour, K.~Zhang \emph{et~al.}, ``Review of causal discovery methods based
  on graphical models,'' \emph{Frontiers in genetics}, vol.~10, p. 524, 2019.

\bibitem{connelly2016fisher}
L.~M. Connelly, ``Fisher's exact test,'' \emph{MedSurg Nursing}, vol.~25,
  no.~1, pp. 58--60, 2016.

\bibitem{colombo2012learning}
D.~Colombo, M.~H. Maathuis \emph{et~al.}, ``Learning high-dimensional directed
  acyclic graphs with latent and selection variables,'' \emph{The Annals of
  Statistics}, pp. 294--321, 2012.

\bibitem{Kocaoglu2020}
M.~Kocaoglu, S.~Shakkottai \emph{et~al.}, ``Applications of common entropy for
  causal inference,'' in \emph{Advances in Neural Information Processing
  Systems}, H.~Larochelle, M.~Ranzato \emph{et~al.}, Eds., vol.~33.\hskip 1em
  plus 0.5em minus 0.4em\relax Curran Associates, Inc., 2020, pp.
  17\,514--17\,525.

\bibitem{shahriari2015taking}
B.~Shahriari, K.~Swersky \emph{et~al.}, ``Taking the human out of the loop: A
  review of bayesian optimization,'' \emph{Proceedings of the IEEE}, vol. 104,
  no.~1, pp. 148--175, 2015.

\bibitem{knowles2006parego}
J.~Knowles, ``Parego: A hybrid algorithm with on-line landscape approximation
  for expensive multiobjective optimization problems,'' \emph{IEEE transactions
  on evolutionary computation}, vol.~10, no.~1, pp. 50--66, 2006.

\bibitem{hernandez2016predictive}
D.~Hern{\'a}ndez-Lobato, J.~Hernandez-Lobato \emph{et~al.}, ``Predictive
  entropy search for multi-objective bayesian optimization,'' in
  \emph{International conference on machine learning}.\hskip 1em plus 0.5em
  minus 0.4em\relax PMLR, 2016, pp. 1492--1501.

\bibitem{ponweiser2008multiobjective}
W.~Ponweiser, T.~Wagner \emph{et~al.}, ``Multiobjective optimization on a
  limited budget of evaluations using model-assisted-metric selection,'' in
  \emph{International conference on parallel problem solving from
  nature}.\hskip 1em plus 0.5em minus 0.4em\relax Springer, 2008, pp. 784--794.

\bibitem{daulton2020differentiable}
S.~Daulton, M.~Balandat \emph{et~al.}, ``Differentiable expected hypervolume
  improvement for parallel multi-objective bayesian optimization,''
  \emph{Advances in Neural Information Processing Systems}, vol.~33, pp.
  9851--9864, 2020.

\bibitem{gpml}
C.~K. Williams and C.~E. Rasmussen, ``Gaussian processes for machine
  learning,'' \emph{MIT Press}, 2006.

\bibitem{hutter2009automated}
F.~Hutter, ``Automated configuration of algorithms for solving hard
  computational problems,'' Ph.D. dissertation, University of British Columbia,
  2009.

\bibitem{shahriaritaking}
B.~Shahriari \emph{et~al.}, ``Taking the human out of the loop: a review of
  bayesian optimization,'' Tech. Rep., 2015.

\bibitem{moveit}
\BIBentryALTinterwordspacing
{ROS} {M}ove{I}t. [Online]. Available: \url{https://moveit.ros.org/}
\BIBentrySTDinterwordspacing

\bibitem{cao2015using}
Y.~Cao, B.~J. Smucker \emph{et~al.}, ``On using the hypervolume indicator to
  compare pareto fronts: Applications to multi-criteria optimal experimental
  design,'' \emph{Journal of Statistical Planning and Inference}, vol. 160, pp.
  60--74, 2015.

\bibitem{zitzler1999multiobjective}
E.~Zitzler and L.~Thiele, ``Multiobjective evolutionary algorithms: a
  comparative case study and the strength pareto approach,'' \emph{IEEE
  transactions on Evolutionary Computation}, vol.~3, no.~4, pp. 257--271, 1999.

\bibitem{zimmerman2017metrics}
T.~A. Zimmerman, \emph{Metrics and key performance indicators for robotic
  cybersecurity performance analysis}, 2017.

\bibitem{falco2018performance}
J.~Falco, K.~Van~Wyk \emph{et~al.}, ``Performance metrics and test methods for
  robotic hands,'' \emph{DRAFT NIST Special Publication}, vol. 1227, pp. 2--2,
  2018.

\bibitem{chickering2004large}
M.~Chickering, D.~Heckerman \emph{et~al.}, ``Large-sample learning of bayesian
  networks is np-hard,'' \emph{Journal of Machine Learning Research}, vol.~5,
  pp. 1287--1330, 2004.

\bibitem{dubslaff2022causality}
C.~Dubslaff, K.~Weis \emph{et~al.}, ``Causality in configurable software
  systems,'' in \emph{Proceedings of the 44th International Conference on
  Software Engineering}, 2022, pp. 325--337.

\bibitem{hossen2023care}
M.~A. Hossen, S.~Kharade \emph{et~al.}, ``Care: Finding root causes of
  configuration issues in highly-configurable robots,'' \emph{IEEE Robotics and
  Automation Letters}, vol.~8, no.~7, pp. 415--4122, Jul. 2023.

\bibitem{aglietti20a}
V.~Aglietti, X.~Lu \emph{et~al.}, ``Causal bayesian optimization,'' in
  \emph{Proceedings of the Twenty Third International Conference on Artificial
  Intelligence and Statistics}, ser. Proceedings of Machine Learning Research,
  S.~Chiappa and R.~Calandra, Eds., vol. 108.\hskip 1em plus 0.5em minus
  0.4em\relax PMLR, 26--28 Aug 2020, pp. 3155--3164.

\bibitem{colombo2014order}
D.~Colombo and M.~H. Maathuis, ``Order-independent constraint-based causal
  structure learning,'' \emph{The Journal of Machine Learning Research},
  vol.~15, no.~1, pp. 3741--3782, 2014.

\bibitem{pearl2016causal}
J.~Pearl, M.~Glymour \emph{et~al.}, \emph{Causal inference in statistics: A
  primer}.\hskip 1em plus 0.5em minus 0.4em\relax John Wiley \& Sons, 2016.

\bibitem{richardson2002ancestral}
T.~Richardson and P.~Spirtes, ``Ancestral graph markov models,'' \emph{The
  Annals of Statistics}, vol.~30, no.~4, pp. 962--1030, 2002.

\bibitem{evans2014markovian}
R.~J. Evans and T.~S. Richardson, ``{Markovian acyclic directed mixed graphs
  for discrete data},'' \emph{The Annals of Statistics}, vol.~42, no.~4, pp.
  1452 -- 1482, 2014.

\end{thebibliography}
\begin{IEEEbiography}[{\includegraphics[width=1in,height=1.25in,keepaspectratio]{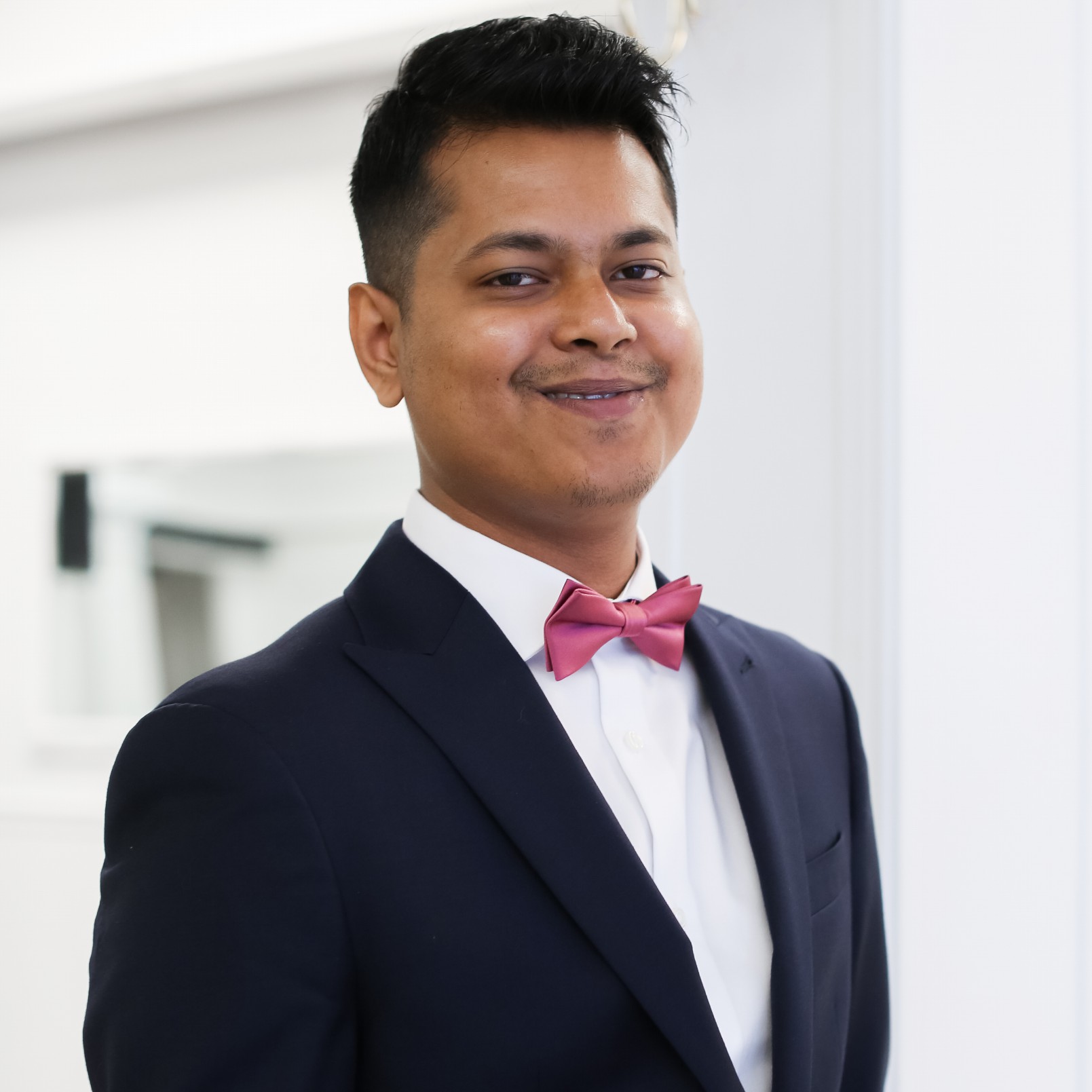}}]%
{Md Abir Hossen} received the B.S. degree in Electrical and Electronics Engineering from American International University-Bangladesh, Dhaka, Bangladesh, in 2017
and the M.S. degree in Electrical Engineering from South Dakota School of Mines and Technology, Rapid City, SD, USA in 2021. He is currently working toward a Ph.D. degree in Computer Science with the Artificial Intelligence and Systems Laboratory~(AISys), University of South Carolina, SC, USA. 

His research interests include artificial intelligence and robot learning. Further information about his research can be found at \url{https://sites.google.com/view/abirhossen/}.
\end{IEEEbiography}

\vspace{-0.9cm}

\begin{IEEEbiography}[{\includegraphics[width=1in,height=1.25in,keepaspectratio]{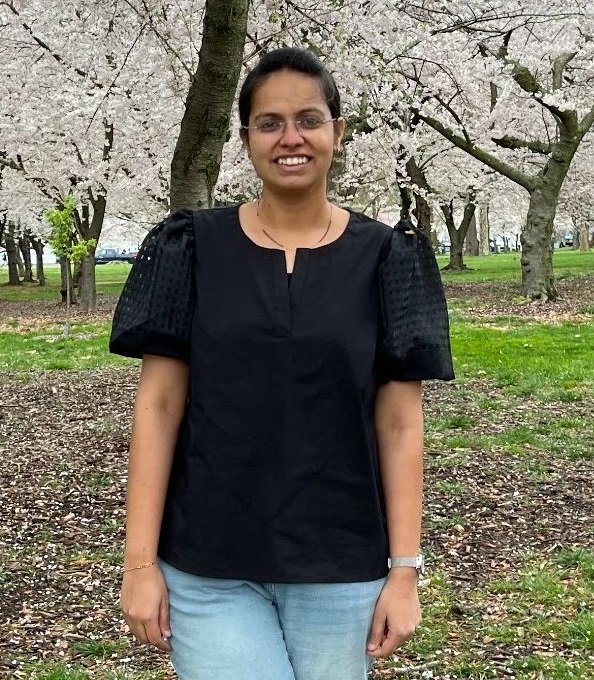}}]%
{Sonam Kharade} received the M.Tech., B.Tech., and Ph.D. degree in Electrical Engineering from Veermata Jijabai Technological Institute (VJTI), Mumbai, India. She was previously a Postdoc at the AISys, University of South Carolina, SC and is now a Postdoc at Argonne National Laboratory, USA. Her research focuses on the development of mathematical frameworks in control theory, incorporating machine learning techniques, and applying them to practical problems across various domains, such as robotics, and power systems.

\end{IEEEbiography}

\vspace{-0.9cm}

\begin{IEEEbiography}[{\includegraphics[width=1in,height=1.25in,keepaspectratio]{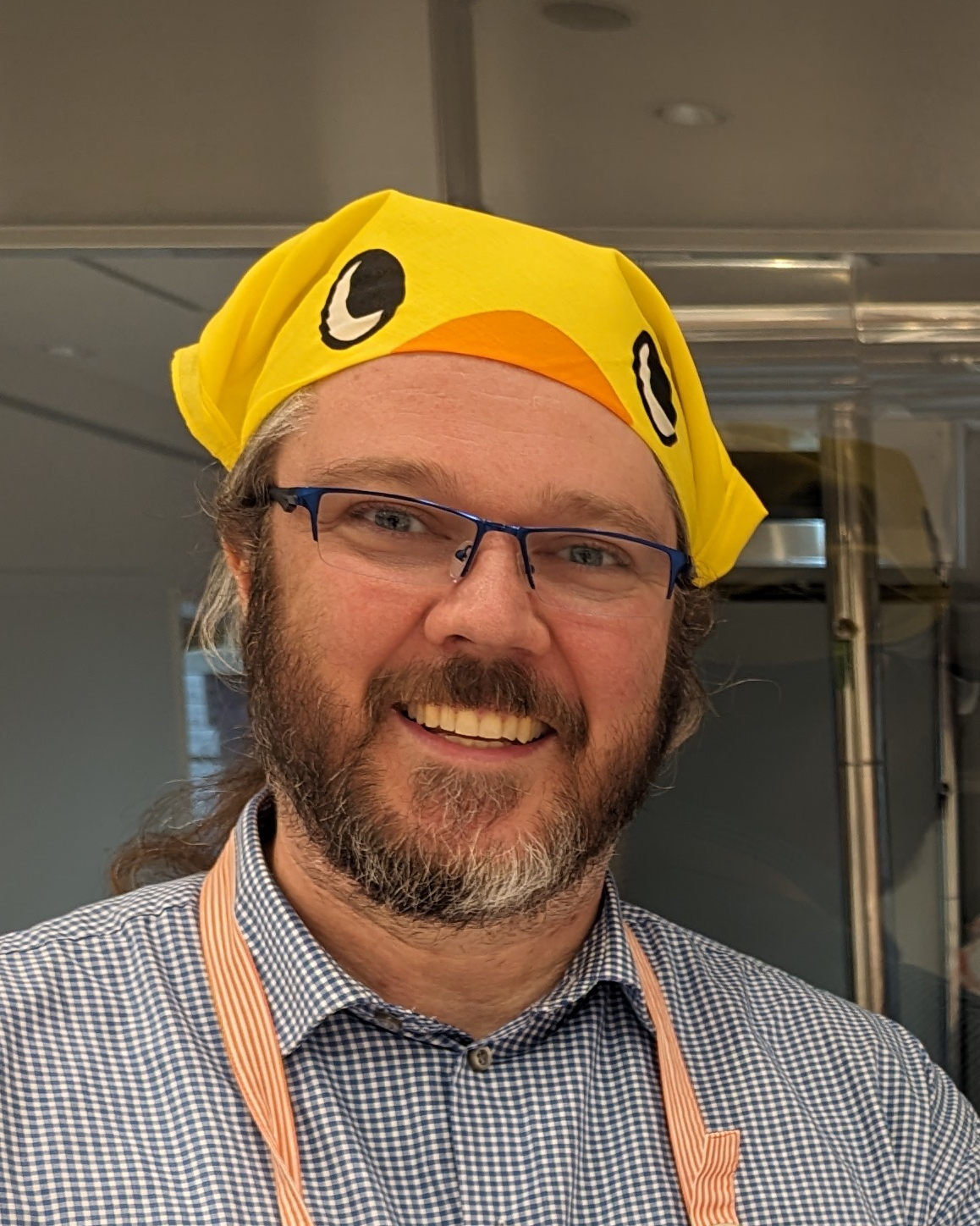}}]%
{Jason M. O'Kane} is Professor of Computer Science and Engineering at Texas A\&M University, TX, USA.
He received the B.S. degree from Taylor University, Upland, IN, USA and the M.S. and Ph.D. degrees from the University of Illinois at Urbana-Champaign, Urbana, IL, USA, all in computer science. His research interests include algorithmic robotics, planning under uncertainty, artificial intelligence, computational geometry, and motion planning.
\end{IEEEbiography}

\vspace{-0.9cm}

\begin{IEEEbiography}[{\includegraphics[width=1in,height=1.25in,keepaspectratio]{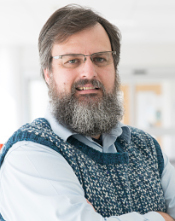}}]%
{Bradley Schmerl} a principal systems scientist at the Institute for Software Research at Carnegie Mellon University, Pittsburgh. His research interests include software architecture, self-adaptive systems, and software engineering tools. Schmerl received a Ph.D. in computer science from Flinders University, Adelaide, Australia. Contact him at schmerl@cs.cmu.edu.
\end{IEEEbiography}

\vspace{-0.9cm}

\begin{IEEEbiography}[{\includegraphics[width=1in,height=1.25in,keepaspectratio]{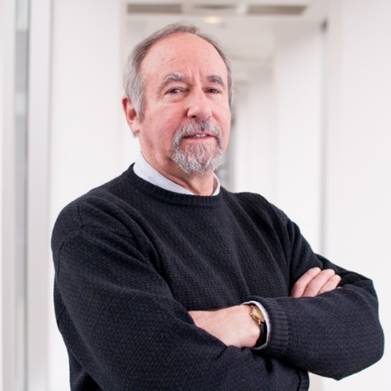}}]%
{David Garlan} is a professor and associate dean of the School of Computer Science at Carnegie Mellon University, Pittsburgh, Pennsylvania, 15213, USA. His research interests include autonomous and self-adaptive systems, software architecture, formal methods, explainability, and cyberphysical systems. Garlan received a Ph.D. in computer science from Carnegie Mellon University. He is a Fellow of the IEEE. More information about him can be found at \url{https://www.cs.cmu.edu/~garlan/}. Contact him at garlan@cs.cmu.edu.
\end{IEEEbiography}

\vspace{-0.9cm}

\begin{IEEEbiography}[{\includegraphics[width=1in,height=1.25in,keepaspectratio]{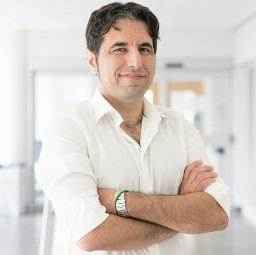}}]%
{Pooyan Jamshidi} is an assistant professor in the Department of Computer Science and Engineering at the University of South Carolina, where he directs the AISys lab. He holds a Ph.D. in Computer Science from Dublin City University and has completed postdoctoral research at Carnegie Mellon University and Imperial College London. Pooyan has also worked in the industry; most recently, he was a visiting researcher at Google in 2021. Dr. Jamshidi, who received the USC 2022 Breakthrough Stars Award, specializes in developing resilient systems for dynamic environments. His work integrates various areas such as distributed systems, statistical and causal learning, and robotics, focusing on areas like autonomous systems, AI accelerators, and software/hardware co-design.
\end{IEEEbiography}\vfill

\end{document}